\documentclass[lettersize,journal]{IEEEtran}
\usepackage{amsmath,amsfonts}
\usepackage{algorithmic}
\usepackage{algorithm}
\usepackage{array}
\usepackage[caption=false,font=normalsize,labelfont=sf,textfont=sf]{subfig}
\usepackage{textcomp}
\usepackage{stfloats}
\usepackage{url}
\usepackage{verbatim}
\usepackage{graphicx}
\usepackage{cite}
\usepackage{booktabs} 
\usepackage{mwe}
\usepackage{graphbox}
\usepackage{amssymb}
\usepackage{mathabx}
\usepackage{mathtools}
\usepackage{amsthm}
\usepackage{multirow}
\DeclareMathOperator{\Tr}{Tr}

\DeclareMathOperator*{\argmin}{arg\,min}
\usepackage{nicefrac}
\usepackage{booktabs}
\usepackage[dvipsnames]{xcolor}

\begin{document}

\title{Graph Laplacian Learning with Exponential Family Noise}

\author{Changhao~Shi and Gal~Mishne
\thanks{C. Shi is with the Electrical and Computer Engineering Department, UC San Diego, CA 92093 USA (e-mail: cshi@ucsd.edu). G. Mishne is with the Halıcıoğlu Data Science Institute and the Neurosciences Graduate Program, UC San Diego, CA 92093 USA (e-mail: gmishne@ucsd.edu). 
}}



\maketitle

\begin{abstract}
Graph signal processing (GSP) is a prominent framework for analyzing signals on non-Euclidean domains.
The graph Fourier transform (GFT) uses the combinatorial graph Laplacian matrix to reveal the spectral decomposition of signals in the graph frequency domain.
However, a common challenge in applying GSP methods is that in many scenarios the underlying graph of a system is unknown.
A solution in such cases is to construct the unobserved graph from available data, which is commonly referred to as graph or network inference.
Although different graph inference methods exist, these are restricted to learning from either smooth graph signals or simple additive Gaussian noise.
Other types of noisy data, such as discrete counts or binary digits, are rather common in real-world applications, yet are underexplored in graph inference. 
In this paper, we propose a versatile graph inference framework for learning from graph signals corrupted by exponential family noise.
Our framework generalizes previous methods from continuous smooth graph signals to various data types.
We propose an alternating algorithm that jointly estimates the graph Laplacian and the unobserved smooth representation from the noisy signals.
We also extend our approach to a variational form to account for the inherent stochasticity of the latent smooth representation.
Finally, since real-world graph signals are frequently non-independent and temporally correlated, we further adapt our original setting to a time-vertex formulation.
We demonstrate on synthetic and real-world data that our new algorithms outperform competing Laplacian estimation methods that suffer from noise model mismatch.
\end{abstract}

\begin{IEEEkeywords}
Network inference, graph learning, graph signal processing, exponential family distributions.
\end{IEEEkeywords}

\section{Introduction}
\label{sec:intro}

\IEEEPARstart{G}{raphs} are ubiquitous in our world. 
From neural systems to molecules, from social networks to traffic flow, objects in assorted fields and their connections can be naturally represented abstractly as graphs.
Studying these abstractions and the data associated with them has been proven to be vital in understanding these scientific fields \cite{bullmore2009complex,trinajstic2018chemical,robins2007introduction,latora2002boston}. 

Graph signal processing (GSP) provides an elegant yet powerful framework for studying these geometric objects and serves as the foundation of many other modern geometric machine learning approaches.
GSP analyzes data or signals on irregular domains, generalizing from signals on a regular grid such as time series to signals on networks and graphs~\cite{ortega2018graph}.
The graph shift operator (GSO), which generalizes the temporal shift operator in traditional signal processing, is chosen to be a graph representation such as the graph adjacency matrix and the graph Laplacian. 
The eigenvectors of the GSO not only encompass rich geometric information about the graph but also transform spatial signals to the spectral domain.
Thus, by making the eigenvectors of the GSO graph Fourier basis, GSP extends Fourier analysis to non-Euclidean domains. 
Functions of the eigenvalues of GSO are called graph filters, lifting or dampening different spectral components of graph signals.
However, applying the graph Fourier transform (GFT) requires the graph of the system to be known, which is not the case in many real-world problems.
When the graph is unknown, one needs resort to various graph inference methods~\cite{mateos2019connecting} to apply GSP or other graph machine learning methods.

Graph inference is a prominent problem in GSP.
With only graph signals available, learning the graph structure is possible when imposing various geometric priors on the signals~\cite{Shen2017,Giannakis2018,mateos2019connecting,li2021graph}.
Arguably the most prevalent is to assume that the observed signals are smooth with respect to the graph to be inferred.
Established methods usually find the graph adjacency matrix or, more commonly, the graph Laplacian, so that total variation of given signals, a measurement of smoothness, will be minimal \cite{hu2015spectral,dong2016learning,kalofolias2016learn,egilmez2017graph,kumar2020unified}.
However, smooth graph signals are rare in the real world, and one is often required to deal with noisy signals which contradicts this assumption.
The smooth model implies that signals are continuous and unbounded, while in the real world, signals can be discrete, categorical, or even binary in extreme cases. 
One way to bypass this problem is to disregard the noise and treat the noisy signals as smooth signals, especially when the noise is assumed to be considerably small.
Although such compromises may yield satisfying results, ignoring the noise is not always appropriate. 
In neuroscience, for example, graph learning methods have been successfully applied to fMRI data to learn functional connectivity between brain regions~\cite{hu2015spectral,WANG2020108649,gao2021smooth}. However, at the cellular level, electrophysiology recordings consist of binary spiking patterns, and to infer a graph of neurons, i.e. corresponding to their functional or anatomical connectivity, the non-continuous and non-negative nature of spiking signals posits difficulty on the existing smooth graph inference methods.
When smoothness is not even properly defined, new graph learning approaches are necessary.

Fortunately, a remedy for the smooth model is feasible, owing to its probabilistic interpretation \cite{dong2016learning,dong2019learning}. 
For the Laplacian-based smooth model, with the graph Laplacian properly regularized, graph inference coincides with the maximum likelihood estimation (MLE) of an improper Gaussian distribution of smooth signals.
More interestingly, Dong et al.~\cite{dong2016learning} considered the case where the noise is additive Gaussian, and formulated graph inference as the maximum a-posterior (MAP) of the unobserved Gaussian smooth signals.
In the same vein as generalized linear models (GLM) generalizing linear models, an appropriate noise distribution can substitute the Gaussian distribution to account for a specific response.
More precisely, we can model the observations by a conditional response distribution whose expected value is parameterized by their underlying smooth representation.
This new layer of hierarchy can adapt the smooth model to other data types, with the underlying smooth signal kept intact for the use of GSP.

In this paper, we study the problem of learning graph Laplacian matrices from graph signals of various types.
We first generalize a GSP-based framework that generates graph signals with Gaussian noise~\cite{dong2016learning} to a versatile framework that is compatible with all exponential family noise.
We then propose \textbf{G}raph \textbf{L}earning with \textbf{E}xponential family \textbf{N}oise (\textbf{GLEN}), an algorithm that estimates the underlying graph Laplacian from the noisy signals without knowing the smooth representation.
Our generalized method provides a unified solution for graph Laplacian learning when the observed data type varies.
We present two concrete examples of this generalized scenario with Poisson and Bernoulli noise.
Furthermore, we extend GLEN to the time-vertex setting~\cite{grassi2017time} to handle data with temporal correlations, e.g., time series on a network.
We demonstrate in synthetic experiments with different graph models that GLEN is competitive against off-the-shelf Laplacian estimation methods under noise model mismatch.
Furthermore, we apply GLEN to different types of real-world data (e.g., binary questionnaires and neural activity) to further demonstrate the efficacy of our methods.
Note that our goal is to learn rigorous combinatorial graph Laplacian matrices, which is different from previous work that aims to learn general precision matrices \cite{biswas2016learning,chiquet2019variational}.

In summary, our contributions are as follows:
\begin{enumerate}
\setlength{\itemsep}{0.5pt}
    \item We establish a generalized GSP-based framework that models graph signals of different data types using exponential family distributions;
    \item We propose an alternating algorithm for joint graph learning and signal denoising under this generalized framework;
    \item We extend our framework to temporally smooth signals using the time-vertex formulation.
\end{enumerate}

The rest of the paper is organized as follows.
We review related work on graph inference in Section~\ref{sec:related-work} and introduce mathematical background in Section~\ref{sec:background}.
In Section~\ref{sec:generalize_graph_inference} we poresent our approach to generalize graph inference beyond simple Gaussian noise to exponential family noise and propose an alternating algorithm for learning the unknown graph.
We evaluate our methods on synthetic and real-world datasets in Section~\ref{sec:experiments}.

We use the following notation throughout the paper.
Lowercase and uppercase bold letters denote vectors and matrices, respectively, and lowercase bold italic letters denote random vectors.
Let $\mathbf{S}\mathbb{R}^{N \times N}$ and $\mathbf{S}\mathbb{R}_{+}^{N \times N}$ denote the set of symmetric real matrices and symmetric real nonnegative matrices.
Let $\mathbf{1}$ and $\mathbf{0}$ denote the all 1 and all 0 vectors, and let $\mathbf{O}$ denote the all 0 matrix.
$\dagger$ denotes the Moore-Penrose pseudo-inverse and ${\det}^\dagger$ denotes the pseudo-determinant. 
For matrix norms, ${\|\cdot\|}_F$ denotes the Frobenius norm, 
and ${\|\cdot\|}_{1}$ sum of the absolute values of all matrix entries.
We use ${\|\cdot\|}_2$ to denote vector 2-norm.

\section{Related Work}
\label{sec:related-work}

The problem of graph inference has been studied from three different perspectives: statistical modeling, GSP, and physically motivated models.

In \noindent \textbf{Probabilistic graphical models} (PGMs), a graph typically refers to the conditional dependencies between random variables.
One important class of PGM is the Gaussian Markov random field (GMRF) \cite{rue2005gaussian}. 
In a GMRF, the conditional dependencies are encoded by the inverse covariance matrix, also named the precision matrix, which plays a similar role as the graph Laplacian. 
The problem of learning the parsimonious precision matrix is known as covariance selection~\cite{dempster1972covariance}, assuming that the graph to be estimated is sparse.
Nowadays, covariance selection is usually solved by the prestigious graphical Lasso algorithm and its different variants \cite{friedman2008sparse,d2008first,lu2009smooth,scheinberg2010sparse,li2010inexact,hsieh2011sparse,witten2011new,mazumder2012graphical,oztoprak2012newton}.
However, it is worth mentioning that learning general GMRFs is intrinsically different from the graph Laplacian inference problem in this paper.
This is because a general precision matrix can have negative conditional dependencies but a graph in this context cannot contain negative edges.
Additionally, since Laplacians are first-order intrinsic, the set of Laplacians is disjoint from the set of general precision, which is nonsingular.
Other PGMs, such as different variants of GMRFs, discrete Markov random fields and Bayesian networks, also have found their applications in many fields, including but not limited to physics \cite{cipra1987introduction}, genetics \cite{yang2015graphical,park2017learning}, microbial ecology \cite{kurtz2015sparse,biswas2016learning,chiquet2019variational} and causal inference \cite{zhang1996exploiting,yu2004advances}.
But again, the learned matrices are not Laplacians.

The graph inference methods in \textbf{GSP} tackle the problem from a different perspective.
GSP typically assumes that the observations are outputs of a graph filtering system which takes some initial signals as inputs.
Inference methods vary on the assumptions of the form of graph filters, i.e. the function of GSO, and the characteristics of initial signals.
The smoothness of the signals with respect to the graph is arguably the most popular assumption \cite{dong2016learning,kalofolias2016learn,kalofolias2017large,egilmez2017graph,kumar2020unified,saboksayr2021accelerated,buciulea2022learning}.
Heat diffusion filters, also relying on the graph Laplacian, were studied in \cite{thanou2017learning}.
When the GSO is the adjacency matrix or the weighted adjacency matrix, (truncated) polynomial filters are widely used \cite{segarra2017network,navarro2020joint,shafipour2021identifying}. 
The normalized graph Laplacian matrix was also explored in \cite{pasdeloup2017characterization}, and its connection to the normalized adjacency matrix was also discussed.
Besides the graph filters, assumptions on the input or transformed signals can also vary.
Maretic, Thanou and Frossard~\cite{maretic2017graph} used a sparsity prior on the input signals instead of a standard Gaussian distribution. 
Recent methods learn graphs that endow the observations with sparse spectral representations~\cite{sardellitti2019graph,humbert2021learning}.
 
Interestingly, connections can be made between statistical models and GSP.
Although a general precision matrix is intrinsically different from a graph Laplacian, one can purposely make a GMRF improper to accommodate the Laplacian structure.
Such an exception is the class of attractive intrinsic GMRFs, where the precision matrices are improper and restricted to M-matrices so that they share the same properties as combinatorial graph Laplacians.
As we show in the next section, fitting this improper GMRF to the data \cite{slawski2015estimation} induces the same smoothness measurement term as learning a valid combinatorial graph Laplacian.

We describe our framework using the language of GSP but show that it also admits a PGM-like interpretation.
For example, when the noise distribution is selected as Poisson, our framework is closely related to the Poisson-Log-Normal (PLN) model. 
While graph inference from different data types such as Poisson was explored in the field of PGMs, e.g., Ising models and Poisson graphical models~\cite{yang2012graphical}, it is rarely discussed in the field of GSP where precision matrices are constrained to be combinatorial Laplacians.
This paper aims to complete this missing piece of rigorous graph Laplacian learning when noise varies across a broad spectrum of exponential family distributions.
Also, because typical PGMs such as PLN do not learn rigorous graph Laplacians, thus we only focus on GSP methods in this paper.

\noindent \textbf{Physically motivated models} consider the observations to be outcomes of some physical processes.
This is however less related to our setting and we refer the interested reader to the comprehensive surveys on graph inference in \cite{dong2019learning,mateos2019connecting}.

\section{Background}
\label{sec:background}

\subsection{Preliminaries}
\label{ssec:preliminary}

We begin with standard notations.
Consider a weighted undirected graph $G=\{\mathcal{V},\mathcal{E},W\}$ , where $\mathcal{V}$ denotes the set of $N$ vertices, $\mathcal{E}$ the set of edges and $\mathbf{W} \in \mathbf{S}\mathbb{R}_{+}^{N \times N}$ the weighted adjacency matrix.
Each entry $\mathbf{W}_{ij}=\mathbf{W}_{ji}\geq0$ corresponds to the weight of edge $\mathcal{E}_{ij}$.
Let $\mathbf{D}$ denote the diagonal $N \times N$ weighted degree matrix, such that the diagonal entry $\mathbf{D}_{ii}=\sum_j \mathbf{W}_{ij}$ is the degree of node $\mathcal{V}_i$.
The combinatorial Laplacian matrix $\mathbf{L}$ of graph $G$ is given by $\mathbf{L}=\mathbf{D}-\mathbf{W}$.

A real graph signal is a function $f:\mathcal{V} \rightarrow \mathbb{R}^N$ that assigns a real value to each vertex of the graph.
The definition of a smooth signal varies across different contexts, but generally a smooth graph signal $\mathbf{X}$ can be seen as the result of applying a low-pass graph filter $\mathcal{F}(\mathbf{L})$ to a non-smooth signal $\mathbf{x}_0$
\begin{equation}
\label{eq:smooth_signal_generation}
    \mathbf{x} = \boldsymbol{\mu}+ \mathcal{F}(\mathbf{L})\mathbf{x}_0 = \boldsymbol{\mu}+ \sum_i \mathbf{u}_i f(\lambda_i) \mathbf{u}_i^T \mathbf{x}_0,
\end{equation}
where $\boldsymbol{\mu}\in \mathbb{R}^{N}$ is an offset vector and applying the graph filter relies on $\{\mathbf{u}_i, \lambda_i\}$, the eigenvector-eigenvalue pairs of $\mathbf{L}$.
When $\mathbf{x}_0 \sim \mathcal{N}(\mathbf{0},\mathbf{I}_N)$, the smooth signal follows the distribution
\begin{equation}
\label{eq:smooth_signal_distribution}
    \mathbf{x} \sim \mathcal{N}(\boldsymbol{\mu},{\mathcal{F}(\mathbf{L})}^2).
\end{equation}
This GSP formulation of applying a low-pass filter to white $\mathbf{x}_0$, as well as different choices of the filter have been well-summarized by \cite{kalofolias2016learn}.

\subsection{Graph Inference from Smooth Signals}
\label{ssec:graph_inference_smooth}

Consider a dataset $\mathbf{X} \in \mathbb{R}^{N \times M}$, where each row corresponds to one of the $N$ nodes of the graph and each of the $M$ columns contains an independent smooth graph signal $\mathbf{X}_{\cdot j} \in \mathbb{R}^N$.
Suppose the graph $G$ does not change across columns, the goal of graph inference is to learn $G$ so that the columns of $\mathbf{X}$ will be smooth on the graph that we learn.
As the Laplacian matrix $\mathbf{L}$ is uniquely determined by the structure of the graph and the weights of $W$, this problem is equivalent to learning $\mathbf{L}$.

Methods of this type can be summarized by the following optimization problem:
\begin{equation}
\label{eq:general_smooth_objective}
    \min_{\mathbf{L} \in \mathcal{L}}  \left\{ \Tr(\mathbf{X}^T \mathbf{L X}) + \alpha h(\mathbf{L}) \right\},
\end{equation}
where $\mathcal{L}$ is the space of valid graph Laplacian matrices
\begin{equation}
    \mathcal{L} = \left\{ \mathbf{L} \in \mathbf{S}\mathbb{R}^{N \times N} \ | \ \mathbf{L}\mathbf{1}=\mathbf{0}, \mathbf{L}_{ij}=\mathbf{L}_{ji} \leq 0, \forall i \neq j  \right\},
\end{equation}
$h(\mathbf{L})$ is a regularization term, and $\alpha$ is a trade-off parameter.
The first term of \eqref{eq:general_smooth_objective} is known as the graph Laplacian quadratic form, which measures the overall smoothness of graph signals.
The smoothness term can be formulated in various ways, resulting in different optimization solutions for different methods.
The smoothness term naturally arises when we choose $\mathcal{F}(\mathbf{L})=\sqrt{\mathbf{L}^\dagger}$ in \eqref{eq:smooth_signal_generation}, where $\mathbf{L}^\dagger$ is the pseudo-inverse of $\mathbf{L}$. 
The second term imposes additional priors on the graph Laplacian, such as connectivity, sparsity, etc., whose specific choice varies across different methods. 

\looseness=-1
Here we briefly describe some popular graph learning methods with smooth signals. Both
\cite{hu2015spectral} and \cite{dong2016learning} use the original smoothness term and choose the regularization term $h(\mathbf{L})$ to be the Frobenius norm of $\mathbf{L}$ with an additional trace equality constraint.
They formulate the problem as a quadratic program with linear constraints and solve it through interior-point methods. 
\cite{kalofolias2016learn} reformulates the smoothness term as $\|\mathbf{W} \circ \mathbf{Z} \|=\frac{1}{2}\Tr(\mathbf{X}^T \mathbf{L X})$, where $\mathbf{Z} \in \mathbb{R}^{N \times N}$ is the squared Euclidean distance matrix across rows of $\mathbf{X}$ and $\circ$ denotes the Hadamard product.
They adapt the previous Frobenius norm regularization with a logarithmic barrier to promote connectivity, and solve for $\mathbf{W}$, instead of $\mathbf{L}$, with primal-dual optimization.
Finally, \cite{egilmez2017graph} use a different smoothness term and propose to solve
\begin{equation}
\label{eq:egilmez_smooth_objective}
    \min_{\mathbf{L} \in \mathcal{L}}  \left\{ \Tr(\mathbf{LS}) - \log{{\det}^\dagger(\mathbf{L})} + \alpha{\| \mathbf{L} \circ \mathbf{H} \|}_1 \right\}, \
\end{equation}
where $\mathbf{S}$ is a data statistic and the design of $\mathbf{H}$ imposes different regularization on $\mathbf{L}$.
The problem is solved using block coordinate descent on the rows and columns of $\mathbf{L}$.
The statistic $\mathbf{S}$ is commonly chosen to be the empirical covariance matrix, which makes the first term a scaled smoothness term since $\frac{1}{N-1}\Tr(\mathbf{LXX}^T)=\Tr{(\mathbf{LS})}$.

\subsection{Graph Inference with Gaussian Noise}
\label{ssec:gaussian_noise}
 
Given a dataset $\mathbf{X}$ of corrupted graph signals, 
~\cite{dong2016learning} modeled noisy observations as underlying smooth representations $\mathbf{y}$ with additive isotropic Gaussian noise $\boldsymbol\epsilon$.
Since smooth representations follow a Gaussian distribution, as shown in Eq.~\eqref{eq:smooth_signal_generation}, the noisy observations also follow a Gaussian distribution
\begin{equation}
\label{eq:noisy_gaussian_signal_distribution}
    \mathbf{x} \sim \mathcal{N}(\boldsymbol{\mu}, {\mathcal{F}(\mathbf{L})}^2 + \sigma_{\epsilon}^2 \mathbf{I}_N),
\end{equation}
where $\sigma_{\epsilon}$ is the standard deviation of Gaussian noise $\epsilon$ and $\mathcal{F}(\mathbf{L})=\sqrt{\mathbf{L}^\dagger}$.
When $\boldsymbol{\mu}=\mathbf{0}$, the MAP estimation of unobserved smooth representation $\mathbf{y}$ from \eqref{eq:noisy_gaussian_signal_distribution} amounts to
\begin{equation}
\label{eq:noisy_gaussian_map}
    \min_{\mathbf{y}} \left\{ {\| \mathbf{x}-\mathbf{y} \|}_2^2 + \beta \mathbf{y}^T\mathbf{Ly} \right\}, 
\end{equation}
where hyper-parameter $\beta$ reflects $\sigma_{\epsilon}^2$.
Based on \eqref{eq:noisy_gaussian_map}, \cite{dong2016learning} proposed to jointly learn $\mathbf{L}$ and the smooth signals $\mathbf{y}$: 
\begin{equation}
\label{eq:dong_full_objective}
    \min_{\mathbf{L} \in \mathcal{L}, \mathbf{Y}} \left\{ {\| \mathbf{X}-\mathbf{Y} \|}_F^2 + \beta (\Tr(\mathbf{Y}^T\mathbf{LY}) + \alpha h(\mathbf{L})) \right\},
\end{equation}
where $\mathbf{Y} \in \mathbb{R}^{N \times M}$ is the matrix of smooth representations, and $h(\mathbf{L})$ is the same Frobenius norm and trace constraint.
To solve Eq.~\eqref{eq:dong_full_objective}, they proposed an alternating algorithm that jointly learns the graph Laplacian and the smooth signal representations: at each iteration they fix one variable and solve for the other.
The advantage of this optimization is that each sub-problem objective is convex even though Eq.~\eqref{eq:dong_full_objective} is not.
When $\mathbf{y}$ is fixed ($\mathbf{y}$-step), the problem coincides with the Laplacian learning from smooth signals as in Eq.~\eqref{eq:general_smooth_objective}.
When $\mathbf{L}$ is fixed ($\mathbf{L}$-step), solving for $\mathbf{y}$
\begin{equation}
\label{eq:gaussian_y_step}
    \min_{\mathbf{Y}} \left\{ {\| \mathbf{X}-\mathbf{Y} \|}_F^2 + \beta (\Tr(\mathbf{Y}^T\mathbf{LY})) \right\}, 
\end{equation}
enjoys a closed-form solution $\mathbf{Y}={(\mathbf{I}_N + \beta \mathbf{L})}^{-1}\mathbf{X}$ that amounts to the Tikhonov filtering of $\mathbf{X}$.

\section{Graph Inference from Noisy Signals}
\label{sec:generalize_graph_inference}
\begin{figure*}
    \centering
    \includegraphics[width=\textwidth]{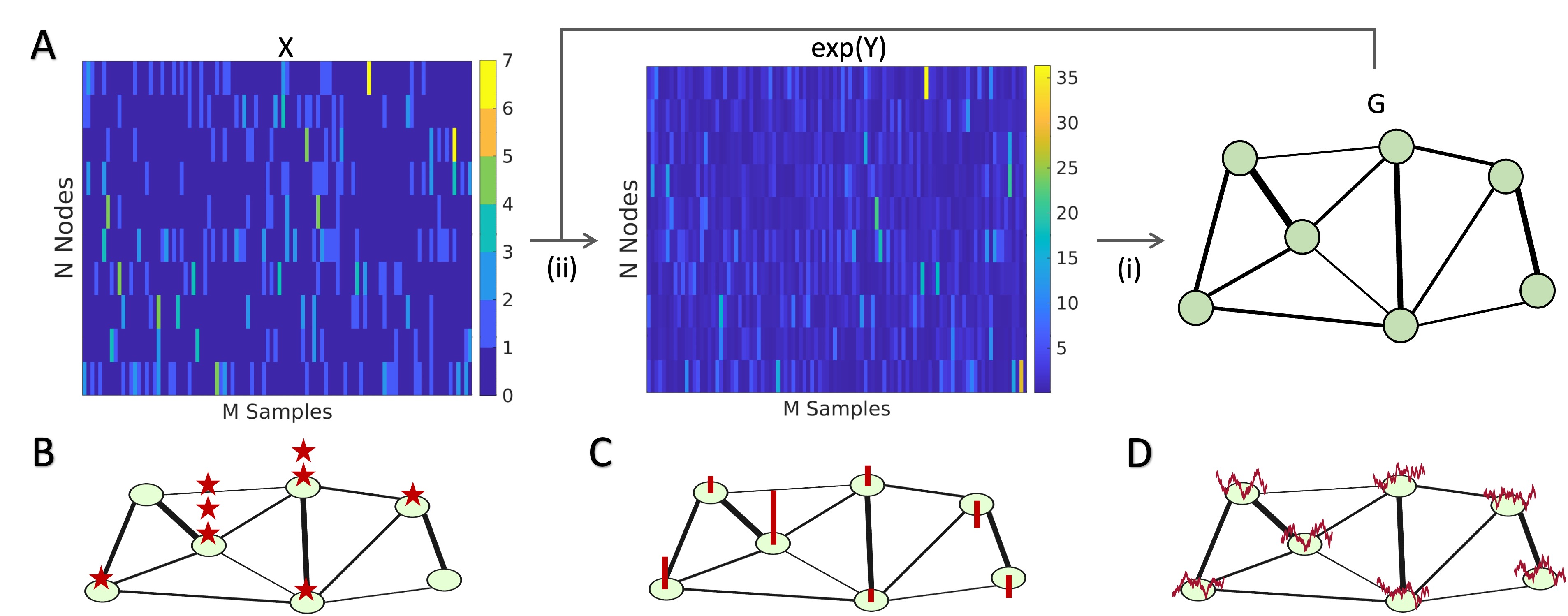}
    \caption{Graph Laplacian learning with Exponential Family Noise (GLEN).
    A. Illustration of GLEN. When only noisy signals $\mathbf{X}$ are available, GLEN (i) learns a graph from the estimation of unobserved smooth signals $\mathbf{Y}$ and (ii) denoises $\mathbf{X}$ using the newly obtained graph estimation to improve the quality of $\mathbf{Y}$ estimation and alternates these two steps.
    B. Example of a Poisson graph signal $\mathbf{x}$. Red star counts the events.
    C. Example of a smooth graph signal $\mathbf{y}$ underlying the noisy Poisson observations.
    D. Example of temporal correlated smooth graph signals.}
    \label{fig:glen}
\end{figure*}

In this section, we present a versatile generative framework that accounts for various types of graph signals with exponential family noise and propose an alternating algorithm for solving the unknown graph.
We elucidate two concrete examples of this generalized scenario with Poisson and Bernoulli noise.
We also enhance our method with variational inference to account for latent stochasticity. 
Finally, we present an adaptation to the time-vertex formulation where the data admits a joint graph-temporal structure, i.e. at every time point we observe a graph signal, and at each node of the graph we observe a time-series.
Our method is demonstrated in Fig.~\ref{fig:glen}.

\subsection{Graph Inference with Exponential Family Noise}
\label{ssec:exp_noise}

\looseness=-1
We first propose a GSP-based framework to model the generative process of noisy signals of different data types, motivated by the Gaussian case.
Specifically, the underlying smooth representation is generated from the same process as in \eqref{eq:smooth_signal_generation}, which is then connected to the mean parameter of the exponential family distribution through a link function $g$
\begin{equation}
\label{eq:noisy_exp_signal_expectation}
    \mathbb{E}[\mathbf{x}|\mathbf{y},\boldsymbol{\mu}] = g^{-1}(\mathbf{y}+\boldsymbol{\mu}),\ 
    s.t.\ \mathbf{y} \sim \mathcal{N}(\mathbf{0},{\mathcal{F}(\mathbf{L})}^2).
\end{equation}
More precisely, we consider exponential family distributions of the following form
\begin{equation}
\label{eq:exp_distribution}
    p(\mathbf{x}|\boldsymbol{\eta}) = k(\mathbf{x})\exp{\{\boldsymbol{\eta}^T T(\mathbf{x}) - A(\boldsymbol{\eta})\}},
\end{equation}
specified by natural parameters $\boldsymbol{\eta}$, sufficient statistics $T(\mathbf{x})$, and the cumulant generating function $A(\boldsymbol{\eta})$, which corresponds to the normalization factor of the probability distribution. 
Then the response model of noisy signals are given by
\begin{align}
\label{eq:noisy_exp_signal_generation}
    p(\mathbf{x}|\mathbf{y},\boldsymbol{\mu}) &= k(\mathbf{x})\exp{\{(\mathbf{y}+\boldsymbol{\mu}) T(\mathbf{x}) - A(\mathbf{y}+\boldsymbol{\mu})\}},\\ 
    &s.t.\ \mathbf{y} \sim \mathcal{N}(\boldsymbol{\mu},{\mathcal{F}(\mathbf{L})}^2). \nonumber
\end{align}
In reminiscence of generalized linear models (GLMs), we let smooth signals only control the mean parameters through the link function $g$ so that \eqref{eq:noisy_exp_signal_expectation} holds.
We list common exponential family distributions in Tab.~\ref{tab:exponential_family}.

\begin{table*}[h]
  \caption{Exponential family and link functions}
  \label{tab:exponential_family}
  \centering
  \begin{tabular}{cccccc}
    \toprule
    Distribution      & $\theta$        & $\eta$                & $T(\mathbf{x})$          & $A(\eta)$                 & g \\
    \midrule
    Bernouli          & $p$             & $\log \frac{p}{1-p}$  & x             & $\log{(1+e^{\eta})}$      & logit \\
    Binomial          & $p$             & $\log \frac{p}{1-p}$  & x             & $n\log{(1+e^{\eta})}$     & logit \\
    Negative Binomial & $p$             & $\log p$              & x             & $-r\log{(1-e^{\eta})}$    & logit \\
    Poisson           & $\lambda$       & $\log \lambda$        & x             & $e^{\eta}$                & log \\
    Gaussian          & $[\boldsymbol{\mu}, \sigma^2]$ & $[\nicefrac{\mu}{\sigma^2}, -\nicefrac{1}{2\sigma^2}]$ & $[x, x^2]$ & $\nicefrac{\mu^2}{2}$ & identity \\
    \bottomrule
  \end{tabular}
\end{table*}

When $p(\mathbf{x}|\mathbf{y},\boldsymbol{\mu})$ takes some specific exponential family distribution, $p(\mathbf{x})$ can find close relatives in the distribution zoo. 
For example, when the response is Poisson and the link function is logarithmic, $p(\mathbf{x})$ can be considered as an improper Poisson-Log-Normal (PLN) distribution \cite{aitchison1989multivariate} with the precision matrix constrained to be a combinatorial graph Laplacian
\begin{equation}
\label{eq:noisy_poisson_signal_distribution}
    \mathbf{x} \sim \mathcal{PLN}(\boldsymbol{\mu},{\mathcal{F}(\mathbf{L})}^2).
\end{equation}
Similarly, one can also obtain improper Bernoulli-Logit-Normal distributions or Binomial-Logit-Normal distributions from this framework, just to name a few.
Note that although these distributions have been studied, we focus on a particular case where the precision matrix is constrained to be a combinatorial graph Laplacian.
Such constraints permit a direction connection to GSP and have not been previously addressed in the literature.

Following \cite{dong2016learning}, the MAP estimate of $\mathbf{y}$ is 
\begin{equation}
    \min_{\mathbf{y}} \left\{-{(\mathbf{y}+\boldsymbol{\mu})}^T T(\mathbf{x}) + {\mathbf{1}}^T A(\mathbf{y}+\boldsymbol{\mu}) + \beta \mathbf{y}^T\mathbf{Ly}\right\}.
\end{equation}
Including the regularization terms and rewriting the objective in a matrix form, we obtain the full objective
\begin{align}
\label{eq:generalized_graph_learning}
    \min_{\mathbf{Y},\boldsymbol{\mu},\mathbf{L} \in \mathcal{L}} \Bigl\{ &- \Tr{((\mathbf{Y}^T+\mathbf{1}{\boldsymbol{\mu}}^T) T(\mathbf{X}))} +  \mathbf{1}^T A(\mathbf{Y}+\boldsymbol{\mu}{\mathbf{1}}^T) \mathbf{1}  \nonumber\\ 
    &+ \beta (\Tr{(\mathbf{Y}^T\mathbf{LY})} + \alpha h(\mathbf{L})) \Bigr\} \\     & \quad\quad\quad s.t.\ \mathbf{Y}^T \mathbf{1} = \mathbf{0}, \nonumber
\end{align}
where we slightly abuse the notation of $A$.
Note that we also add a constraint for $\mathbf{y}$ which we explain below.
The first two terms form a measure of the fidelity of the inferred smooth representations $\mathbf{y}$ with respect to noisy observation $\mathbf{X}$.
The last two terms are the same as \eqref{eq:general_smooth_objective}, imposing smoothness and other structural priors on the inferred graph.
When the distribution is isotropic Gaussian and $\boldsymbol{\mu}=\mathbf{0}$, \eqref{eq:generalized_graph_learning} coincides with \eqref{eq:dong_full_objective}.
If the regularization is chosen to be the same as \eqref{eq:dong_full_objective}, one can fully recover the method in \cite{dong2016learning} for learning the graph from Gaussian observations (Sec.~\ref{ssec:gaussian_noise}).

To solve \eqref{eq:generalized_graph_learning}, we propose GLEN, an alternating optimization algorithm, to learn $\mathbf{L}$, $\mathbf{Y}$ and $\boldsymbol{\mu}$ for general exponential family distributions, inspired by \cite{dong2016learning}. 
Similarly, although the original problem is not convex with respect to all variables, each sub-problem is convex with respect to a single variable.
Within each iteration, we update $\mathbf{L}$, $\mathbf{y}$ and $\boldsymbol{\mu}$ sequentially, during which the other two are fixed to their current estimations.

For the update of $\mathbf{L}$, since the choice of the exponential family distribution only affects the fidelity terms of \eqref{eq:generalized_graph_learning}, the $\mathbf{L}$-step is unaffected and is equivalent to the smooth signal learning scenario, where we simply replace $\mathbf{X}$ in \eqref{eq:general_smooth_objective} with $\mathbf{y}$
\begin{equation}
\label{eq:general_L_step}
    \min_{\mathbf{L} \in \mathcal{L}}  \left\{ \Tr(\mathbf{Y}^T\mathbf{LY}) + \alpha h(\mathbf{L}) \right\}.
\end{equation}
Depending on the choice of regularization $h(\mathbf{L})$, previous methods \cite{dong2016learning,kalofolias2016learn,egilmez2017graph} can be readily plugged in. In the simulations in Sec.~\ref{sec:experiments} we use the solution in~\cite{egilmez2017graph} for learning $\mathbf{L}$ for demonstration.

\looseness=-1
The update of $\mathbf{y}$ can be more challenging.
With Gaussian noise, we have shown an analytical solution exists for \eqref{eq:gaussian_y_step}, but this is not true with other exponential family distributions. 
More importantly, because $\mathbf{L}$ is a combinatorial graph Laplacian with first-order intrinsic property $\mathbf{L1}=\mathbf{0}$, \eqref{eq:generalized_graph_learning} is under-determined with infinite solutions.
Consider a set of minimizers $\widehat{\mathbf{Y}},\widehat{\boldsymbol{\mu}},\widehat{\mathbf{L}}$ and an arbitrary scaler $s$, $\widehat{\mathbf{Y}}+s\mathbf{1},\widehat{\boldsymbol{\mu}}-s\mathbf{1},\widehat{\mathbf{L}}$ is also a set of minimizers for \eqref{eq:generalized_graph_learning}.
Therefore, we want $\mathbf{y}$ to satisfy an additional constraint $\mathbf{Y}^T \mathbf{1} = \mathbf{0}$ to prevent floating $\mathbf{y}$ and $\boldsymbol{\mu}$.
Therefore, for the $\mathbf{y}$-step, we instead solve the following constrained optimization problem 
\begin{equation}
\label{eq:general_Y_step}
    \begin{gathered}
    \min_{\mathbf{Y}}  \Bigl\{ - \Tr{((\mathbf{Y}^T+\mathbf{1}{\boldsymbol{\mu}}^T) T(\mathbf{X}))}  \\  
    +\mathbf{1}^T A(\mathbf{Y}+\boldsymbol{\mu}{\mathbf{1}}^T) \mathbf{1}
    + \beta (\Tr{(\mathbf{Y}^T\mathbf{LY})} \Bigr\},\\
    s.t.\ \mathbf{Y}^T \mathbf{1} = \mathbf{0}.
    \end{gathered}
\end{equation}
To solve \eqref{eq:general_Y_step}, we apply the Newton-Raphson method with equality constraints to each smooth signal representation $\mathbf{y}_j=\mathbf{Y}_{\cdot j}$.
Let the gradient and Hessian of the unconstrained problem be $\nabla_j$ and $\nabla_j^2$, where
\begin{align}
    \nabla_j & = - \mathbf{x}_j + A'(\boldsymbol{\mu}+\mathbf{y}_j) + \beta \mathbf{L y}_j , \label{eq:y_update_grad}\\ 
    \nabla_j^2 & = \textrm{diag}(A''(\boldsymbol{\mu}+\mathbf{y}_j)) + \beta \mathbf{L} . \label{eq:y_update_hess}
\end{align}
The update $\Delta \mathbf{y}$ of the constrained problem is given by solution of $\mathbf{v}$ in the following linear system.
\begin{equation}
\label{eq:newton_with_equality}
    \begin{bmatrix} \nabla_j^2 & \mathbf{1} \\
     \mathbf{1}^T &  0 \end{bmatrix}
     \begin{bmatrix} \mathbf{v} \\
     w \end{bmatrix} =
     \begin{bmatrix} -\nabla_j \\
     0 \end{bmatrix}.
\end{equation}

Finally, $\boldsymbol{\mu}$ can be updated via a simple GLM fitting.
For each node $\mathcal{V}_i$, the update amounts to fitting a GLM (with the same exponential distribution) from invented predictors $\mathbf{1} \in \mathbb{R}^M$ to responses $\mathbf{X}_{i:}^T$ with known offset $\mathbf{Y}_{i:}^T$.
We summarize the full algorithm for general exponential family distributions in Alg.~\ref{alg:alternating_optimization} and present two specific examples with Poisson and Bernoulli distribution in the following sections.

\begin{algorithm}[tb]
   \caption{GLEN}
   \label{alg:alternating_optimization}
\begin{algorithmic}
   \STATE {\bfseries Input:} noisy signals $\mathbf{y}$
   \STATE {\bfseries Onput:} graph Laplacian $\mathbf{L}$, smooth signals $\mathbf{X}$, bias $\boldsymbol{\mu}$, stepsize $\rho$
   \REPEAT
   \STATE Initialize $\mathbf{y}$ and $\boldsymbol{\mu}$.
   \STATE Solve  $\mathbf{L}= \argmin_{\mathbf{L} \in \mathcal{L}}  \left\{ \Tr(\mathbf{Y}^T\mathbf{LY}) + \alpha h(\mathbf{L}) \right\}$
   \FOR{$j=1$ {\bfseries to} $M$}
   \STATE Calculate $\nabla_j$ and $\nabla_j^2$ as in \eqref{eq:y_update_grad} and \eqref{eq:y_update_hess}
   \STATE Solve $\mathbf{v}$ in \eqref{eq:newton_with_equality}
   \STATE Update $\mathbf{Y}_{:,j} \leftarrow \mathbf{Y}_{:,j} + \rho \mathbf{v}$
   \ENDFOR
   \FOR{$i=1$ {\bfseries to} $N$}
   \STATE Fit GLM from $\mathbf{1}$ to $\mathbf{X}_{i,:}^T$ with fixed offset $\mathbf{Y}_{i,:}^T$
   \ENDFOR
    \STATE Calculate the loss using \eqref{eq:generalized_graph_learning}
    \UNTIL convergence.
    
\end{algorithmic}
\end{algorithm}

\subsubsection{Derivation of Poisson Observations}
\label{sssec:glen_poisson}

We plug-in the Poisson distribution to our framework and derive the corresponding objective functions and the update rules.
Given Poisson distribution as shown in Table.~\ref{tab:exponential_family} and the generalized objective in Eq.~\eqref{eq:generalized_graph_learning}, we obtain the graph learning objective with Poisson noise:
\begin{equation}
\label{eq:poisson_graph_learning}
    \begin{gathered}
    \min_{\mathbf{Y},\boldsymbol{\mu},\mathbf{L} \in \mathcal{L}} \Bigl\{ - \Tr{((\mathbf{Y}^T+\mathbf{1}{\boldsymbol{\mu}}^T) \mathbf{X})} \\
    + \mathbf{1}^T \exp{(\mathbf{Y}+\boldsymbol{\mu}{\mathbf{1}}^T)} \mathbf{1}\\
    + \beta (\Tr{(\mathbf{Y}^T\mathbf{LY})} + \alpha h(\mathbf{L})) \Bigr\},\\
    s.t.\ \mathbf{Y}^T \mathbf{1} = \mathbf{0}.
    \end{gathered}
\end{equation}
The gradient and Hessian for unconstrained Newton–Raphson are then given by
\begin{align}
    \nabla_j & = - \mathbf{x}_j + \exp{(\boldsymbol{\mu}+\mathbf{y}_j)} + \beta \mathbf{L y}_j , \label{eq:y_update_grad_poisson}\\ 
    \nabla_j^2 & = \textrm{diag}(\exp{(\boldsymbol{\mu}+\mathbf{y}_j)}) + \beta \mathbf{L} , \label{eq:y_update_hess_poisson}
\end{align}
from which the true update $\Delta \mathbf{y}_j$ is obtained from Eq.~\eqref{eq:newton_with_equality}.

\subsubsection{Derivation of Bernoulli Observation}
\label{sssec:glen_bernoulli}

We now plug-in the Bernoulli distribution to our framework and derive the corresponding objective functions and the update rules.
Given Bernoulli distribution as shown in Table.~\ref{tab:exponential_family} and the generalized objective in Eq.~\eqref{eq:generalized_graph_learning}, we obtain the graph learning objective with Bernoulli noise
\begin{equation}
\label{eq:bernoulli_graph_learning}
    \begin{gathered}
    \min_{\substack{\mathbf{Y},\boldsymbol{\mu}\\\mathbf{L} \in \mathcal{L}}} \Bigl\{ - \Tr{((\mathbf{Y}^T+\mathbf{1}{\boldsymbol{\mu}}^T) \mathbf{X})} \\
    + \mathbf{1}^T \log(\mathbf{1}\mathbf{1}^T+\exp{(\mathbf{Y}+\boldsymbol{\mu}{\mathbf{1}}^T)}) \mathbf{1}\\
    + \beta (\Tr{(\mathbf{Y}^T\mathbf{LY})} + \alpha h(\mathbf{L})) ] \Bigr\}, \\
    s.t.\ \mathbf{Y}^T \mathbf{1} = \mathbf{0}.
    \end{gathered}
\end{equation}
The gradient and Hessian for unconstrained Newton–Raphson are then given by
\begin{align}
    & \nabla_j = - \mathbf{x}_j + \frac{1}{\mathbf{1}+\exp{(-\boldsymbol{\mu}-\mathbf{y}_j)}} + \beta \mathbf{L y}_j , \label{eq:y_update_grad_bernoulli}\\ 
    & \begin{gathered}
    \nabla_j^2 = \left. \textrm{diag}(\frac{1}{\exp{(\boldsymbol{\mu}+\mathbf{y}_j)}+\mathbf{2}+\exp{(-\boldsymbol{\mu}-\mathbf{y}_j)}}) \right.\\
    \left. + \beta \mathbf{L} \right. , \label{eq:y_update_hess_bernoulli}
    \end{gathered}
\end{align}
from which the true update $\Delta \mathbf{y}_j$ is obtained from Eq.~\eqref{eq:newton_with_equality}.

\subsection{Graph Inference with Stochastic Latent Variables}
\label{ssec:variational_exp_noise}
Taking the probabilistic view of graph inference, MLE is usually preferable due to its consistent performance and other attractive properties.
However, the likelihood is difficult to compute as it is an integral marginalized over latent smooth signal representations $\mathbf{Y}$.
One remedy is to use variational approaches to approximate the latent distributions of $\mathbf{Y}$ for a simplified likelihood function.
Since the MAP in Eq.~\ref{eq:generalized_graph_learning} only differs from MLE in the fact that it ignores the distributions of latent variables $\mathbf{Y}$, this results in a modified version of MAP.

We now consider the model where $\mathbf{y}$ is a stochastic latent variable.
Since the posterior of a smooth signal representation given its noisy observation $p(\mathbf{y}|\mathbf{x})$ is usually not tractable, we approximate that with an isotropic Gaussian distribution $q(\mathbf{y})$.
Denote the mean parameter of each Gaussian distribution $q(\mathbf{y}_j)$ with mean $\bar{\mathbf{Y}}_j$, and fix their covariance to be $\mathbf{\Lambda}=\lambda \mathbf{I}_N$.
We then derive the evidence lower bound (ELBO), and use its negation as a stochastic version of Eq.~\eqref{eq:generalized_graph_learning}:
\begin{equation}
    \begin{gathered}
        \min_{\substack{\bar{\mathbf{Y}},\boldsymbol{\mu}\\ \mathbf{L} \in \mathcal{L}} }  \Bigl\{  - \Tr{((\bar{\mathbf{Y}}^T+\mathbf{1}{\boldsymbol{\mu}}^T) T(\mathbf{X}))} \\
        + \mathbf{1}^T \mathbb{E}_{q(\mathbf{Y})} A(\mathbf{Y}+\boldsymbol{\mu}{\mathbf{1}}^T) \mathbf{1} \\
        + \beta (\Tr{(\bar{\mathbf{Y}}^T \mathbf{L} \bar{\mathbf{Y}})} + N \Tr(\mathbf{L\Lambda}) + \alpha h(\mathbf{L})) \Bigr\}, \\
        s.t.\ \bar{\mathbf{Y}}^T \mathbf{1} = \mathbf{0}.
    \end{gathered}
\end{equation}
The $\mathbf{L}$-step then becomes
\begin{equation}
\label{eq:general_vi_L_step}
    \min_{\mathbf{L} \in \mathcal{L}}  \left\{ \Tr(\bar{\mathbf{Y}}^T \mathbf{L} \bar{\mathbf{Y}}) + N \Tr(\mathbf{L\Lambda}) + \alpha h(\mathbf{L}) \right\}.
\end{equation}
This is similar to Eq.~\eqref{eq:egilmez_smooth_objective}, since $h(\mathbf{L})$ also contains $-\log{{\det}^\dagger(\mathbf{L})}$.
Using the same regularization, Eq.~\eqref{eq:general_vi_L_step} can be solved with \cite{egilmez2017graph} using the data statistic $\mathbf{S}=\mathbf{Y} \mathbf{Y}^T+N\mathbf{\Lambda}$.
The $\mathbf{y}$-step now involves the expectation of $A(\mathbf{Y}+\boldsymbol{\mu}\mathbf{1}^T)$ which reflects the stochasticity of $\mathbf{Y}$
\begin{equation}
\label{eq:general_vi_Y_step}
    \begin{gathered}
    \min_{\mathbf{y}}  \Bigl\{ - \Tr{((\bar{\mathbf{Y}}^T+\mathbf{1}{\boldsymbol{\mu}}^T) T(\mathbf{X}))}  +\\  
    \mathbf{1}^T \mathbb{E}_{q(\mathbf{Y})} A(\mathbf{Y}+\boldsymbol{\mu}{\mathbf{1}}^T) \mathbf{1}
    + \beta (\Tr{(\bar{\mathbf{Y}}^T L \bar{\mathbf{Y}})} \Bigr\},\\
    s.t.\ \bar{\mathbf{Y}}^T \mathbf{1} = \mathbf{0}.
    \end{gathered}
\end{equation}
As we choose $q(\mathbf{Y})$ to be Gaussian, some exponential family distributions will enjoy a closed-form $\mathbb{E}_{q(\mathbf{Y})} A(\mathbf{Y}+\boldsymbol{\mu}\mathbf{1}^T)$, while others need to seek approximations.
The update of $\boldsymbol{\mu}$ remains the same.
We term this variational based approach as GLEN-VI.
See details and the full derivation in Appendix~\ref{sec:exp_noise_vi}.

\subsection{Graph Inference with Temporal Correlation}
\label{ssec:time_exp_noise}

In the previous section, it is assumed that each column of the input matrix $\mathbf{X}$ is an independent graph signal sampled from the 2-layer generative model.
This is not always true in practice.
A common case is that the columns are correlated, for example, when the data has a natural temporal ordering.
In such a scenario, each row of the matrix is a canonical 1-D signal on the nodes, and the whole matrix can then be seen as a multivariate time-varying signal lying on the graph.
For example, in neural data analysis, the input is often a count matrix whose rows correspond to neurons and columns correspond to time bins.
Then we can model the neurons (rows) as nodes of a graph reflecting their (functional) connectivity, while the columns that are close in time should also exhibit similar firing patterns.
Such an assumption of temporal correlation merges classical signal processing with GSP and proves to be useful in many scenarios.

The Time-Vertex signal processing framework \cite{grassi2017time,liu2019graph} is specifically designed for this kind of setting.
For a time-varying graph signal $\mathbf{X} \in \mathbb{R}^{N \times M}$, the time-vertex framework considers a joint graph $J$ which is the Cartesian product of a graph $G$ that underlies the $N$ rows and a temporal ring graph $T$ that underlies the $M$ columns.
A smooth signal on this product graph is not only smooth upon $G$, but also enjoys temporal smoothness upon $T$.
The joint time-vertex Fourier transform (JFT) is simply applying the graph Fourier transform (GFT) on the $G$ dimension and the discrete Fourier transform (DFT) on the $T$ dimension.
It can further be shown that the graph Laplacian quadratic form of a matrix $\mathbf{X}$ on $J$ is simply the summation of the quadratic forms along each dimension
\begin{equation}
\label{eq:time_vertex_smoothness}
    \Tr{(\mathbf{x}^T \mathbf{L}_J \mathbf{x})} = \Tr{(\mathbf{X}^T \mathbf{L}_G \mathbf{X})} + \Tr{(\mathbf{X L}_T \mathbf{X}^T)}
\end{equation}
where $\mathbf{X}$ is the column-wise vectorization of $\mathbf{X}$, and $\mathbf{L}_J$, $\mathbf{L}_G$, $\mathbf{L}_T$ denote the Laplacian of $J$, $G$, $T$.

Here we assume that the underlying representation of the input matrix is smooth on the time-vertex graph $J$.
As the temporal graph $T$ is known a priori (modeled as a path graph), learning $G$ is essentially the same as learning $J$.
The optimization problem becomes the following:
\begin{equation}
\label{eq:generalized_tvgraph_learning}
\hspace*{-0.15cm}
    \begin{gathered}
    \min_{\substack{\mathbf{Y},\boldsymbol{\mu}, \\\mathbf{L} \in \mathcal{L}}} \Bigl\{ - \Tr{((\mathbf{Y}^T+\mathbf{1}{\boldsymbol{\mu}}^T) T(\mathbf{X}))} + \mathbf{1}^T A(\mathbf{Y}+\boldsymbol{\mu}{\mathbf{1}}^T) \mathbf{1}  \\  + \gamma \Tr{(\mathbf{YTY}^T)} 
    + \beta (\Tr{(\mathbf{Y}^T\mathbf{LY})} + \alpha h(\mathbf{L})) \Bigr\}\\
    s.t.\ \mathbf{Y}^T \mathbf{1} = \mathbf{0}, 
    \end{gathered}
\end{equation}
where $\gamma$ controls the weight of temporal edges in $T$.
Similar alternating optimization algorithm can be applied for \eqref{eq:generalized_tvgraph_learning}, with the only change being the $\mathbf{y}$-step.
Again the new gradient and Hessian of the unconstrained problem are
\begin{align}
    & \begin{gathered}
        \nabla_j = \left. - \mathbf{x}_j + A'(\boldsymbol{\mu}+\mathbf{y}_j) + \beta \mathbf{L y}_j + \right. \\ \left.
        \ 2 \gamma (2 \mathbf{y}_j - \mathbf{y}_{j-1} - \mathbf{y}_{j+1}) \right.,
    \end{gathered}\\
    & \nabla_j^2 = \textrm{diag}(A''(\boldsymbol{\mu}+\mathbf{y}_j)) + \beta \mathbf{L} + 4 \gamma \mathbf{I}_N, 
\end{align}
and the update $\Delta \mathbf{y}$ of the constrained problem is given by Eq.~\eqref{eq:newton_with_equality}. We term our solution GLEN-TV.

\section{Experiments}
\label{sec:experiments}

\begin{table*}[ht]
\setlength\tabcolsep{3pt}
    \caption{Numerical comparison of structure prediction (N=20).}
    \label{tab:structure_prediction}
    \centering
    \small
    \begin{tabular}{|c||c|c|c|c|c|c|c|c|c|}
        \hline
        \multirow{2}{*}{Method}
        & \multicolumn{3}{c|}{Erd\H{o}s-R\'enyi} & \multicolumn{3}{c|}{Stochastic Block} & \multicolumn{3}{c|}{Watts-Strogatz}\\
        \cline{2-10}
         & F-score $\uparrow$ & PR-AUC $\uparrow$ & NMI $\uparrow$ & F-score $\uparrow$ & PR-AUC $\uparrow$ & NMI $\uparrow$ & F-score $\uparrow$ & PR-AUC $\uparrow$ & NMI $\uparrow$ \\
         \hline
         SCGL \cite{lake2010discovering} & $0.64\pm0.03$ & $0.51\pm0.04$ & $0.16\pm0.03$ & $0.68\pm0.03$ & $0.60\pm0.06$ & $0.26\pm0.05$ & $0.67\pm0.03$ & $0.64\pm0.04$ & $0.28\pm0.05$ \\
         GLS-1 \cite{dong2016learning} & $0.65\pm0.03$ & $0.68\pm0.04$ & $0.18\pm0.04$ & $0.67\pm0.04$ & $0.70\pm0.04$ & $0.26\pm0.06$ & $0.75\pm0.06$ & $0.79\pm0.04$ & $0.41\pm0.09$\\
         GLS-2 \cite{kalofolias2016learn} & $0.62\pm0.03$ & $0.53\pm0.04$ & $0.14\pm0.03$ & $0.64\pm0.04$ & $0.61\pm0.07$ & $0.21\pm0.04$ & $0.65\pm0.04$ & $0.69\pm0.05$ & $0.27\pm0.06$ \\
         CGL \cite{egilmez2017graph} & $\textcolor{red}{0.78\pm0.03}$ & $\textcolor{red}{0.82\pm0.03}$ & $\textcolor{red}{0.40\pm0.07}$ & $0.83\pm0.04$ & $0.82\pm0.05$ & \textbf{$0.51\pm0.08$} & $0.90\pm0.04$ & $0.91\pm0.03$ & $0.64\pm0.08$\\
         GLEN & $0.76\pm0.04$ & $0.79\pm0.04$	& $0.34\pm0.06$	& $\textcolor{red}{0.82\pm0.04}$ & \textcolor{violet}{$0.84\pm0.04$} & $\textcolor{red}{0.49\pm0.08}$ & \textcolor{violet}{$0.92\pm0.02$} & $\textcolor{red}{0.93\pm0.02}$ & \textcolor{violet}{$0.72\pm0.06$}\\
         GLEN-VI & \textcolor{violet}{$0.77\pm0.04$}	& \textcolor{violet}{$0.81\pm0.04$}	& \textcolor{violet}{$0.36\pm0.07$} & $\textcolor{red}{0.82\pm0.04}$ & $\textcolor{red}{0.86\pm0.04}$ & $\textcolor{red}{0.49\pm0.08}$ & $\textcolor{red}{0.93\pm0.03}$ & $\textcolor{red}{0.93\pm0.02}$ & $\textcolor{red}{0.73\pm0.07}$\\
         \hline
    \end{tabular}
\end{table*}

\begin{table*}[ht]
\setlength\tabcolsep{3pt}
    \caption{Numerical comparison of weight prediction (N=20).}
    \label{tab:weight_prediction}
    \centering
    \small
    \begin{tabular}{|c||c|c|c|c|c|c|c|c|c|}
        \hline
        \multirow{2}{*}{Method}
        & \multicolumn{3}{c|}{Erd\H{o}s-R\'enyi} & \multicolumn{3}{c|}{Stochastic Block} & \multicolumn{3}{c|}{Watts-Strogatz}\\
        \cline{2-10}
         & $\textrm{RE}_L$ $\downarrow$ & $\textrm{RE}_{edge}$ $\downarrow$ & $\textrm{RE}_{deg}$ $\downarrow$ & $\textrm{RE}_L$ $\downarrow$ & $\textrm{RE}_{edge}$ $\downarrow$ & $\textrm{RE}_{deg}$ $\downarrow$ & $\textrm{RE}_L$ $\downarrow$ & $\textrm{RE}_{edge}$ $\downarrow$ & $\textrm{RE}_{deg}$ $\downarrow$ \\
         \hline
         SCGL \cite{lake2010discovering} & $0.89\pm0.07$ & $1.06\pm0.06$ & $0.85\pm0.07$ & $0.83\pm0.10$ & $0.89\pm0.06$ & $0.82\pm0.11$ & $0.80\pm0.07$ & $0.81\pm0.06$ & $0.80\pm0.07$\\
         GLS-1 \cite{dong2016learning} & $0.44\pm0.03$ & $0.80\pm0.01$ & $0.32\pm0.05$ & $0.48\pm0.03$ & $0.78\pm0.02$ & $0.37\pm0.04$ & $0.44\pm0.03$ & $0.73\pm0.03$ & $0.31\pm0.05$\\
         GLS-2 \cite{kalofolias2016learn} & $0.45\pm0.03$ & $0.85\pm0.01$ & $0.32\pm0.05$ & $0.51\pm0.03$ & $0.87\pm0.01$ & $0.37\pm0.04$ & $0.49\pm0.02$ & $0.89\pm0.01$ & $0.29\pm0.05$\\
         CGL \cite{egilmez2017graph} & $0.43\pm0.08$ & $0.68\pm0.13$ & $0.36\pm0.08$ & $0.46\pm0.06$ & $0.65\pm0.09$ & $0.40\pm0.06$ & $0.41\pm0.08$ & $0.52\pm0.07$ & $0.37\pm0.08$\\
         GLEN & $\textcolor{red}{0.32\pm0.03}$	& $\textcolor{red}{0.56\pm0.05}$	& $\textcolor{red}{0.24\pm0.04}$ & $\textcolor{red}{0.32\pm0.03}$ & \textcolor{violet}{$0.55\pm0.09$} & $\textcolor{red}{0.24\pm0.03}$ & $\textcolor{red}{0.25\pm0.04}$ & $\textcolor{red}{0.38\pm0.04}$ & $\textcolor{red}{0.19\pm0.04}$\\
         GLEN-VI & \textcolor{violet}{$0.34\pm0.03$}	& \textcolor{violet}{$0.61\pm0.03$}	& \textcolor{violet}{$0.25\pm0.04$} & \textcolor{violet}{$0.34\pm0.03$} & $\textcolor{red}{0.53\pm0.05}$ & \textcolor{violet}{$0.28\pm0.03$} & $\textcolor{red}{0.25\pm0.04}$ & $\textcolor{red}{0.38\pm0.04}$ & \textcolor{violet}{$0.19\pm0.05$}\\
         \hline
    \end{tabular}
\end{table*}

\begin{table*}[ht]
\setlength\tabcolsep{3pt}
    \caption{Numerical comparison of structure prediction (N=50).}
    \label{tab:structure_prediction_n50}
    \centering
    \small
    \begin{tabular}{|c||c|c|c|c|c|c|c|c|c|}
        \hline
        \multirow{2}{*}{Method}
        & \multicolumn{3}{c|}{Erd\H{o}s-R\'enyi} & \multicolumn{3}{c|}{Stochastic Block} & \multicolumn{3}{c|}{Watts-Strogatz}\\
        \cline{2-10}
         & F-score $\uparrow$ & PR-AUC $\uparrow$ & NMI $\uparrow$ & F-score $\uparrow$ & PR-AUC $\uparrow$ & NMI $\uparrow$ & F-score $\uparrow$ & PR-AUC $\uparrow$ & NMI $\uparrow$ \\
         \hline
         SCGL \cite{lake2010discovering} & $0.55\pm0.01$ & $0.42\pm0.02$ & $0.08\pm0.01$ & $0.61\pm0.02$ & $0.52\pm0.02$ & $0.17\pm0.02$ & $0.72\pm0.05$ & $0.78\pm0.05$ & $0.46\pm0.07$ \\
         GLS-1 \cite{dong2016learning} & $0.57\pm0.01$ & $0.54\pm0.02$ & $0.09\pm0.01$ & $0.61\pm0.02$ & $0.63\pm0.02$ & $0.12\pm0.02$ & $0.80\pm0.04$ & $0.80\pm0.04$ & $0.56\pm0.06$\\
         GLS-2 \cite{kalofolias2016learn} & $0.54\pm0.01$ & $0.42\pm0.02$ & $0.07\pm0.01$ & $0.56\pm0.02$ & $0.52\pm0.02$ & $0.21\pm0.04$ & $0.77\pm0.04$ & $0.79\pm0.05$ & $0.54\pm0.05$ \\
         CGL \cite{egilmez2017graph} & $0.47\pm0.02$ & $\textcolor{red}{0.66\pm0.02}$ & $\textcolor{red}{0.14\pm0.02}$ & $0.56\pm0.02$ & $0.66\pm0.01$ & $0.20\pm0.03$ & $0.82\pm0.03$ & $0.86\pm0.03$ & $0.59\pm0.04$\\
         GLEN & \textcolor{violet}{$0.57\pm0.02$}	& $0.61\pm0.02$	& $0.10\pm0.02$ & \textcolor{violet}{$0.63\pm0.02$} & \textcolor{violet}{$0.69\pm0.03$} & \textcolor{violet}{$0.22\pm0.03$} & $\textcolor{red}{0.86\pm0.07}$ & \textcolor{violet}{$0.86\pm0.10$}  & $\textcolor{red}{0.66\pm0.11}$\\
         GLEN-VI & $\textcolor{red}{0.60\pm0.02}$ & \textcolor{violet}{$0.62\pm0.03$}	& \textcolor{violet}{$0.13\pm0.02$}	& $\textcolor{red}{0.65\pm0.02}$ & $\textcolor{red}{0.71\pm0.02}$ & $\textcolor{red}{0.26\pm0.02}$ & \textcolor{violet}{$0.85\pm0.04$} & $\textcolor{red}{0.90\pm0.05}$& \textcolor{violet}{$0.64\pm0.06$}\\
         \hline
    \end{tabular}
\end{table*}

\begin{table*}[ht]
\setlength\tabcolsep{3pt}
    \caption{Numerical comparison of weight prediction (N=50).}
    \label{tab:weight_prediction_n50}
    \centering
    \small
    \begin{tabular}{|c||c|c|c|c|c|c|c|c|c|}
        \hline
        \multirow{2}{*}{Method}
        & \multicolumn{3}{c|}{Erd\H{o}s-R\'enyi} & \multicolumn{3}{c|}{Stochastic Block} & \multicolumn{3}{c|}{Watts-Strogatz}\\
        \cline{2-10}
         & $\textrm{RE}_L$ $\downarrow$ & $\textrm{RE}_{edge}$ $\downarrow$ & $\textrm{RE}_{deg}$ $\downarrow$ & $\textrm{RE}_L$ $\downarrow$ & $\textrm{RE}_{edge}$ $\downarrow$ & $\textrm{RE}_{deg}$ $\downarrow$ & $\textrm{RE}_L$ $\downarrow$ & $\textrm{RE}_{edge}$ $\downarrow$ & $\textrm{RE}_{deg}$ $\downarrow$ \\
         \hline
         SCGL \cite{lake2010discovering} & $0.93\pm0.03$ & $1.20\pm0.02$ & $0.90\pm0.04$ & $0.91\pm0.03$ & $1.03\pm0.02$ & $0.90\pm0.04$ & $0.61\pm0.05$ & $0.64\pm0.04$ & $0.60\pm0.05$\\
         GLS-1 \cite{dong2016learning} & $0.33\pm0.02$ & $0.84\pm0.01$ & $0.24\pm0.03$ & $0.36\pm0.02$ & $0.85\pm0.01$ & $0.26\pm0.03$ & $0.46\pm0.02$ & $0.78\pm0.02$ & $0.30\pm0.03$\\
         GLS-2 \cite{kalofolias2016learn} & $0.33\pm0.02$ & $0.87\pm0.01$ & $0.24\pm0.03$ & $0.36\pm0.02$ & $0.89\pm0.00$ & $0.26\pm0.03$ & $0.52\pm0.01$ & $0.96\pm0.00$ & $0.28\pm0.03$\\
         CGL \cite{egilmez2017graph} & $0.42\pm0.03$ & $1.03\pm0.06$ & $0.32\pm0.04$ & $0.42\pm0.03$ & $0.95\pm0.04$ & $0.33\pm0.04$ & $0.42\pm0.04$ & $0.54\pm0.03$ & $0.38\pm0.05$\\
         GLEN & $\textcolor{red}{0.28\pm0.01}$ & \textcolor{violet}{$0.79\pm0.03$}	& $\textcolor{red}{0.18\pm0.02}$ & $\textcolor{red}{0.28\pm0.02}$ & \textcolor{violet}{$0.72\pm0.02$} & $\textcolor{red}{0.19\pm0.03}$ & \textcolor{violet}{$0.34\pm0.12$} & \textcolor{violet}{$0.53\pm0.17$} & \textcolor{violet}{$0.26\pm0.10$}\\
         GLEN-VI & \textcolor{violet}{$0.29\pm0.02$}	& $\textcolor{red}{0.77\pm0.01}$	& \textcolor{violet}{$0.21\pm0.02$} & \textcolor{violet}{$0.30\pm0.02$} & $\textcolor{red}{0.70\pm0.01}$ & \textcolor{violet}{$0.22\pm0.03$} & $\textcolor{red}{0.32\pm0.04}$ & $\textcolor{red}{0.50\pm0.07}$ & $\textcolor{red}{0.24\pm0.03}$\\
         \hline
    \end{tabular}
\end{table*}

We evaluate GLEN on synthetic graphs as well as multiple real datasets.
All experiments are done with MATLAB.

\subsection{Synthetic Graphs}
\looseness=-1
We consider three random graph models: 
1) Erd\H{o}s-R\'enyi model with parameter $p=0.3$;
2) Stochastic block model with two equal-sized blocks, intra-community parameter $p=0.4$ and inter-community parameter $q=0.1$; and 3) Watts-Strogatz small-world model, where we create an initial ring lattice with node degree $2K=4$ and rewire every edge of the graph with probability $p=0.1$.
Using each model, we generate 20 random graphs of $N=20$ or $N=50$ nodes, respectively.
The weight matrix $\mathbf{W}$ of each graph is randomly sampled from a uniform distribution $\mathcal{U}(0.1,2)$, and the unnormalized Laplacian is computed as $\mathbf{L}_{u}=\mathbf{D}-\mathbf{W}$.
We then normalize it as $\mathbf{L}_0=N\frac{\mathbf{L}_u}{\Tr{(\mathbf{L}_u)}}$ to obtain the ground-truth Laplacian.
To generate count signals, we set the offset to:
\[
\boldsymbol{\mu}_i = \begin{cases}
2 & 1 \leq i \leq \lfloor\frac{N}{2}\rfloor \\
-2 & \lfloor\frac{N}{2}\rfloor < i \leq N.
\end{cases}
\]
To demonstrate the efficacy of GLEN, we simulate $M=2000$ count signals with Poisson distribution following \eqref{eq:noisy_exp_signal_generation}, and infer the synthetic graphs from these synthetic signals.
We use the the GSPBOX \cite{perraudin2014gspbox} and the SWGT toolbox \cite{hammond2011wavelets} for graph generation and signal processing.

\looseness=-1
The baselines are selected to be leading graph Laplacian learning methods: shifted combinatorial graph Laplacian learning (SCGL) \cite{lake2010discovering}, two methods for graph learning from smooth signals: GLS-1~\cite{dong2016learning} and GLS-2~\cite{kalofolias2016learn}, and combinatorial graph Laplacian learning (CGL) \cite{egilmez2017graph}.
Because existing methods are designed for Gaussian distributions, which are rather different from Poisson distribution statistically, preprocessing is required for to accommodate the model mismatch.
For SCGL, GLS-1 and GLS-2, we first log-transform and then centralize the count matrices: $\log(\mathbf{X}+\mathbf{1})-\frac{1}{N}\log(\mathbf{X}+\mathbf{1})^T\mathbf{1}$.
For the baseline CGL, we select the data statistic matrix $S$ as the empirical covariance of $\log(\mathbf{X}+\mathbf{1})$.
As the experiments will show, it is not a good estimation of $\mathbf{\Sigma}$.
For all the CGL experiments in this paper, we use $\mathbf{H}=2\mathbf{I}-\mathbf{11}^T$ to regularize ${\|\mathbf{L}\|}_1$, corresponding to type-1 regularization in \cite{egilmez2017graph}.

\vspace{0.1cm}
We quantitatively compare our proposed methods with the baselines in terms of structure prediction and weight prediction.
For structure prediction, we report area under the precision-recall curve (PR-AUC), F-score and normalized mutual information (NMI). 
These three metrics evaluate the binary prediction of edge existence given by the inferred Laplacian $\mathbf{L}^*$, but ignore the graph weights. 
We use a series of thresholds to obtain the precision-recall curve and calculate the F-score and NMI using the best threshold on the curve (in terms of F-score).
For weight prediction, we report the relative error (RE) of estimated Laplacians, edges, and degrees against the ground-truth, which reflect both structure and weight prediction.
Following \cite{egilmez2017graph}, we first normalize the inferred Laplacian $\mathbf{L}^*$ to obtain $\widehat{\mathbf{L}}=\frac{\mathbf{L}^*}{\Tr{(\mathbf{L}^*)}}N$.
The relative error of Laplacian, is then computed as $\textrm{RE}_\mathbf{L}=\frac{{\|\widehat{\mathbf{L}}-\mathbf{L}_0\|}_F}{{\|\mathbf{L}_0\|}_F}$.
For relative error of edges, we vectorize the upper triangle of $\mathbf{W}$ to obtain $\textrm{vech}(\mathbf{L}) \in \mathbb{R}^{N*(N-1)/2}$, and compute the relative $\ell_2$ norm $\textrm{RE}_{edge}=\frac{{\|\textrm{vech}(\widehat{\mathbf{L}})-\textrm{vech}(\mathbf{L}_0)\|}_2}{{\|\textrm{vech}(\mathbf{L}_0)\|}_2}$.
For the relative error of degrees, we compute the relative $\ell2$ norm of the degrees $\textrm{RE}_{deg}=\frac{{\|\textrm{diag}(\widehat{\mathbf{L}})-\textrm{diag}(\mathbf{L}_0)\|}_2}{{\|\textrm{diag}(\mathbf{L}_0)\|}_2}$, where $\textrm{diag}(\mathbf{L})$ is the vector of diagonal elements of the Laplacian.

For the hyperparameters selection, we perform a grid search on the parameter space for each method.
For the comparison of structure prediction, we perform a grid search on each method's hyperparameters, and report the performance of the hyperparameter setting that achieved the highest average PR-AUC across 20 random graphs.
For the comparison of weight prediction, we report the performance of the hyper-parameter setting with lowest $\textrm{RE}_L$ among them.
Our evaluation ensures that we can compare different methods by their best weight prediction with reasonably good structure prediction.

The quantitative comparison of structure and weight prediction for each graph model with $N=20$ is shown in Table.~\ref{tab:structure_prediction} and \ref{tab:weight_prediction}.
Similarly, results of $N=50$ are shown in Table.~\ref{tab:structure_prediction_n50} and \ref{tab:weight_prediction_n50}.
We color the \textcolor{red}{best} performance with red and the \textcolor{violet}{second best} performance with violet for each metric.
As we can see, our methods outperform the baselines on almost all structure prediction metrics and all weight prediction metrics.
For structure prediction, both our methods GLEN and GLEN-VI are competitive on Erd\H{o}s-R\'enyi graphs and substantially superior on stochastic block models and Watt-Strogatz small-world graphs.
For weight prediction, our methods, especially GLEN-VI, significantly outperform the baselines on all type of graphs.

\begin{figure*}[hb]
    \centering
    \begin{tabular}{cccccccc}
    SCGL    & GLS-1 & GLS-2 & CGL   & GLEN  &GLEN-VI    & Ground Truth \\
    \includegraphics[width=2cm]{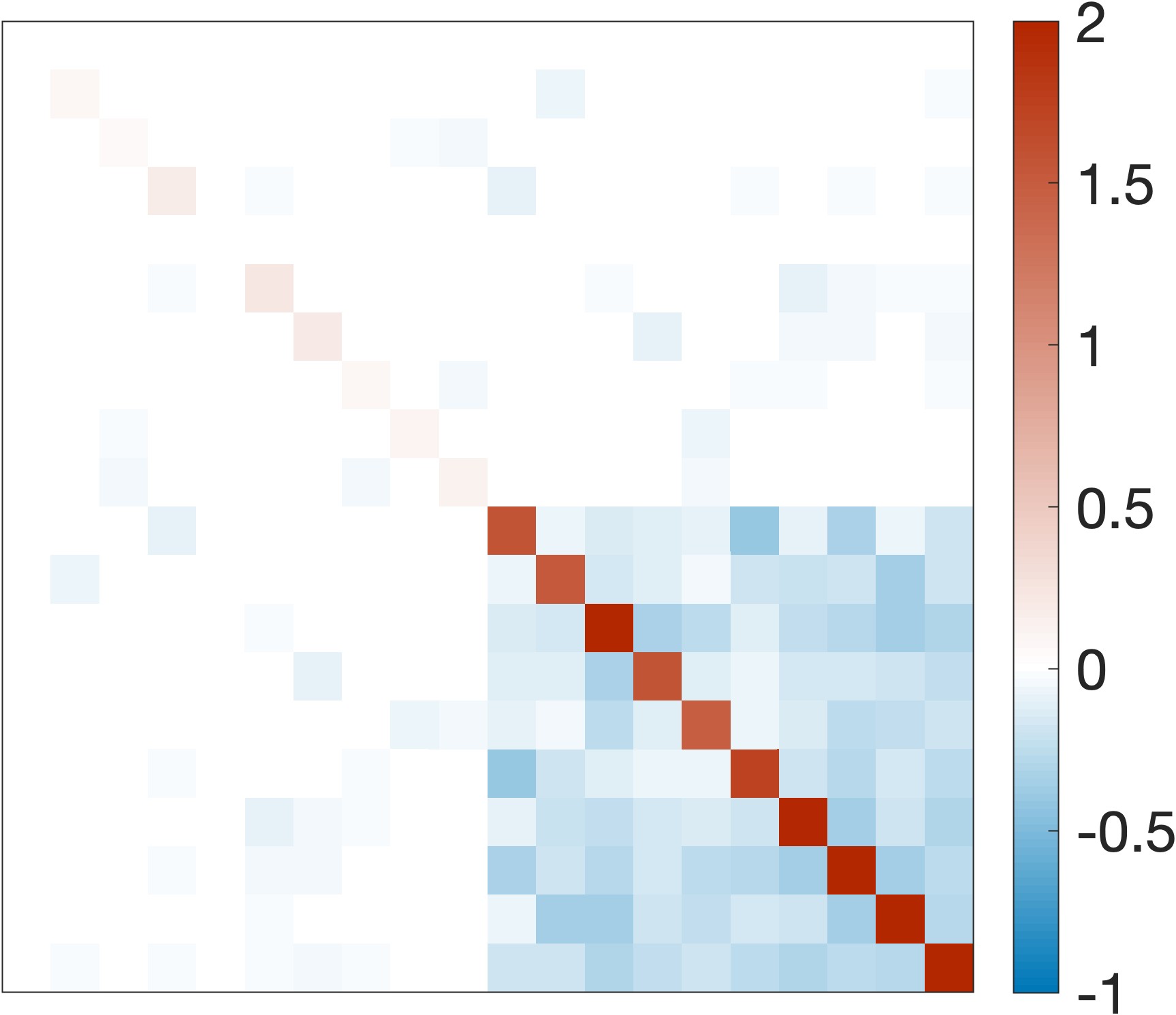}     &  \includegraphics[width=2cm]{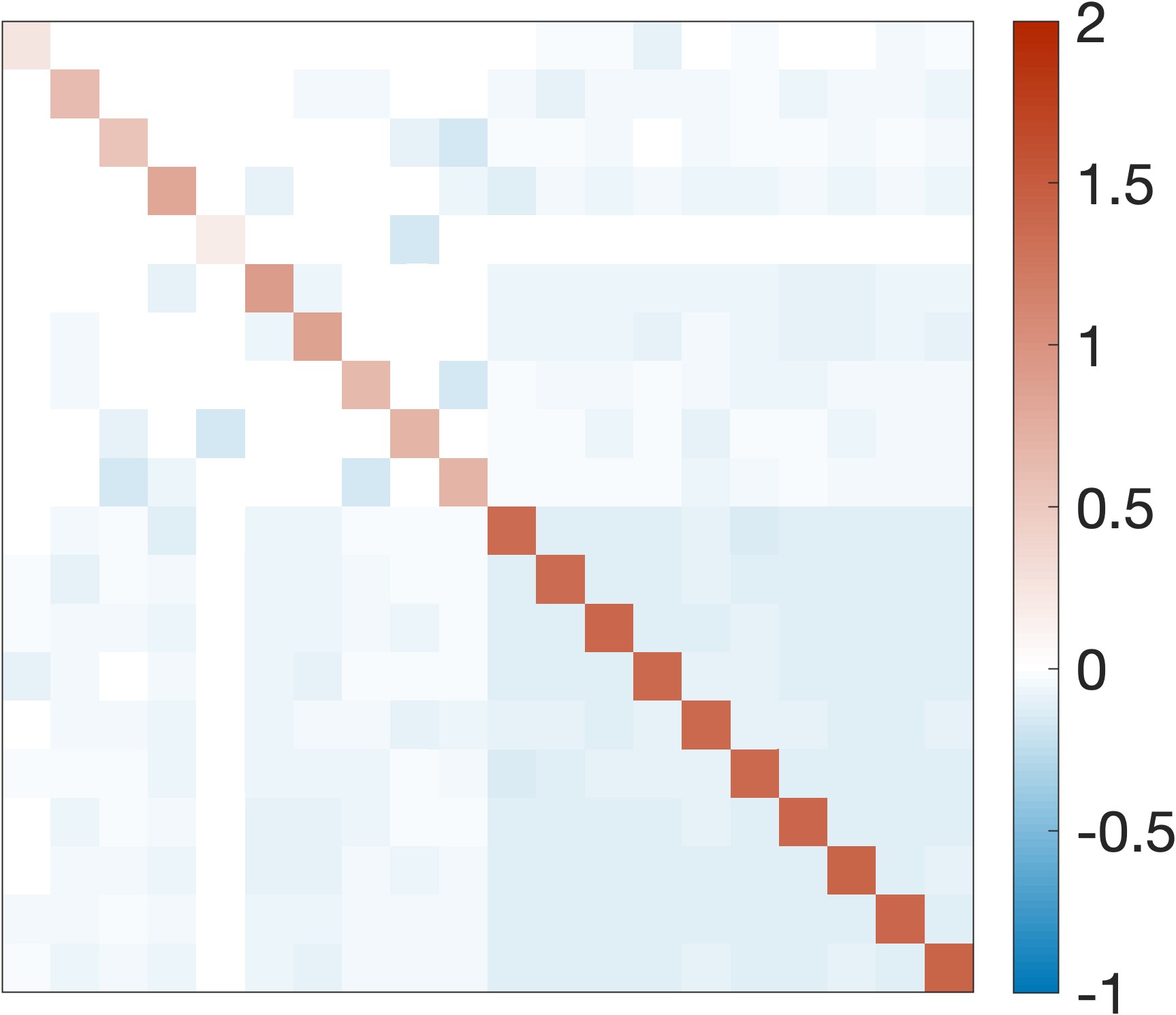}  &
    \includegraphics[width=2cm]{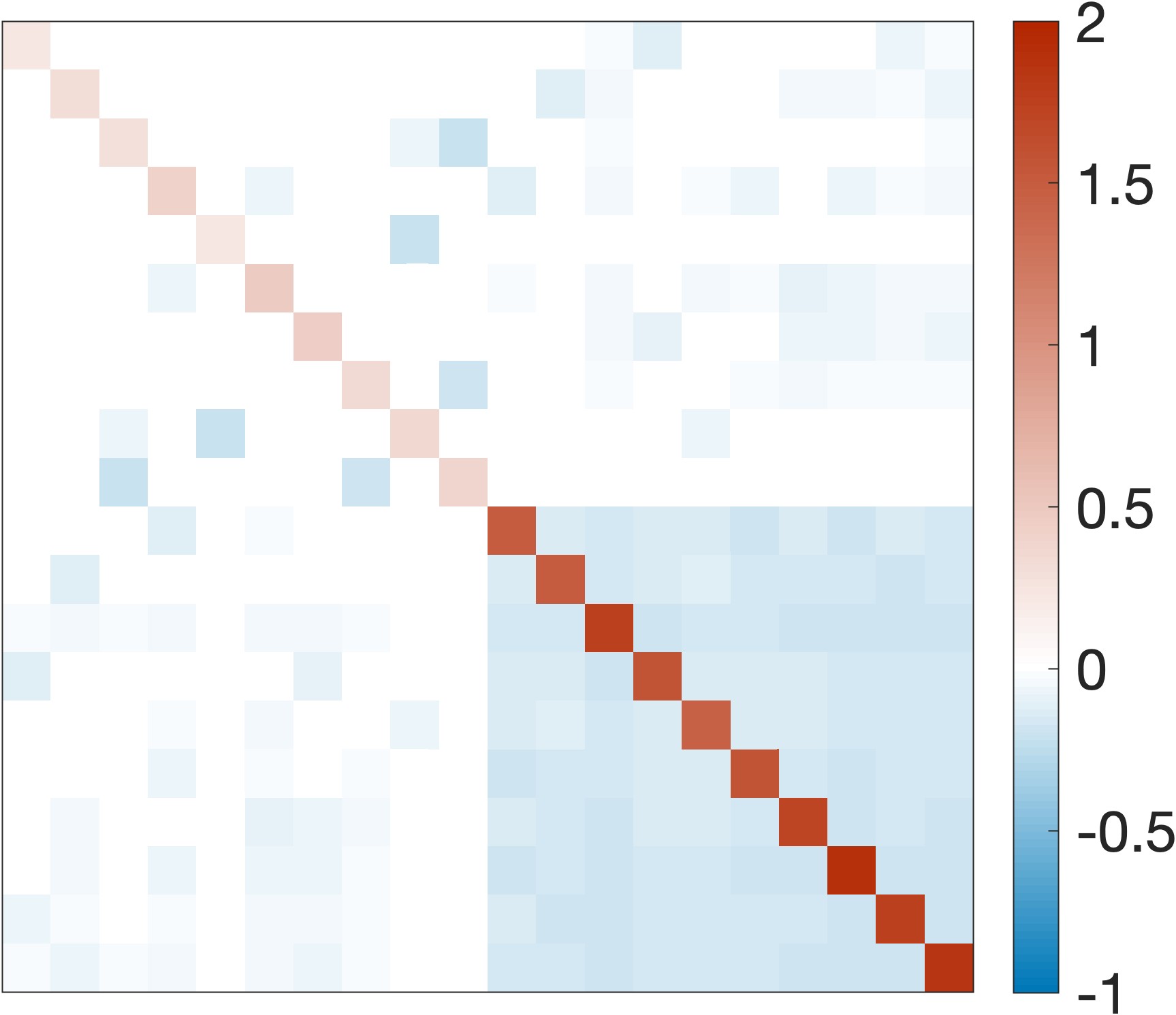} &
    \includegraphics[width=2cm]{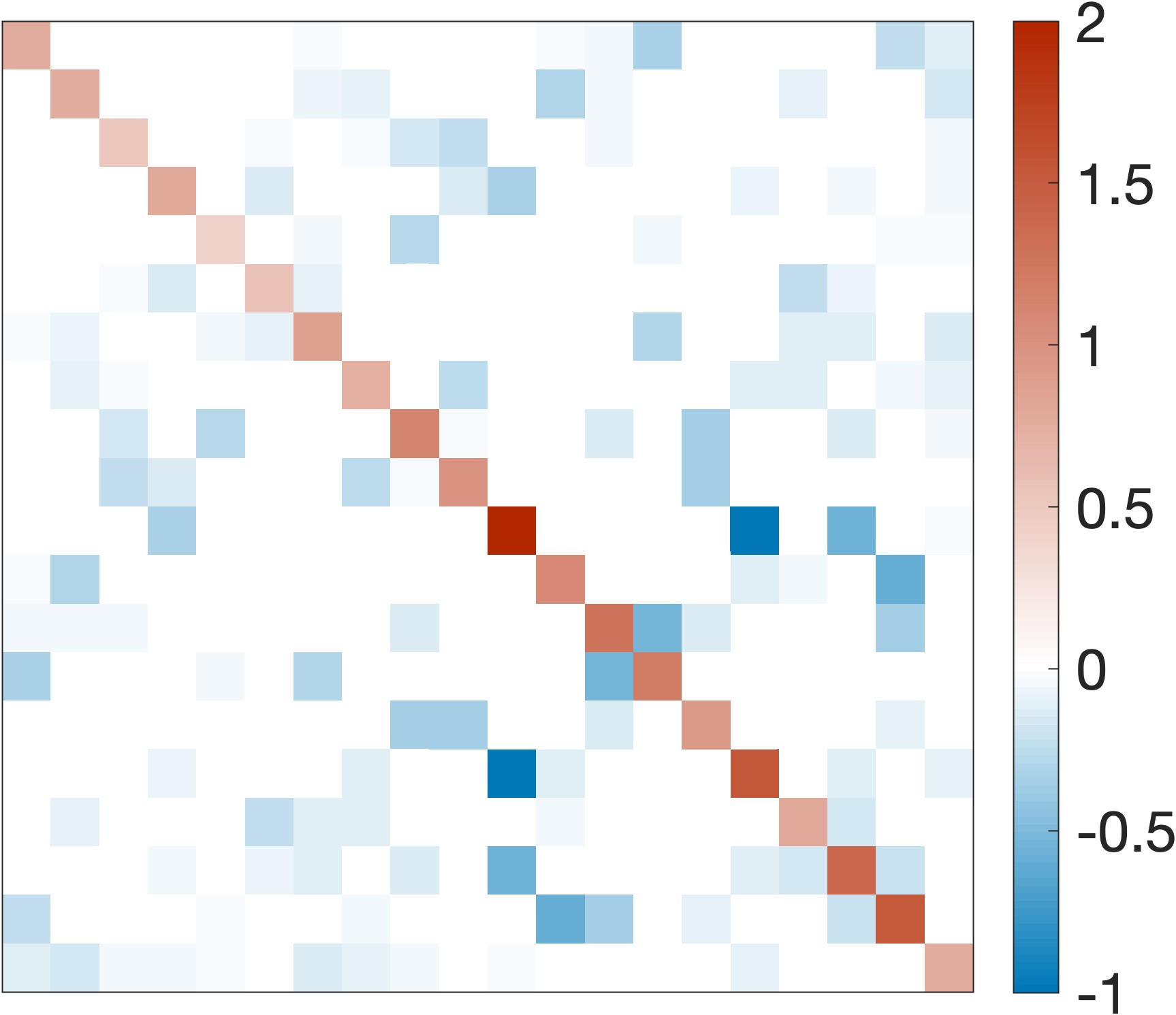}    &
    \includegraphics[width=2cm]{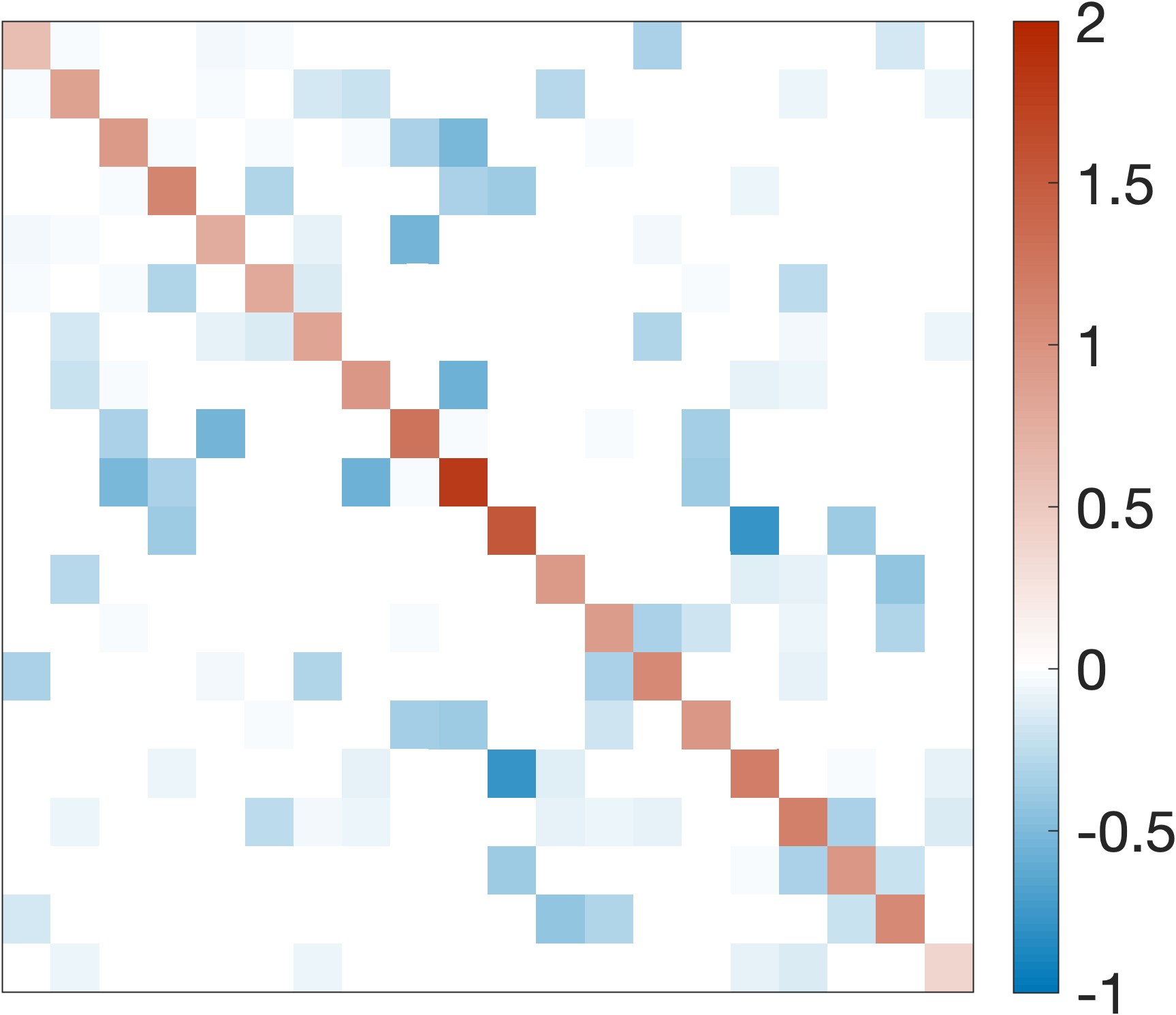}   &
    \includegraphics[width=2cm]{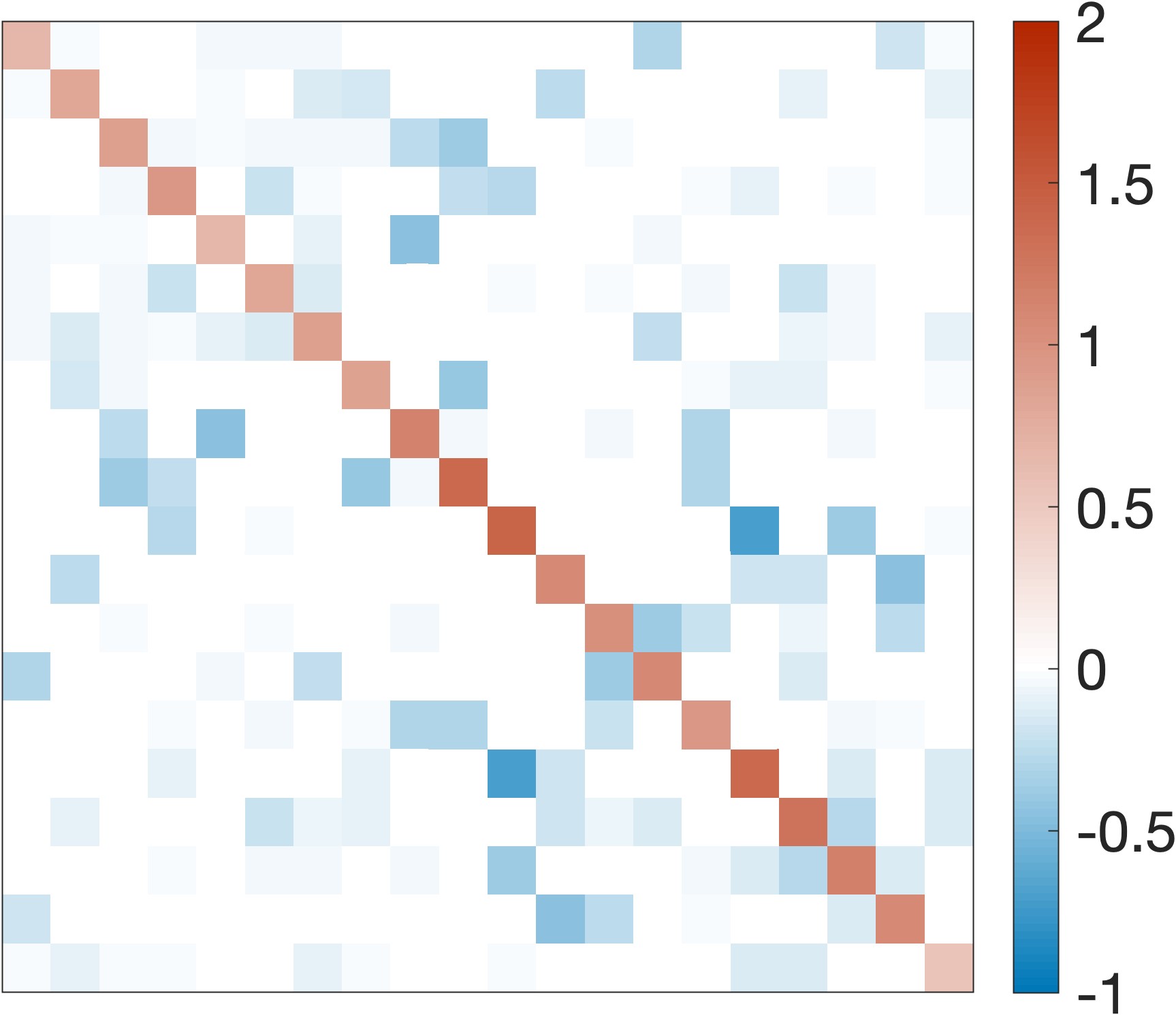} &
    \includegraphics[width=2cm]{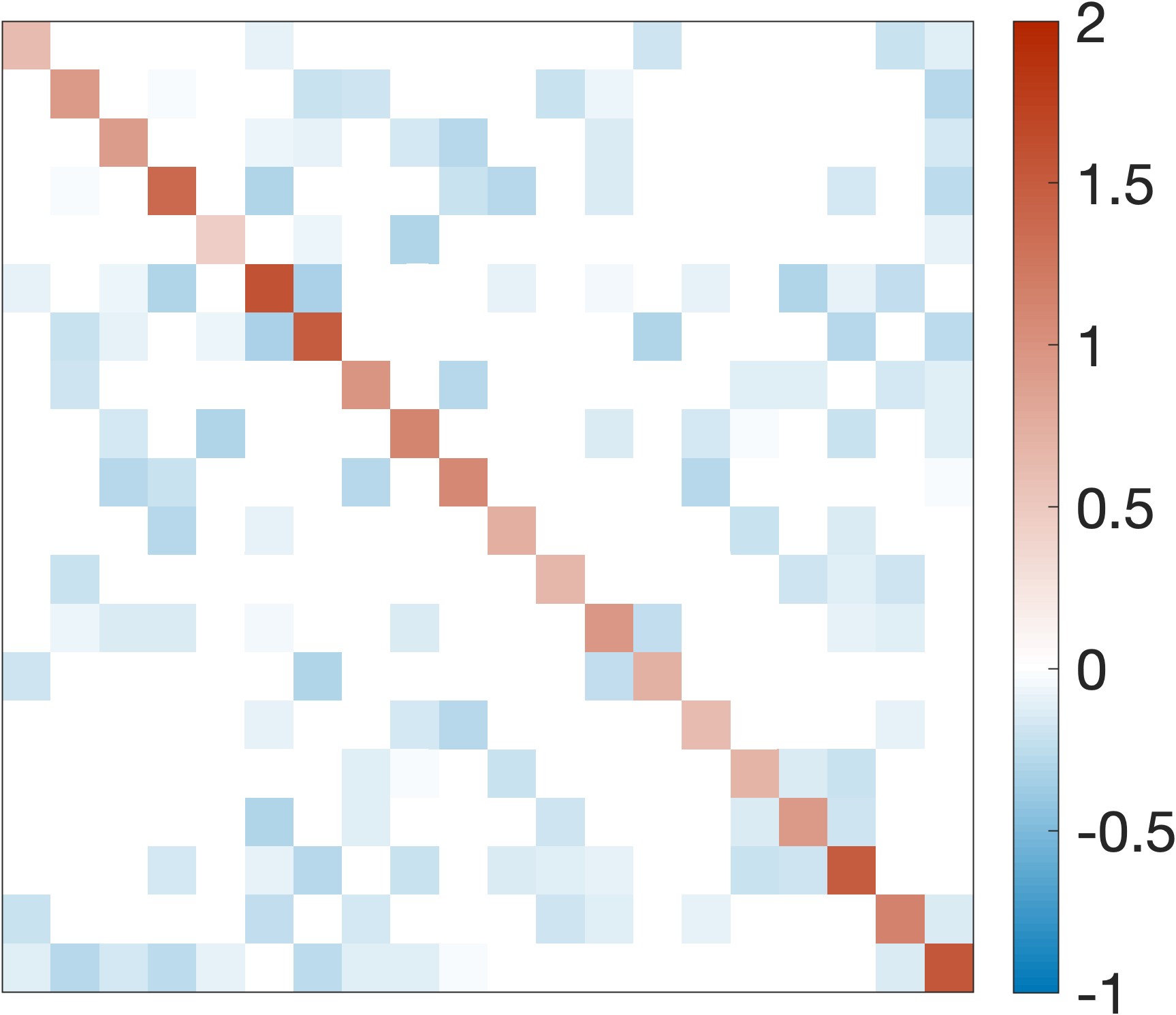}\\

    \includegraphics[width=2cm]{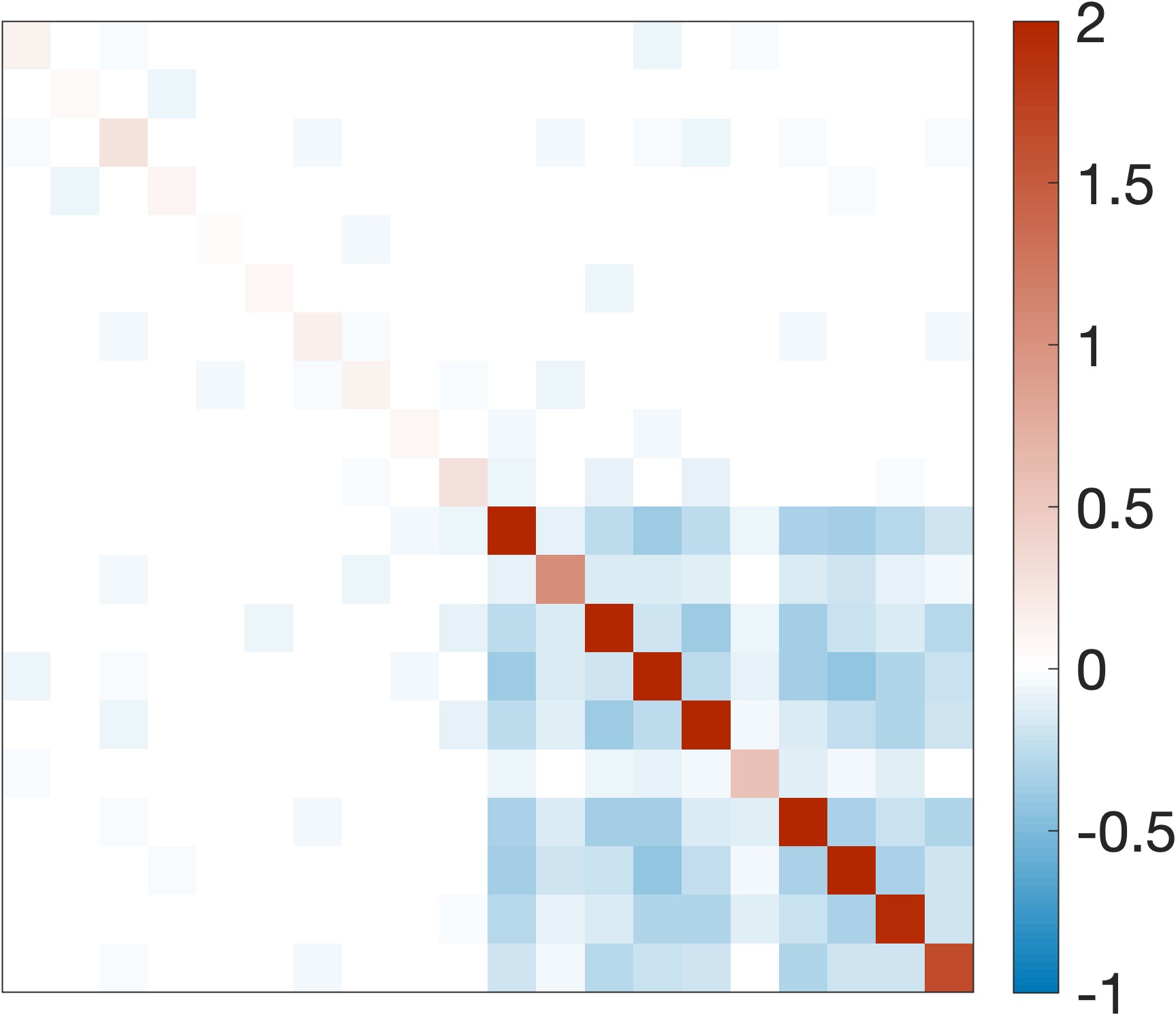}     &  \includegraphics[width=2cm]{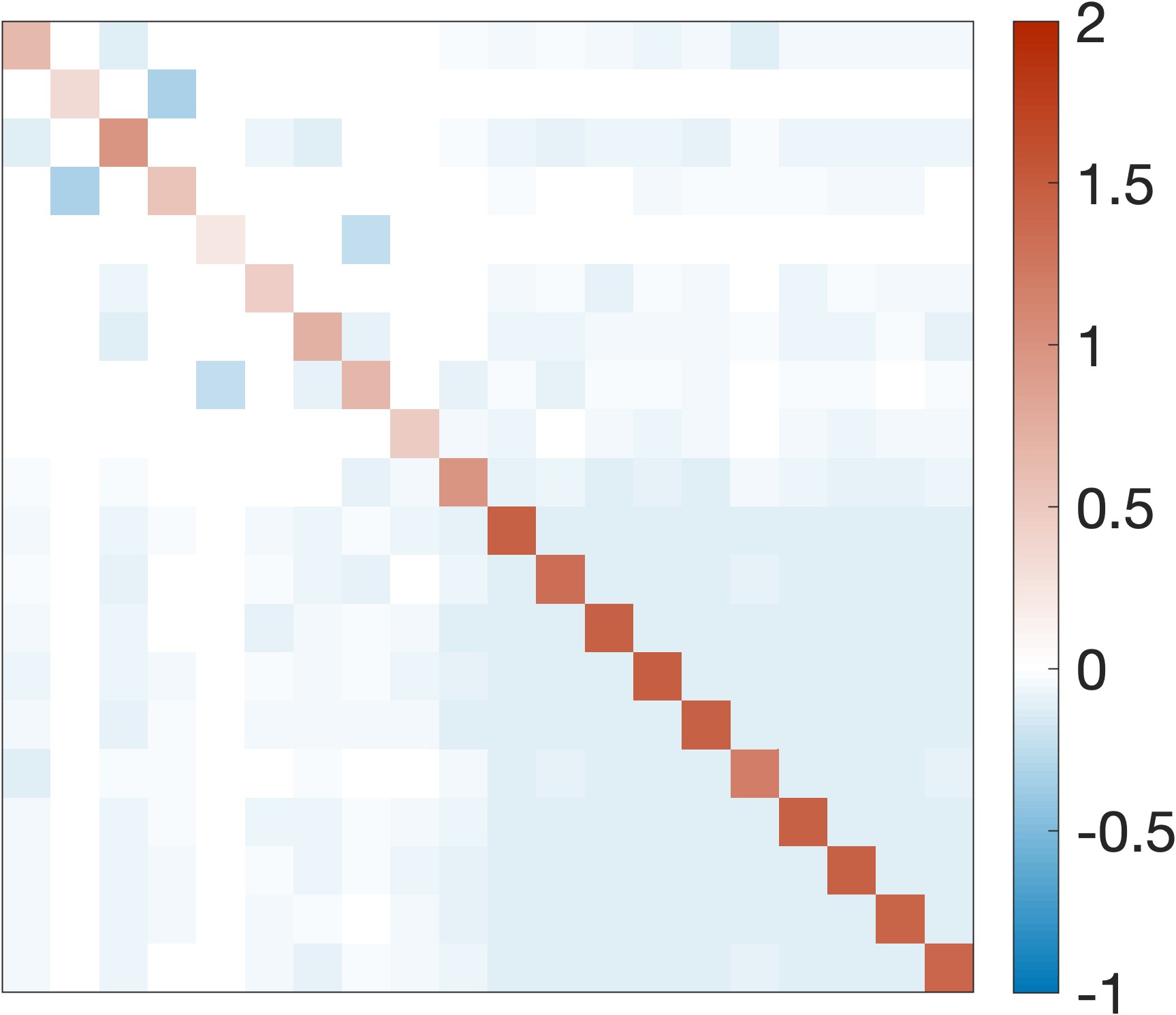}  &
    \includegraphics[width=2cm]{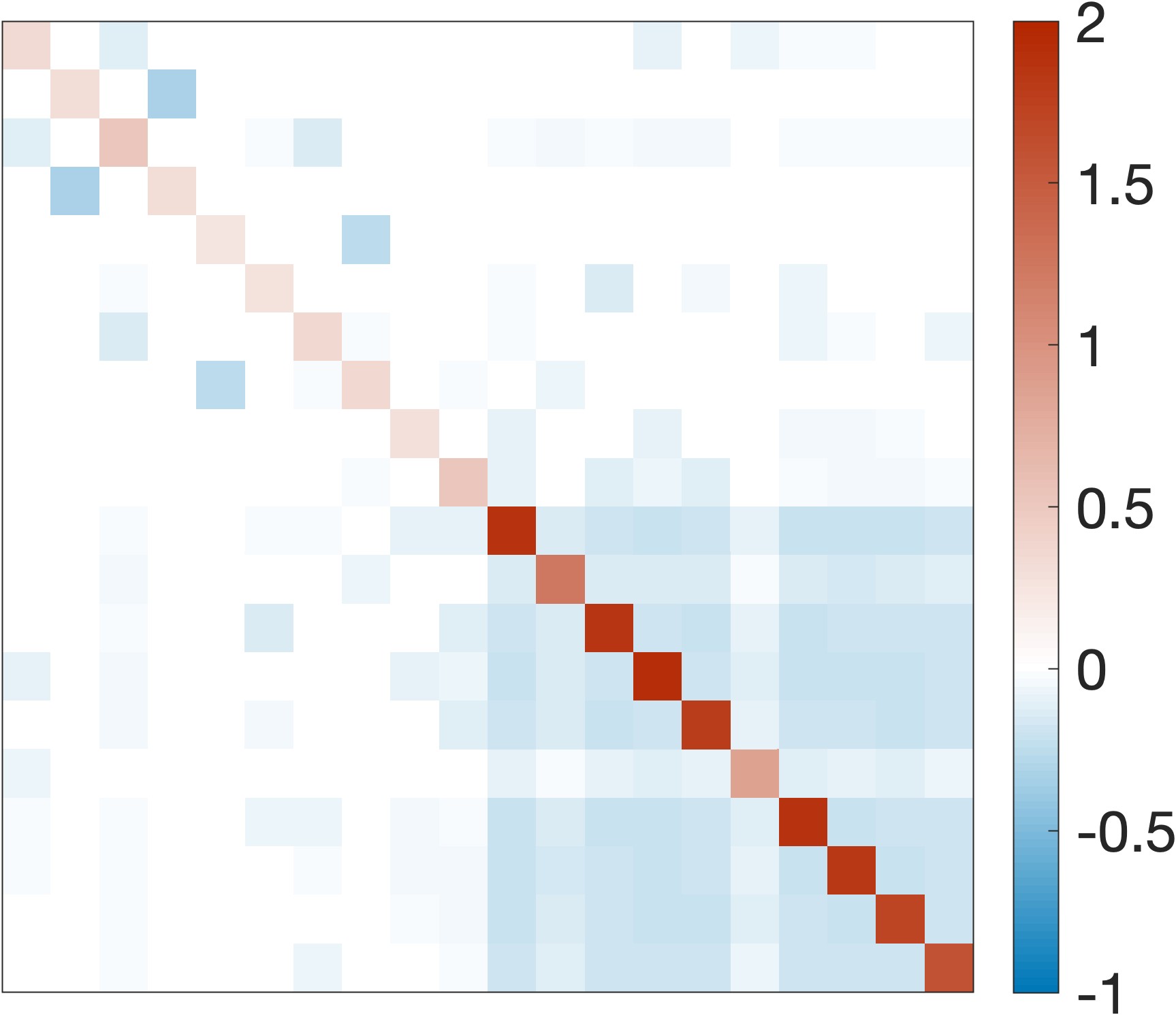} &
    \includegraphics[width=2cm]{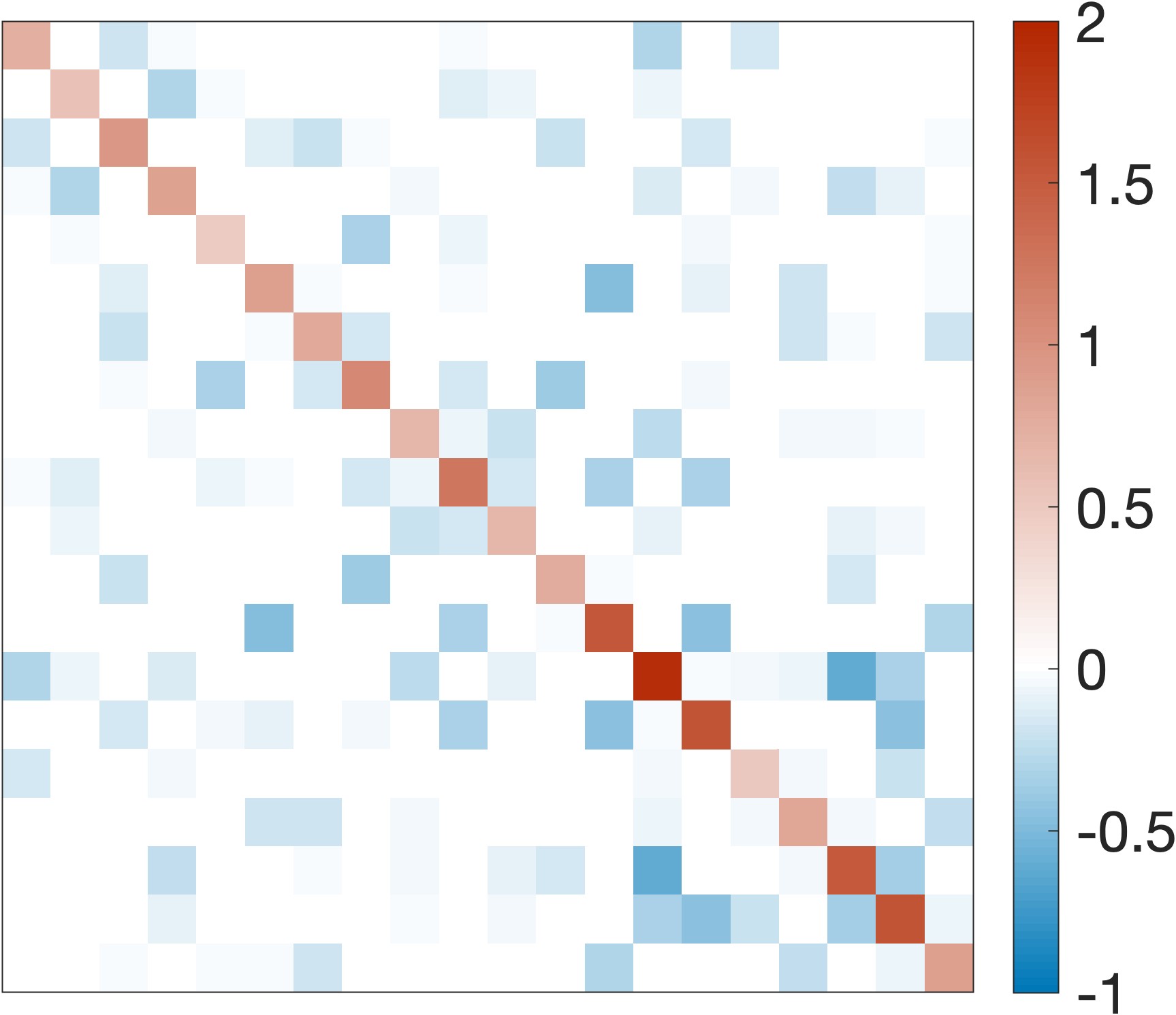}    &
    \includegraphics[width=2cm]{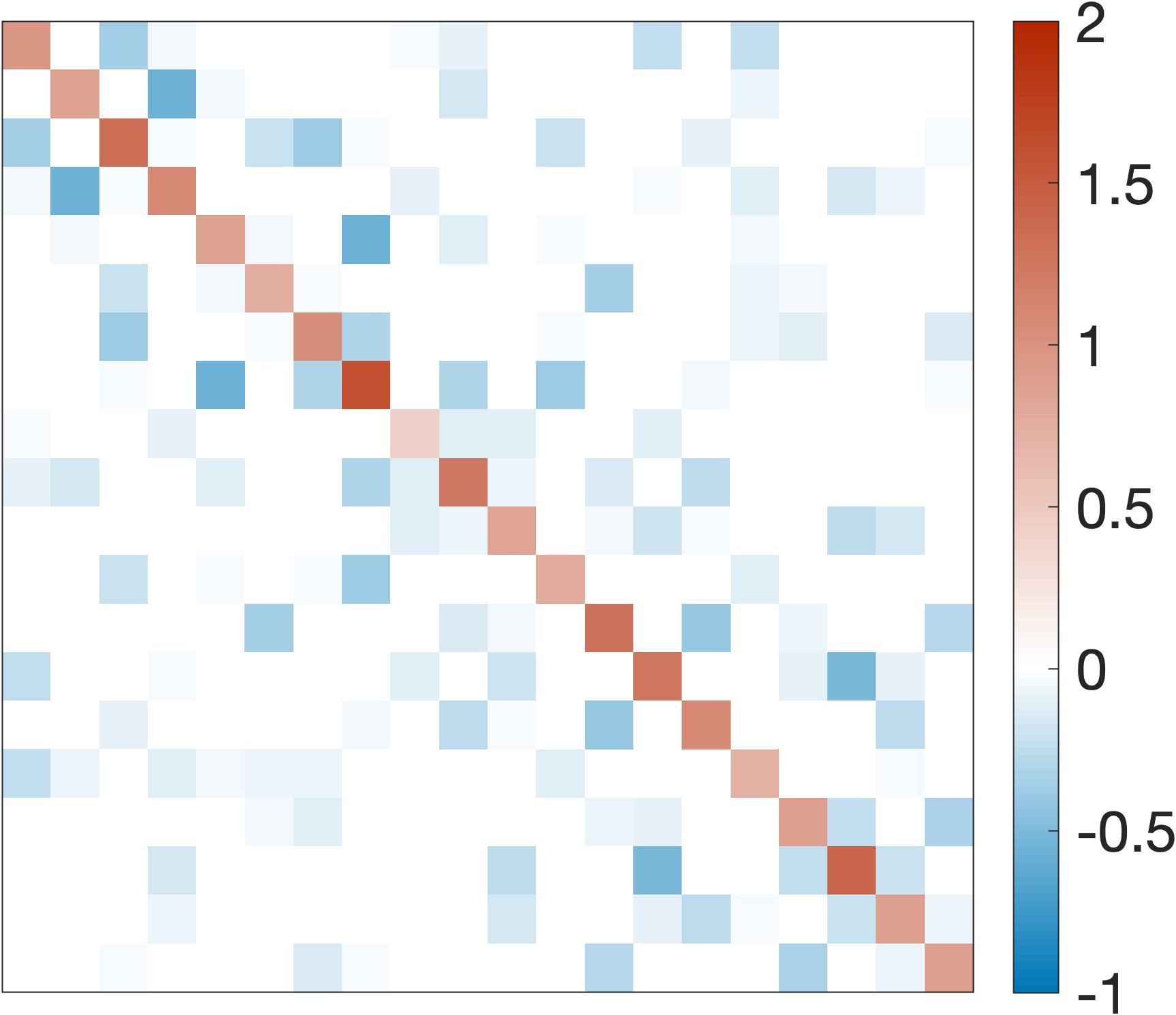}   &
    \includegraphics[width=2cm]{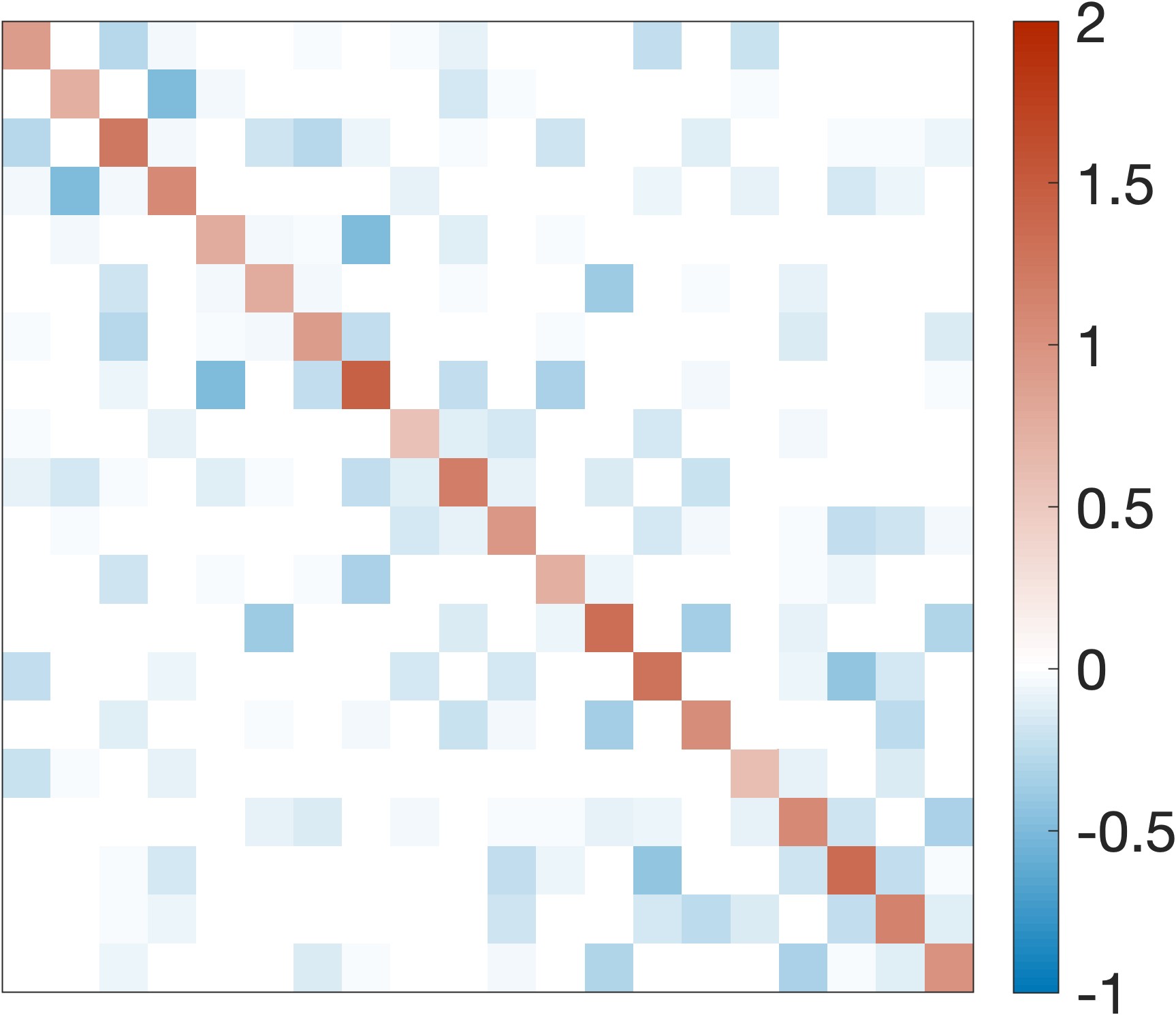} &
    \includegraphics[width=2cm]{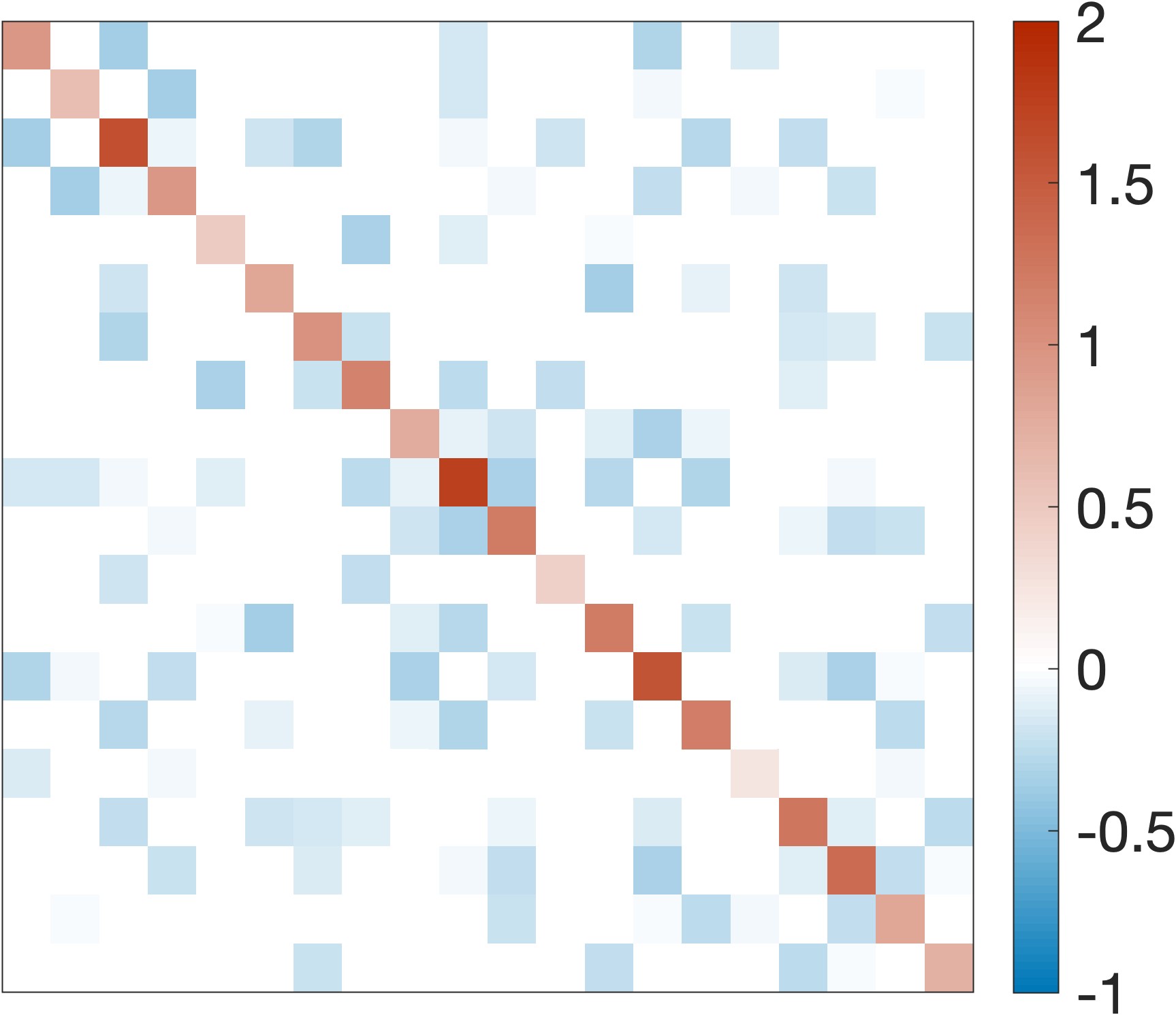}\\

    \includegraphics[width=2cm]{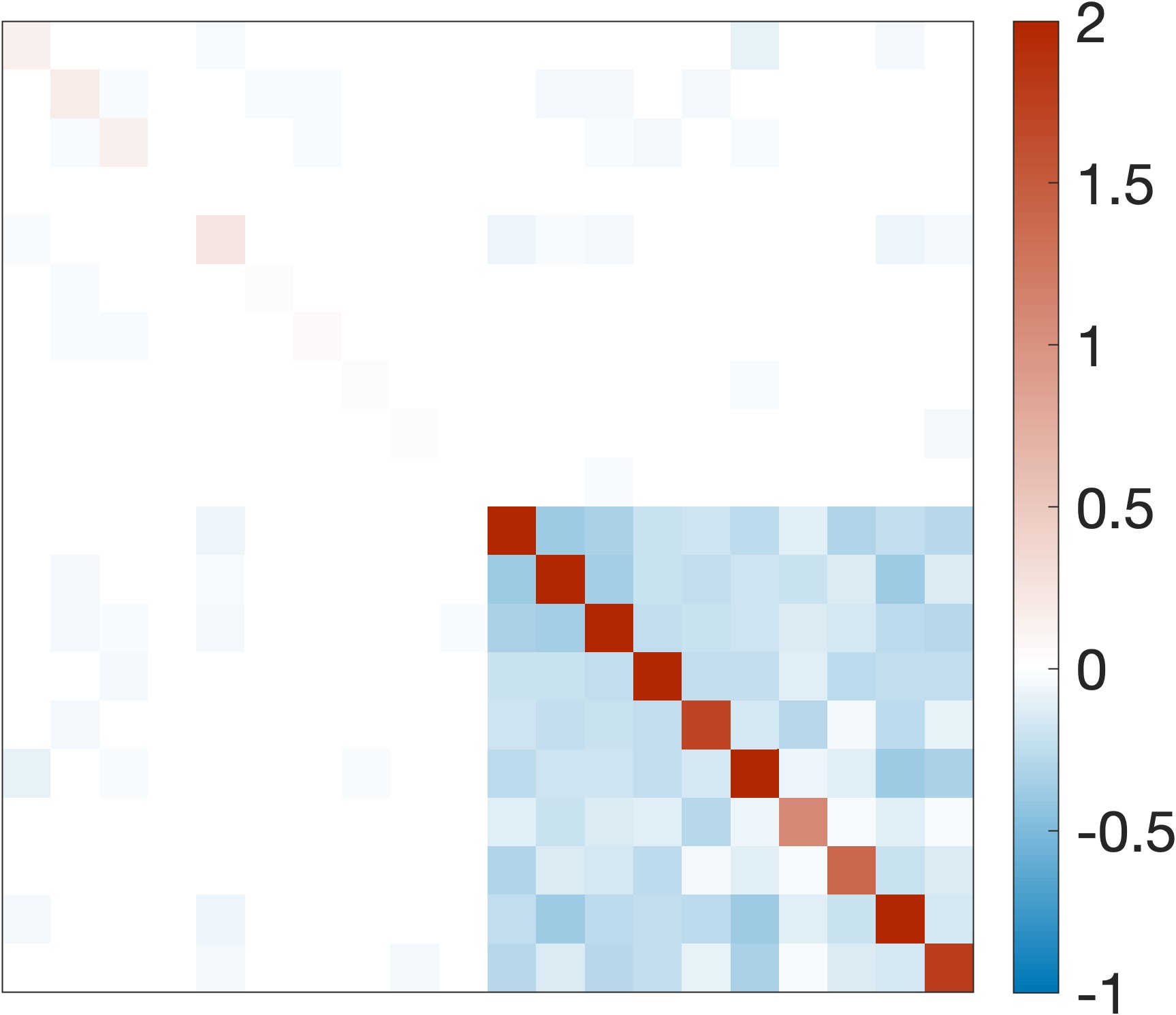}     &  \includegraphics[width=2cm]{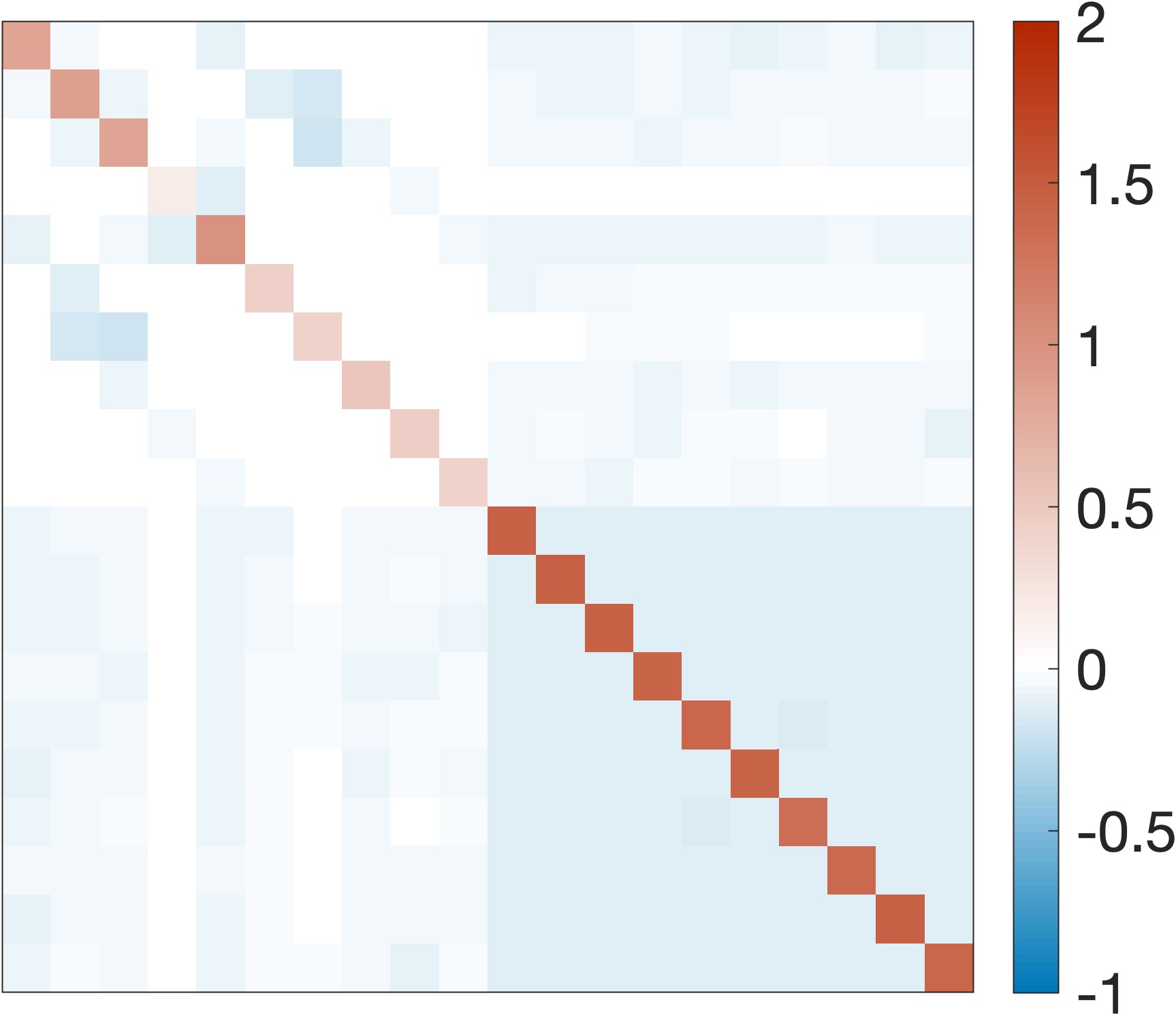}  &
    \includegraphics[width=2cm]{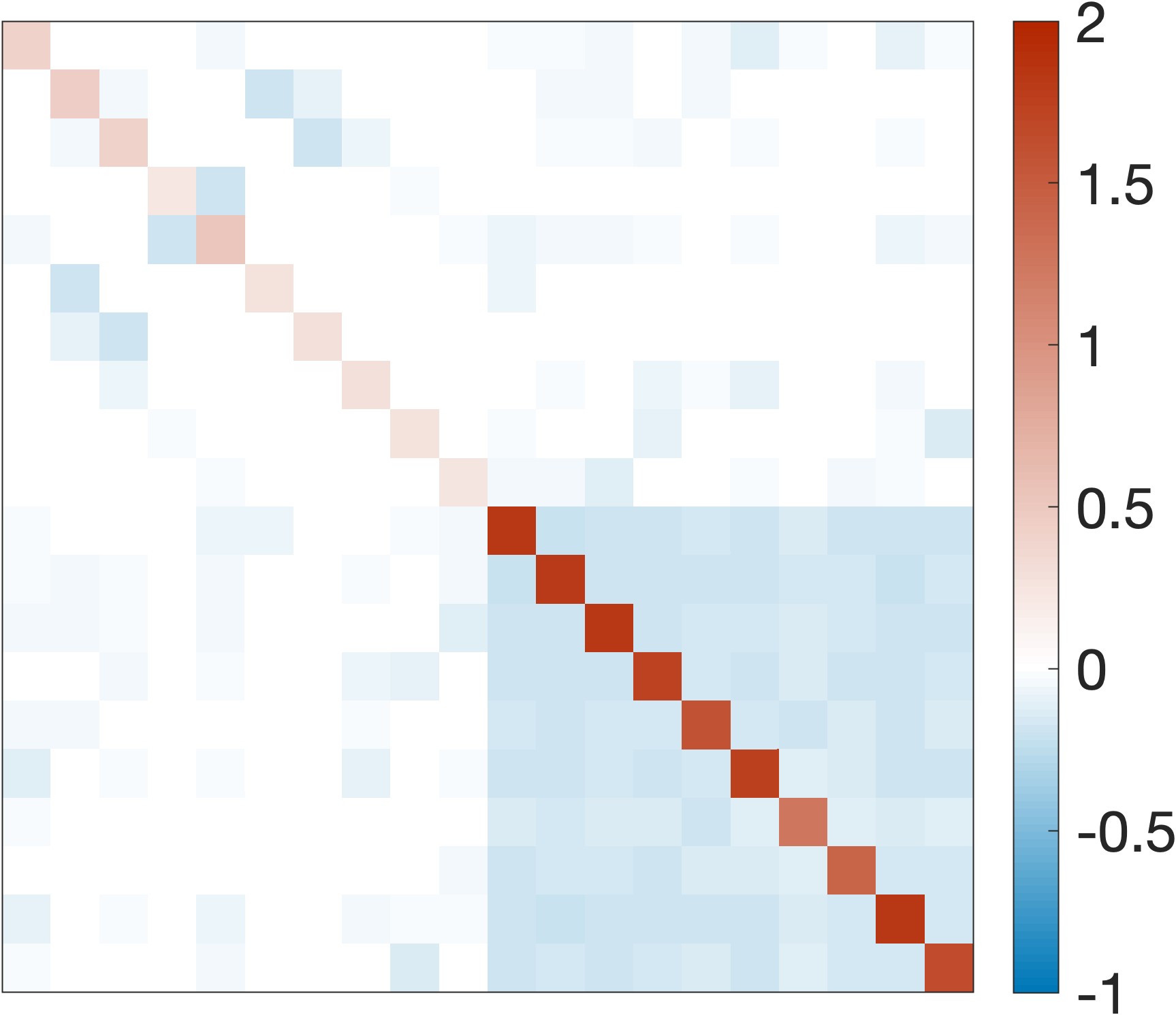} &
    \includegraphics[width=2cm]{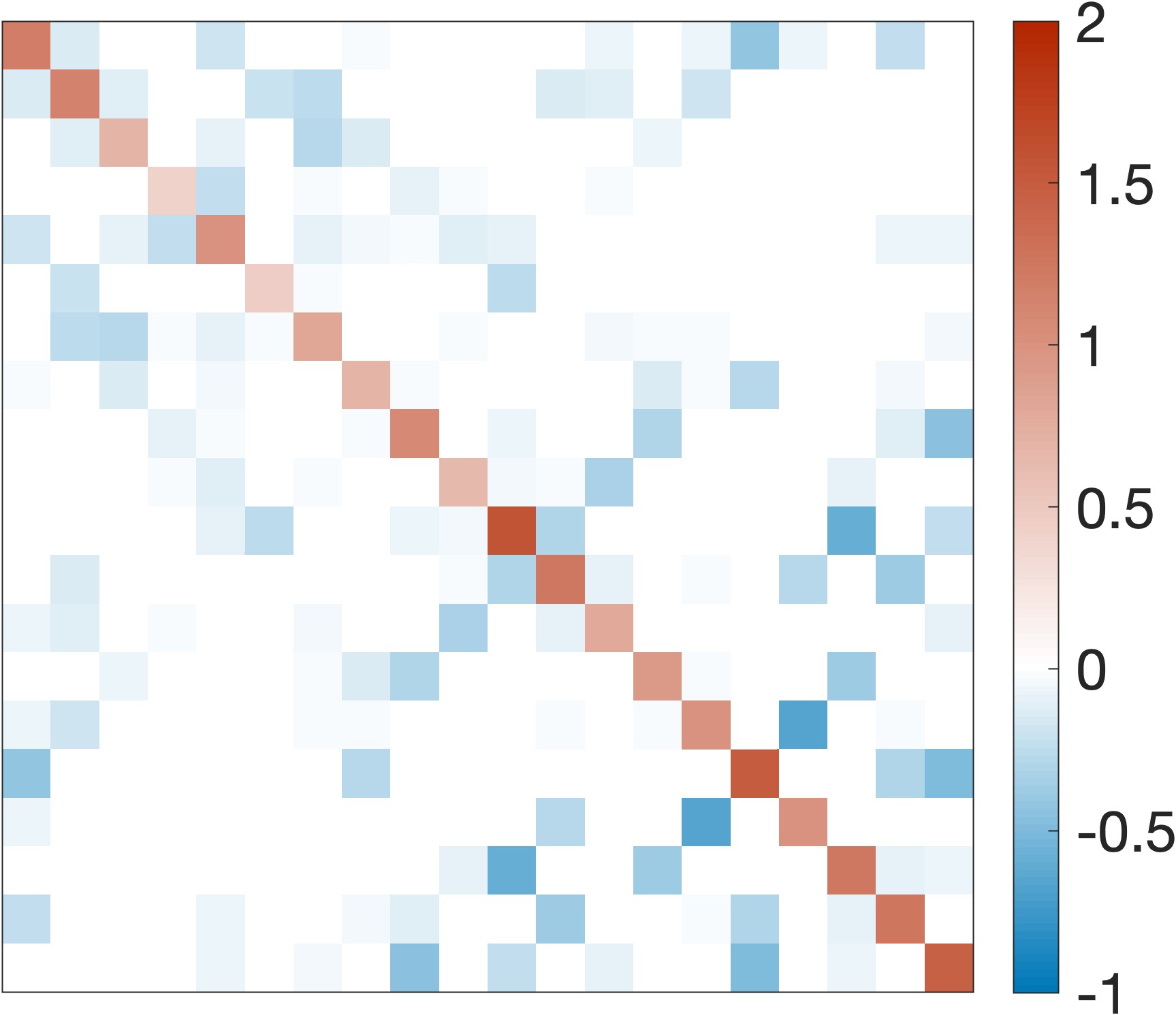}    &
    \includegraphics[width=2cm]{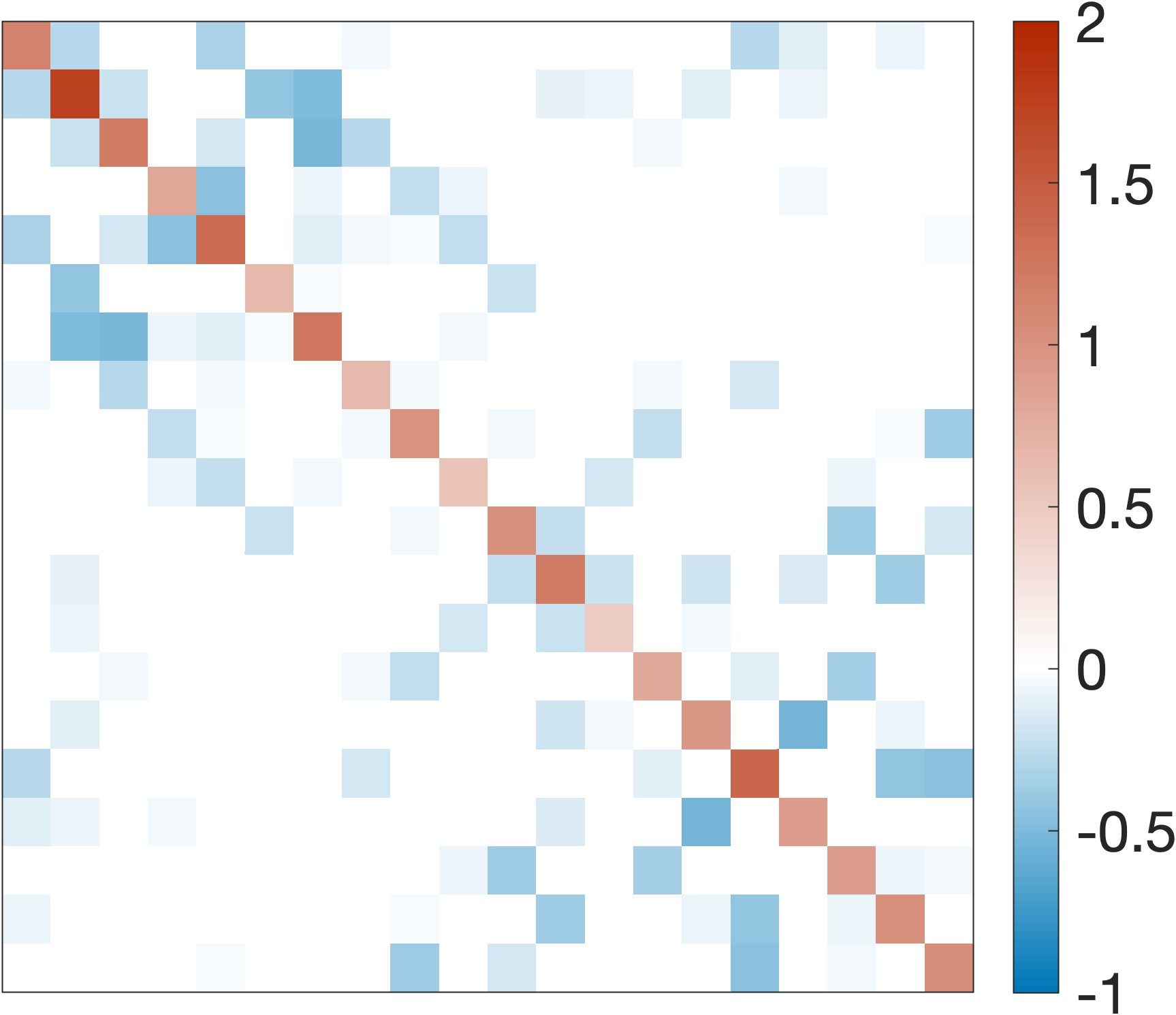}   &
    \includegraphics[width=2cm]{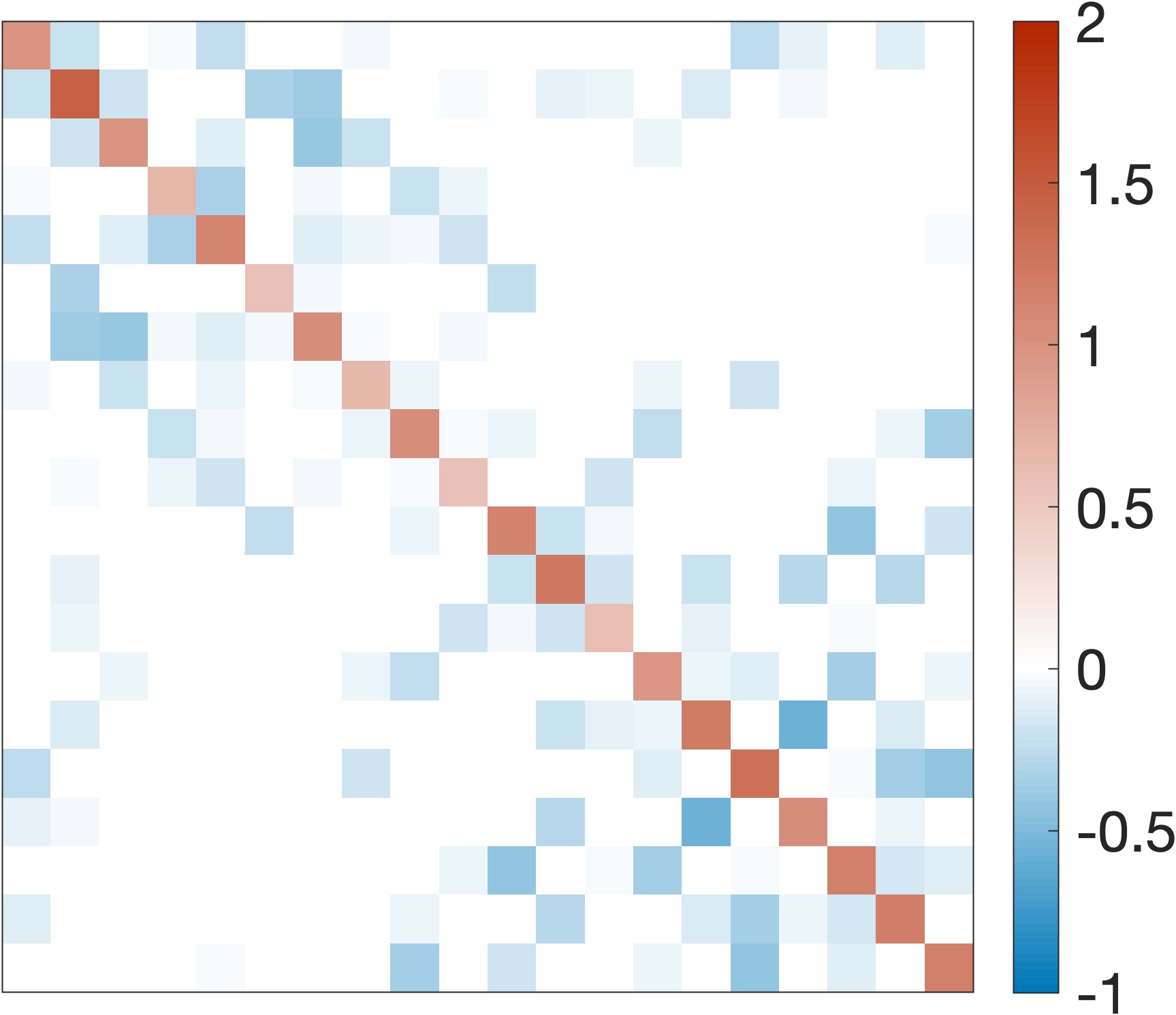} &
    \includegraphics[width=2cm]{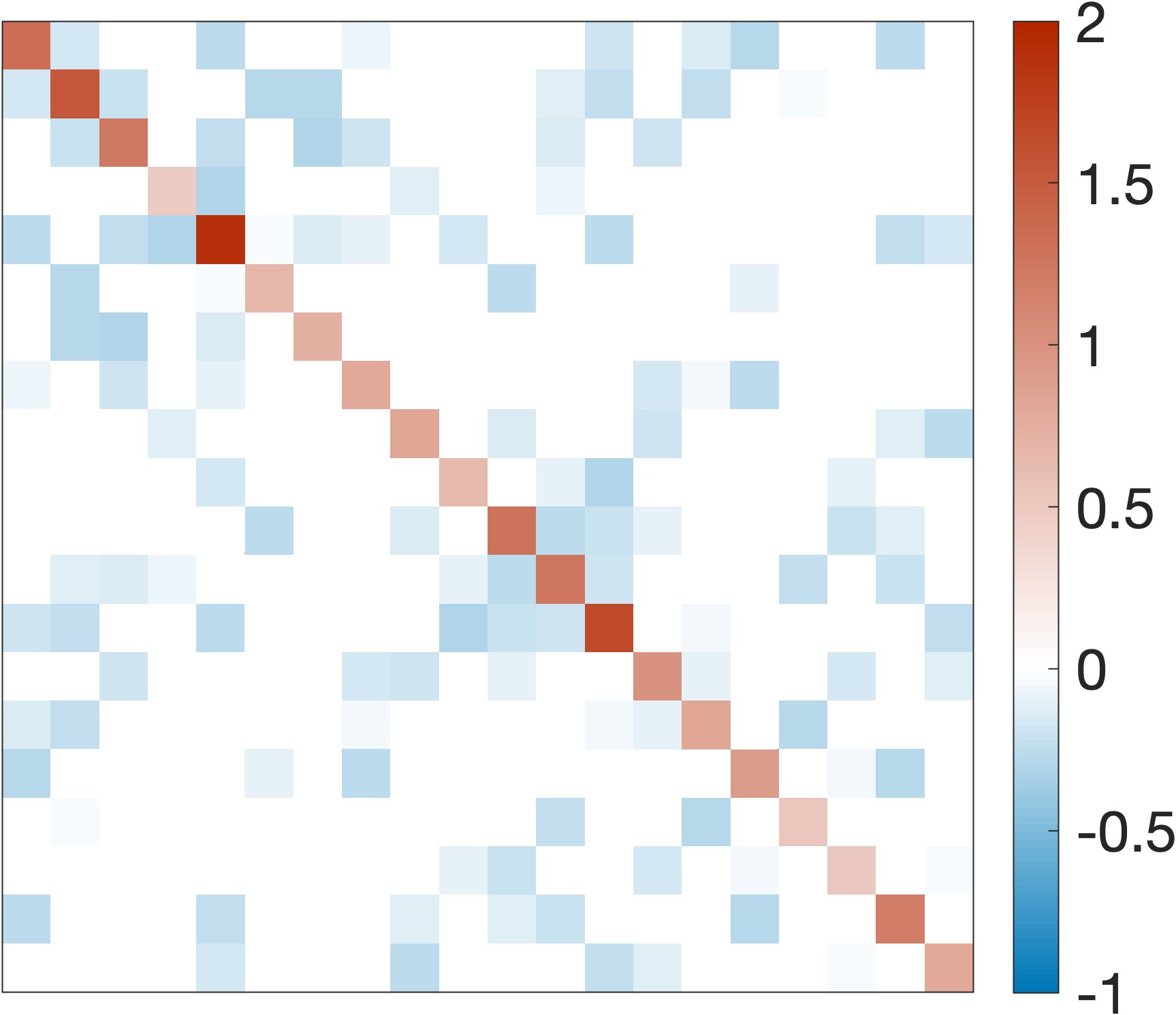}\\

    \end{tabular}
    \caption{Graph Laplacians estimated by different methods and the ground truth Erd\H{o}s-R\'enyi graph Laplacian (p=0.3).}
    \label{fig:more_er_laplacians}

    \centering
    \begin{tabular}{cccccccc}
    SCGL    & GLS-1 & GLS-2 & CGL   & GLEN  &GLEN-VI    & Ground Truth \\
    
    \includegraphics[width=2cm]{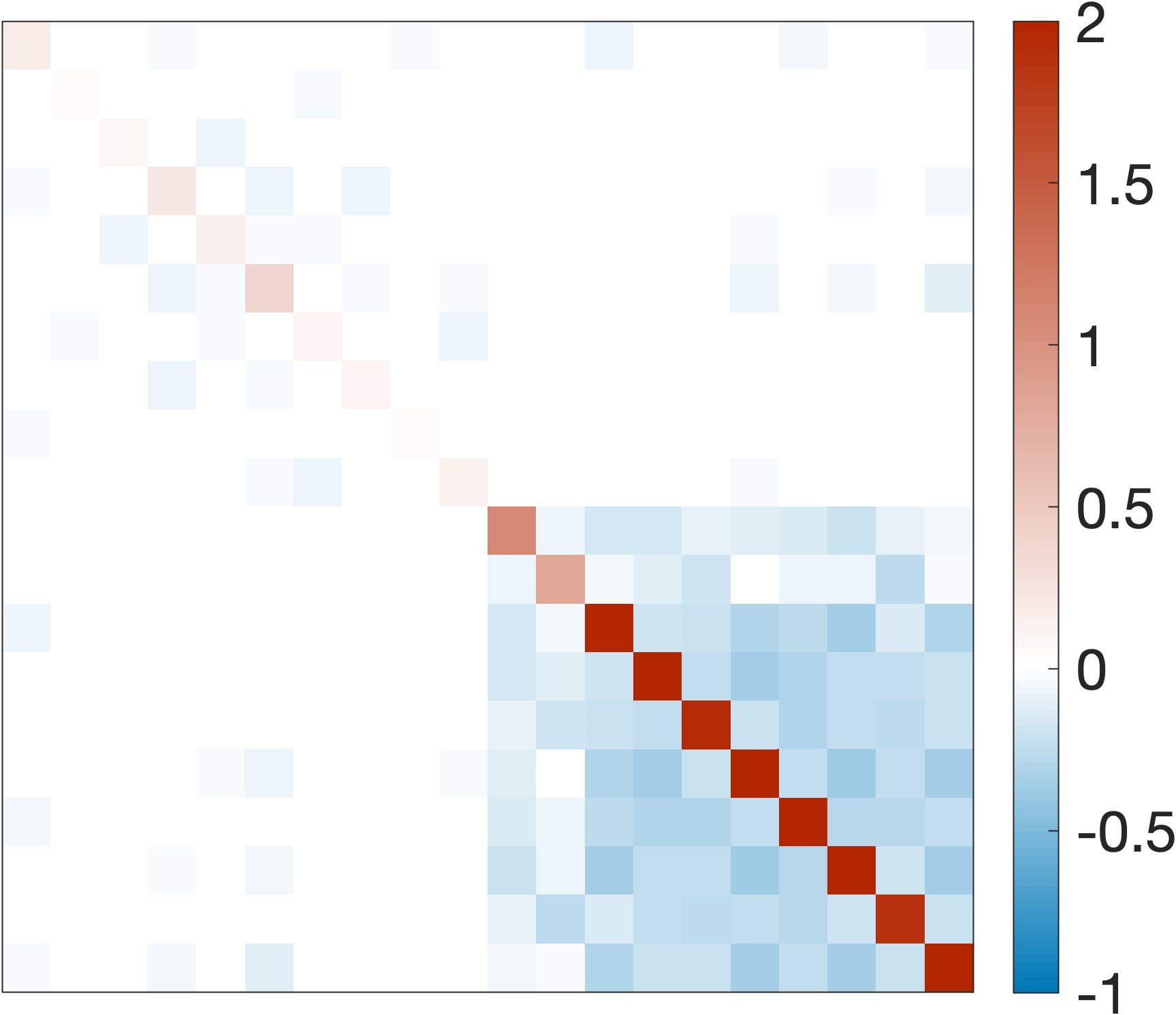}     &  \includegraphics[width=2cm]{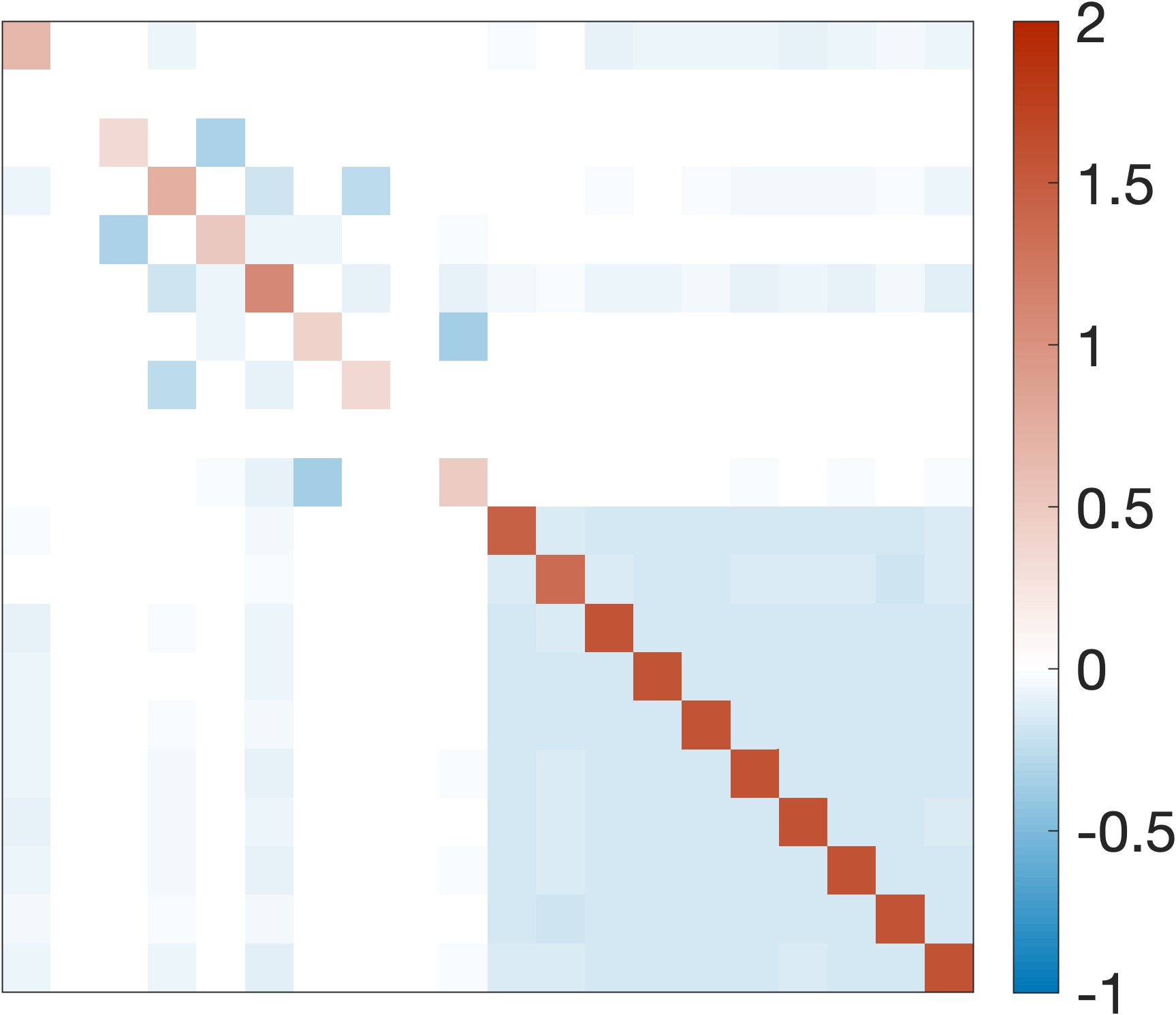}  &
    \includegraphics[width=2cm]{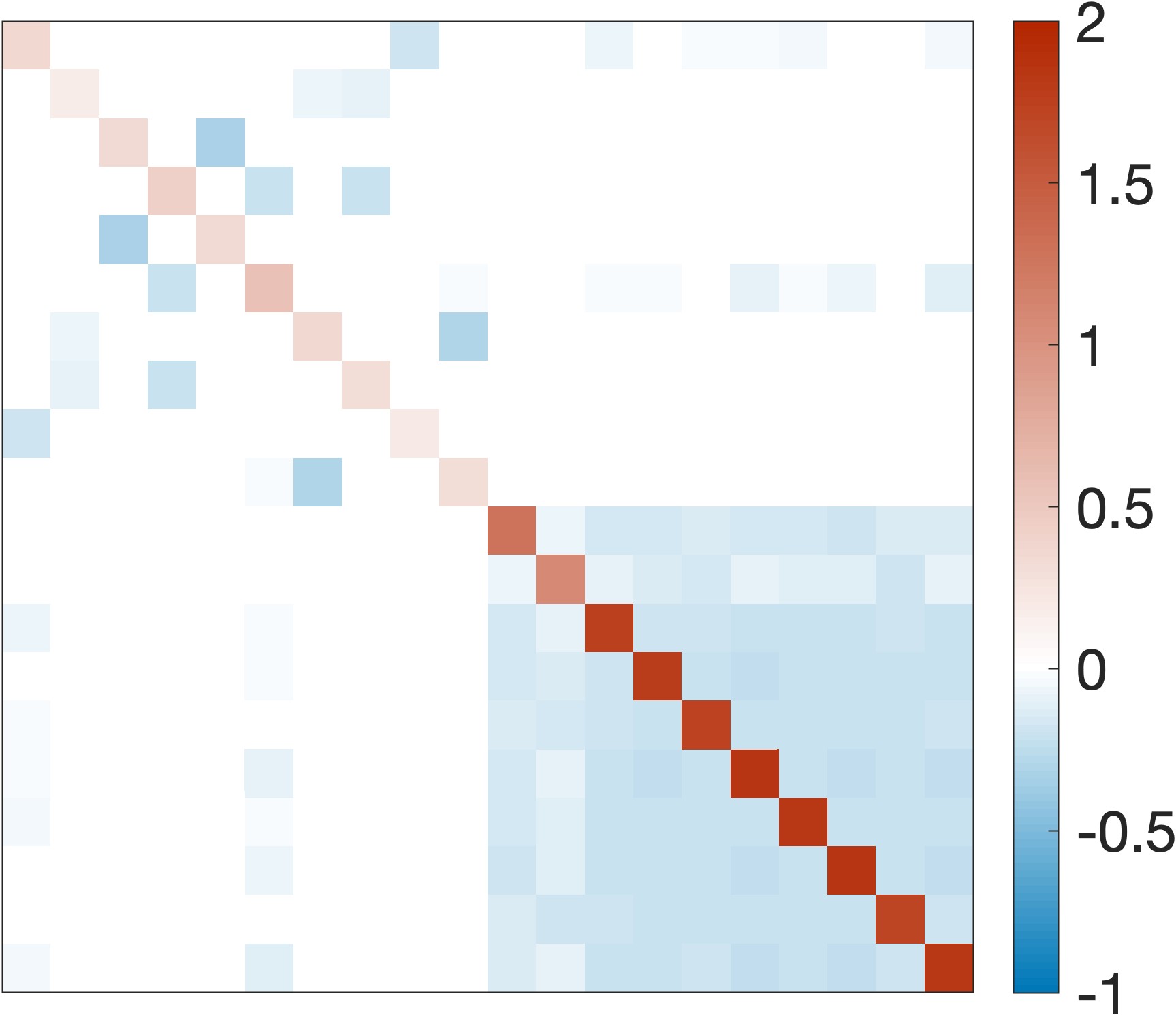} &
    \includegraphics[width=2cm]{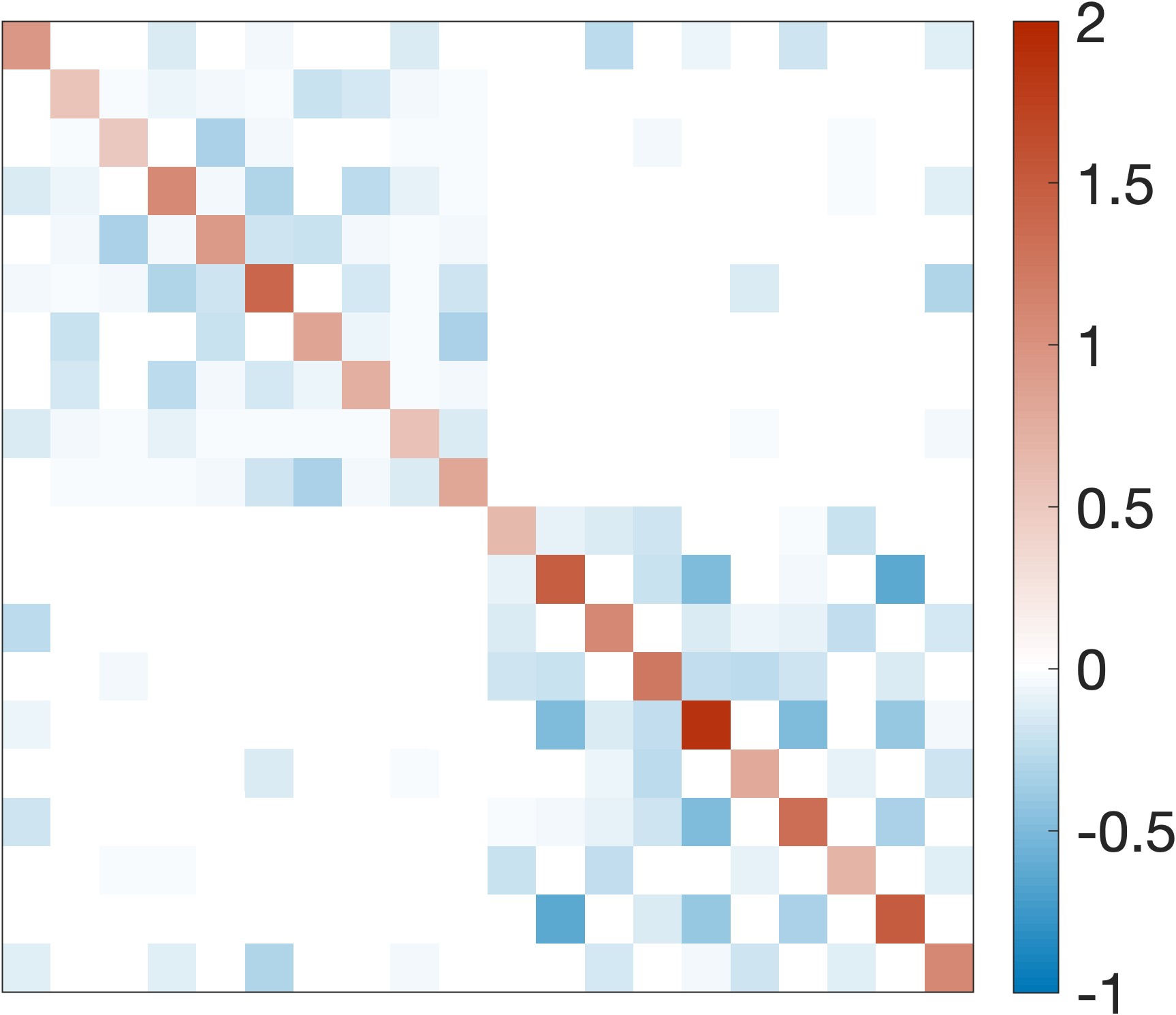}    &
    \includegraphics[width=2cm]{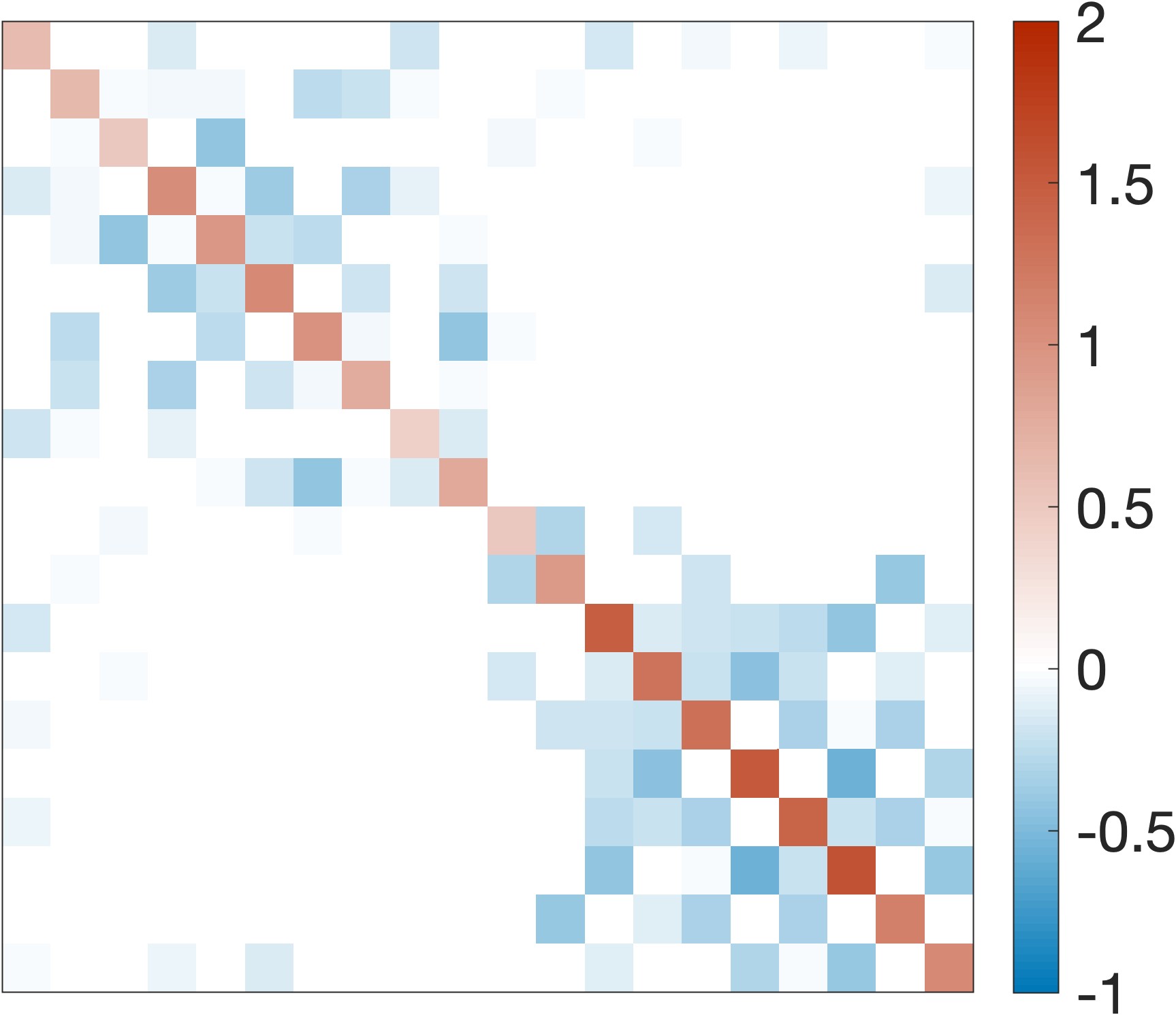}   &
    \includegraphics[width=2cm]{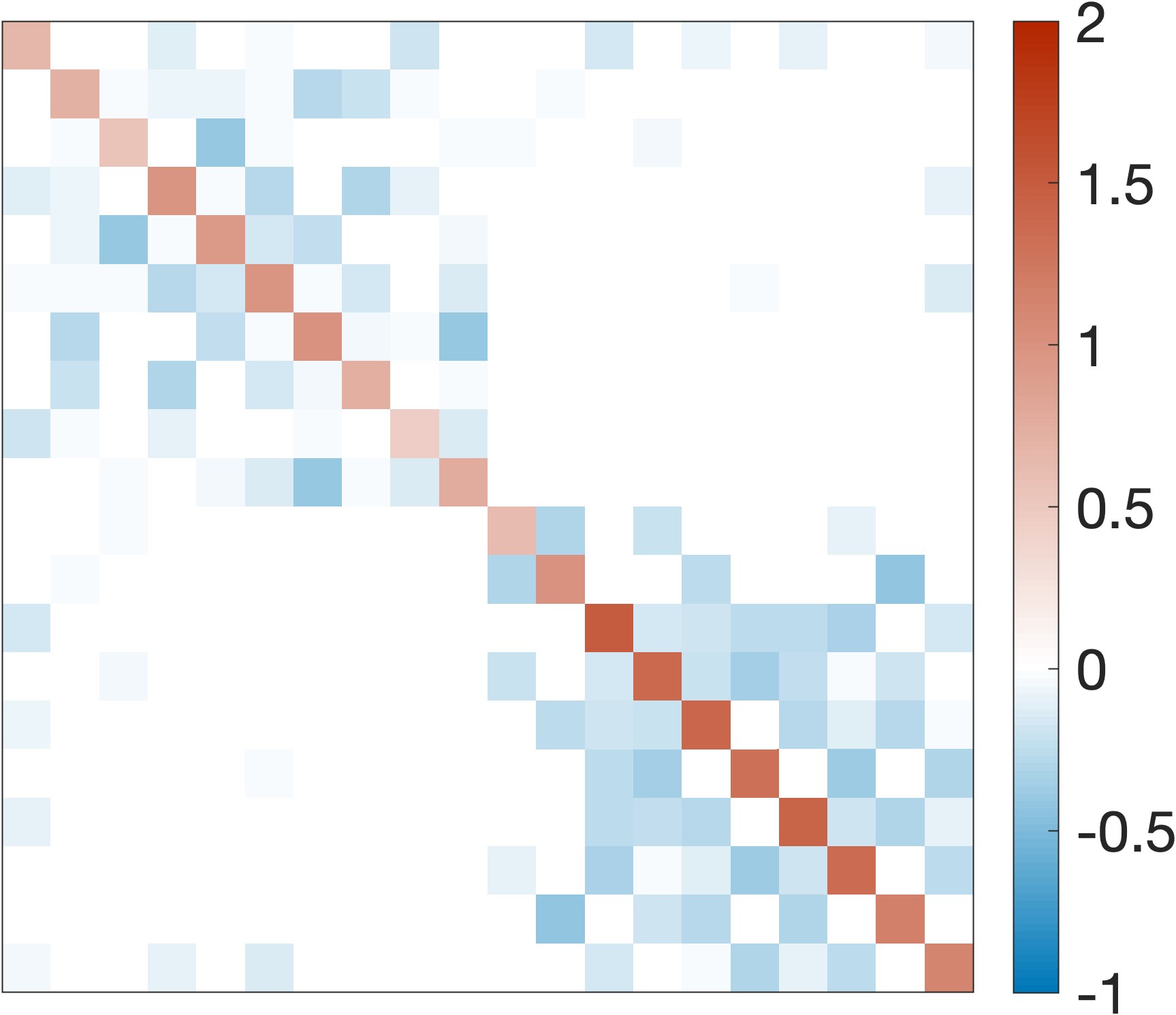} &
    \includegraphics[width=2cm]{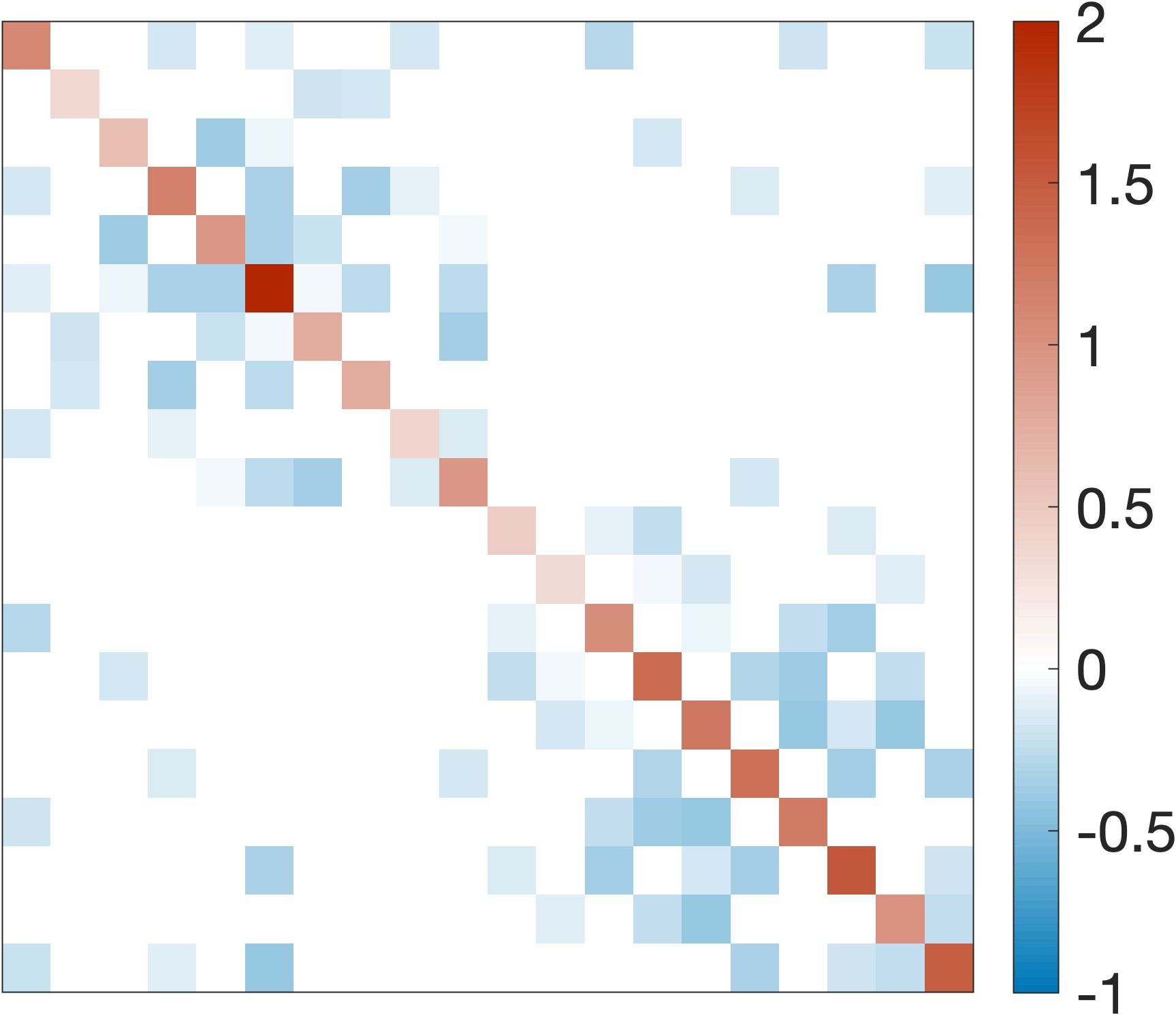}\\

    \includegraphics[width=2cm]{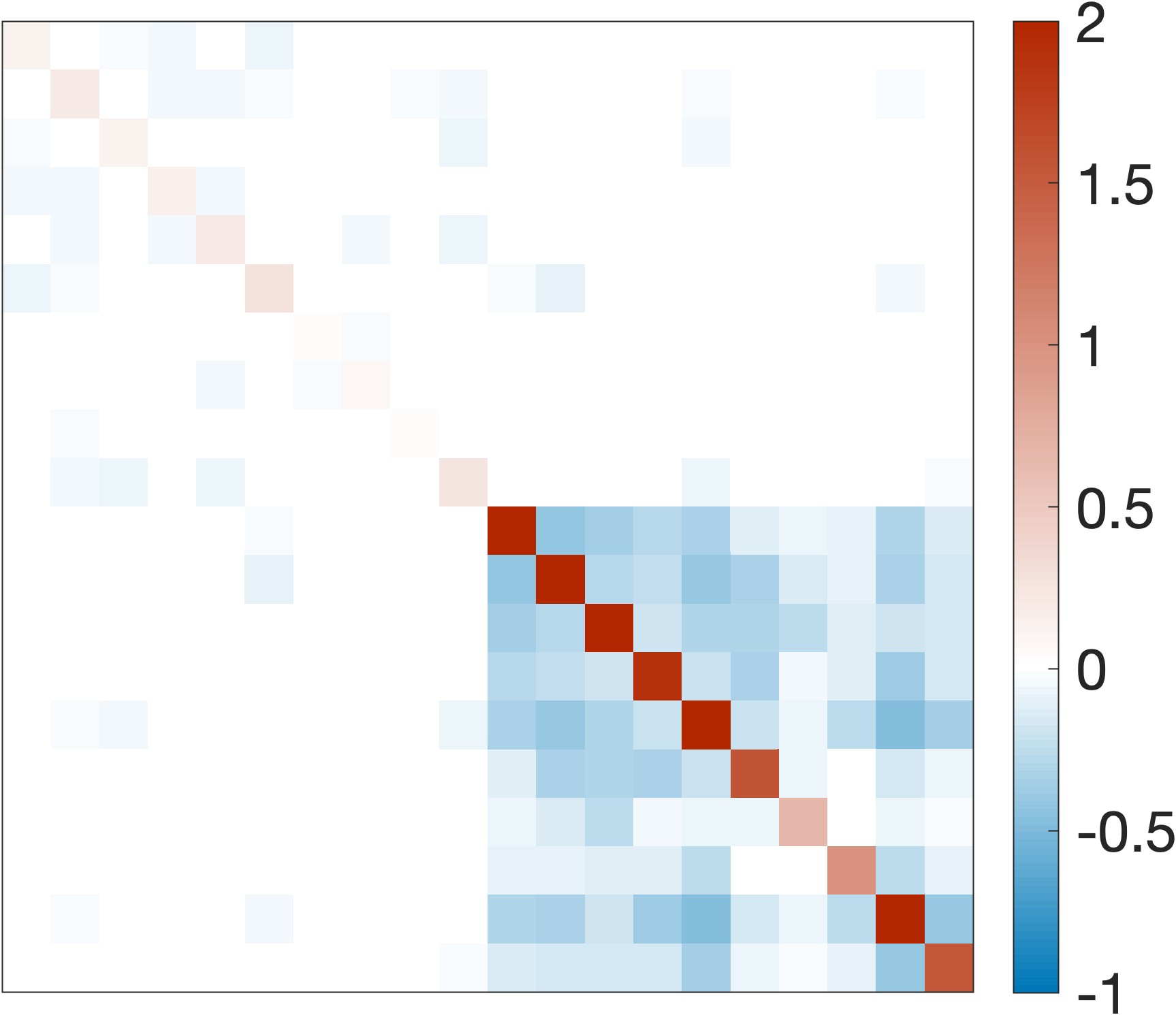}     &  \includegraphics[width=2cm]{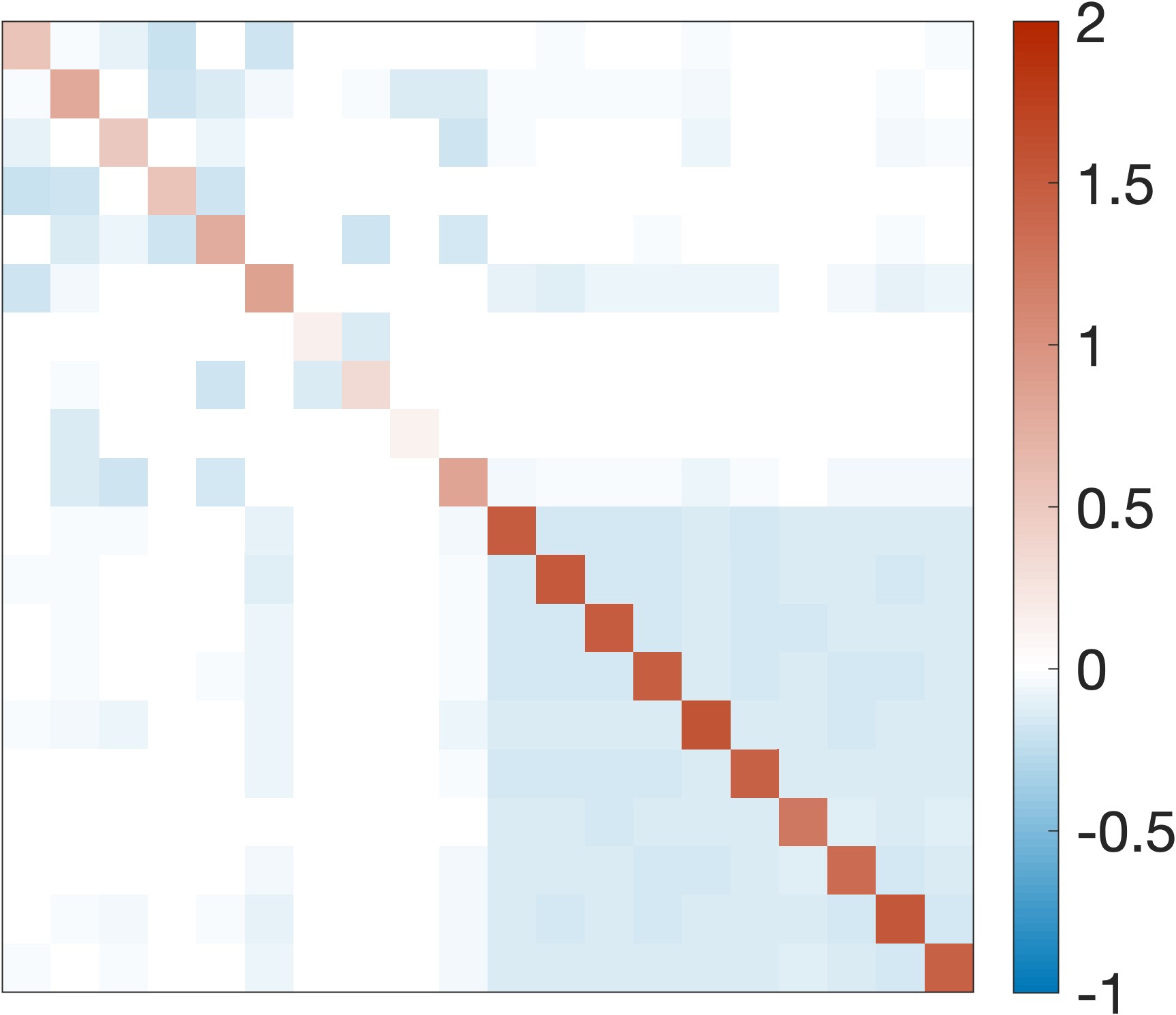}  &
    \includegraphics[width=2cm]{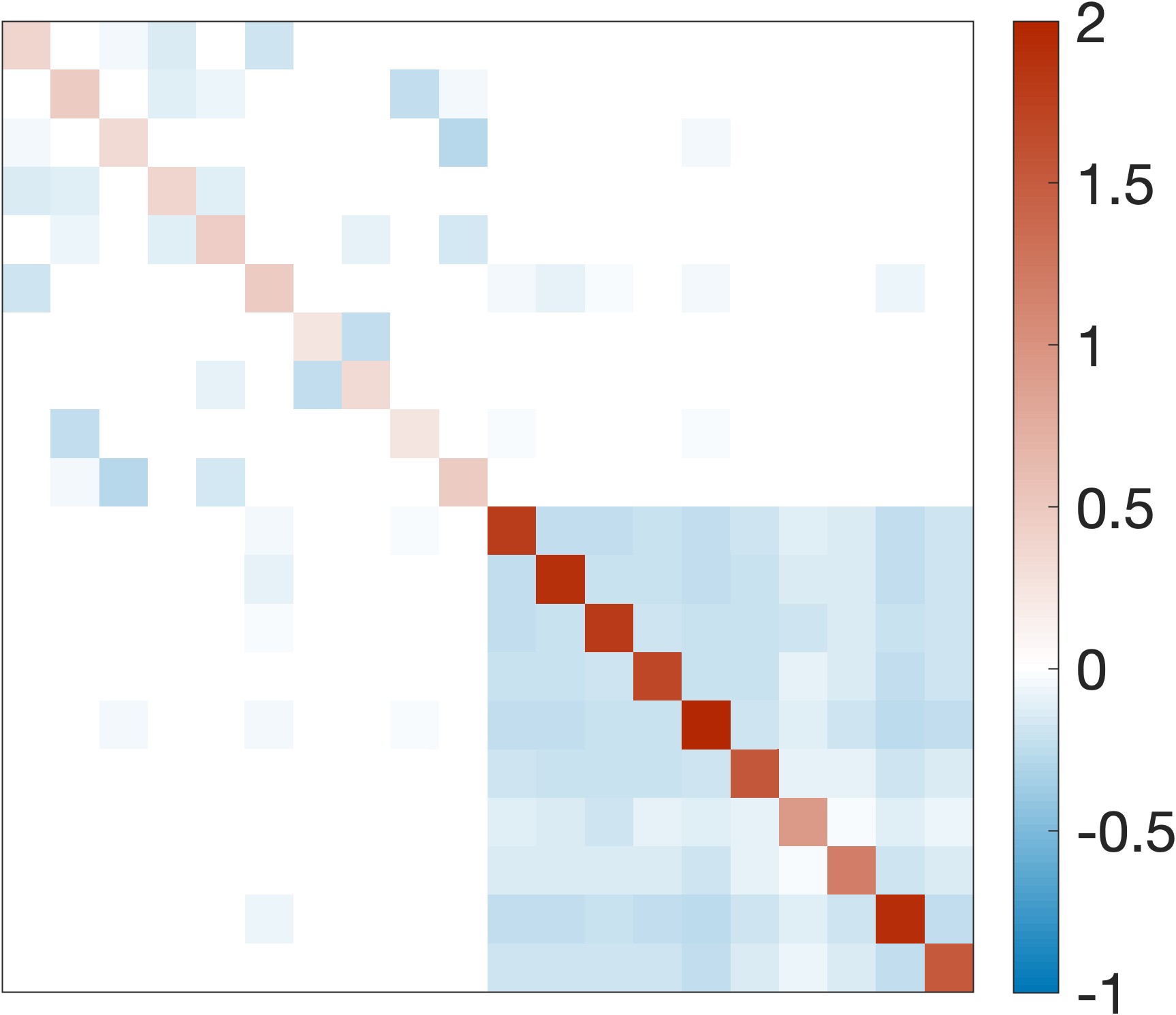} &
    \includegraphics[width=2cm]{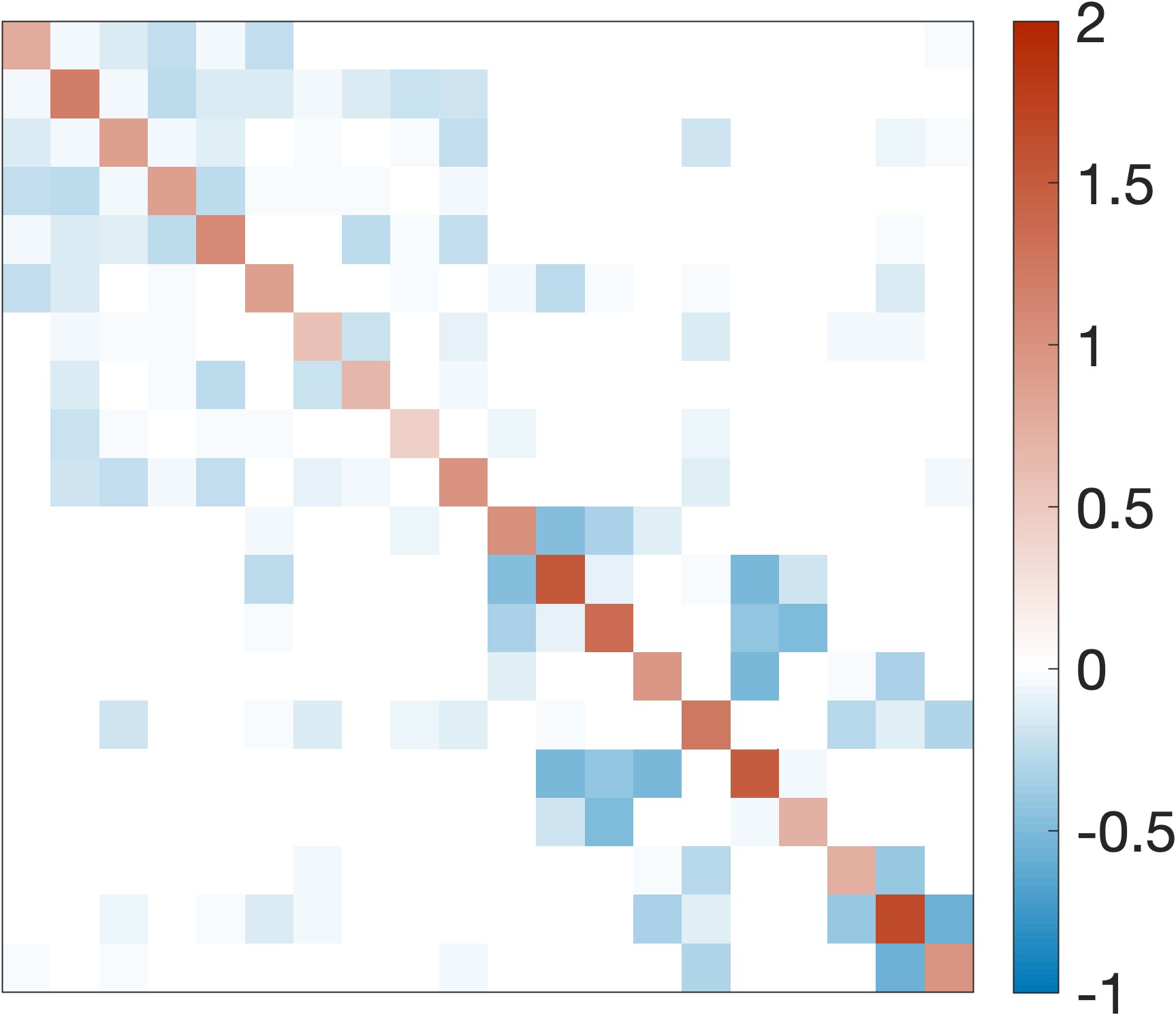}    &
    \includegraphics[width=2cm]{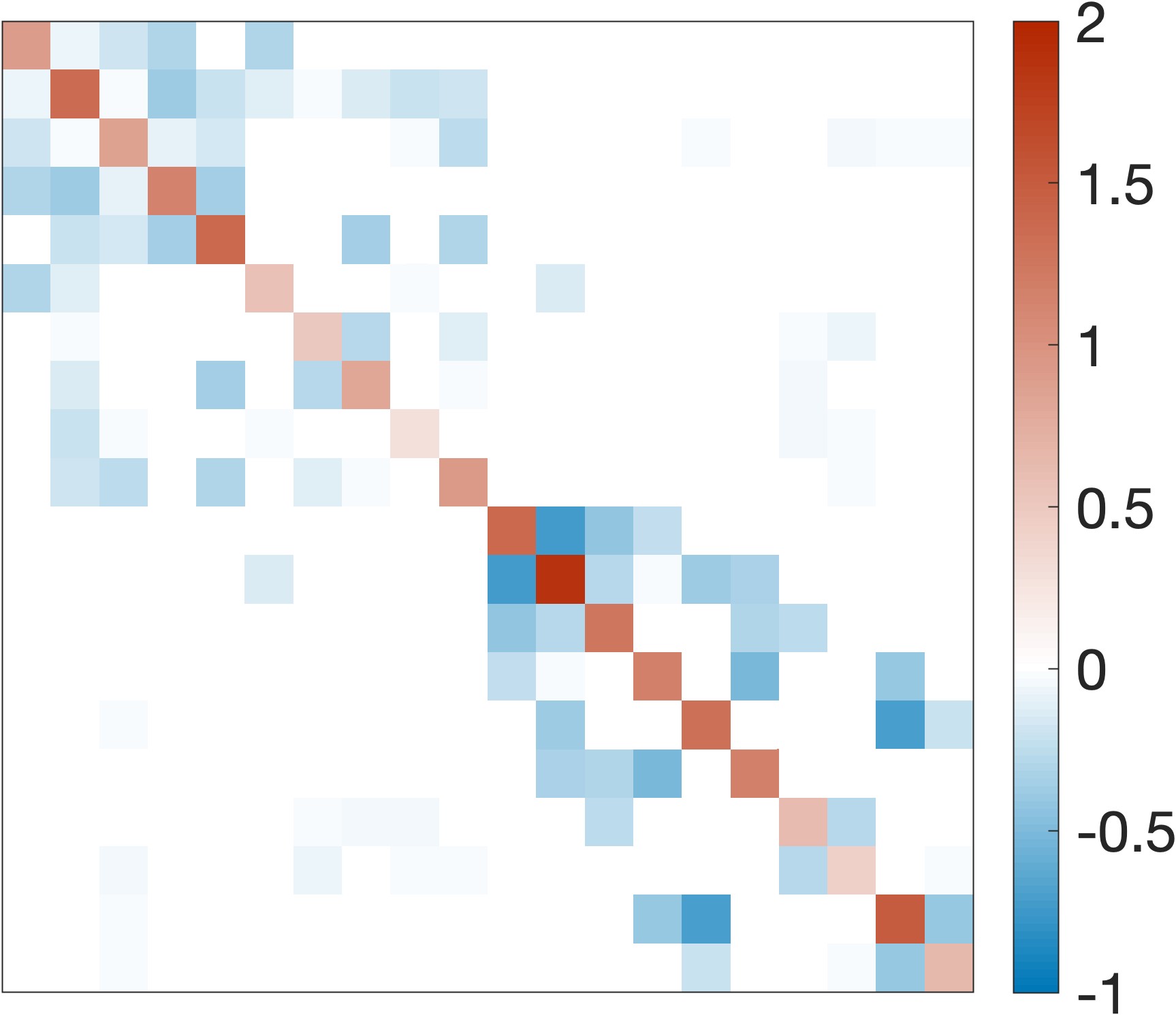}   &
    \includegraphics[width=2cm]{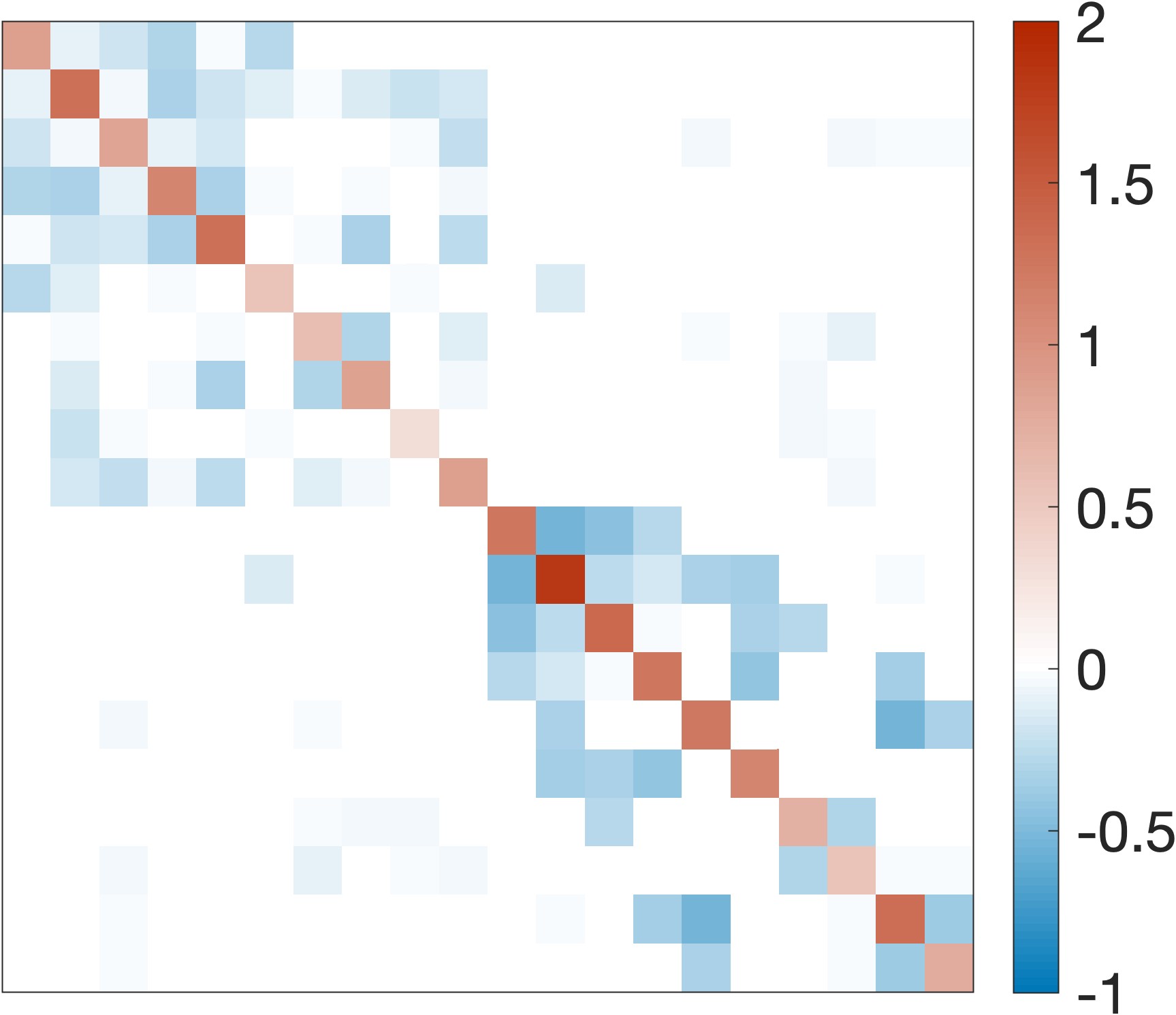} &
    \includegraphics[width=2cm]{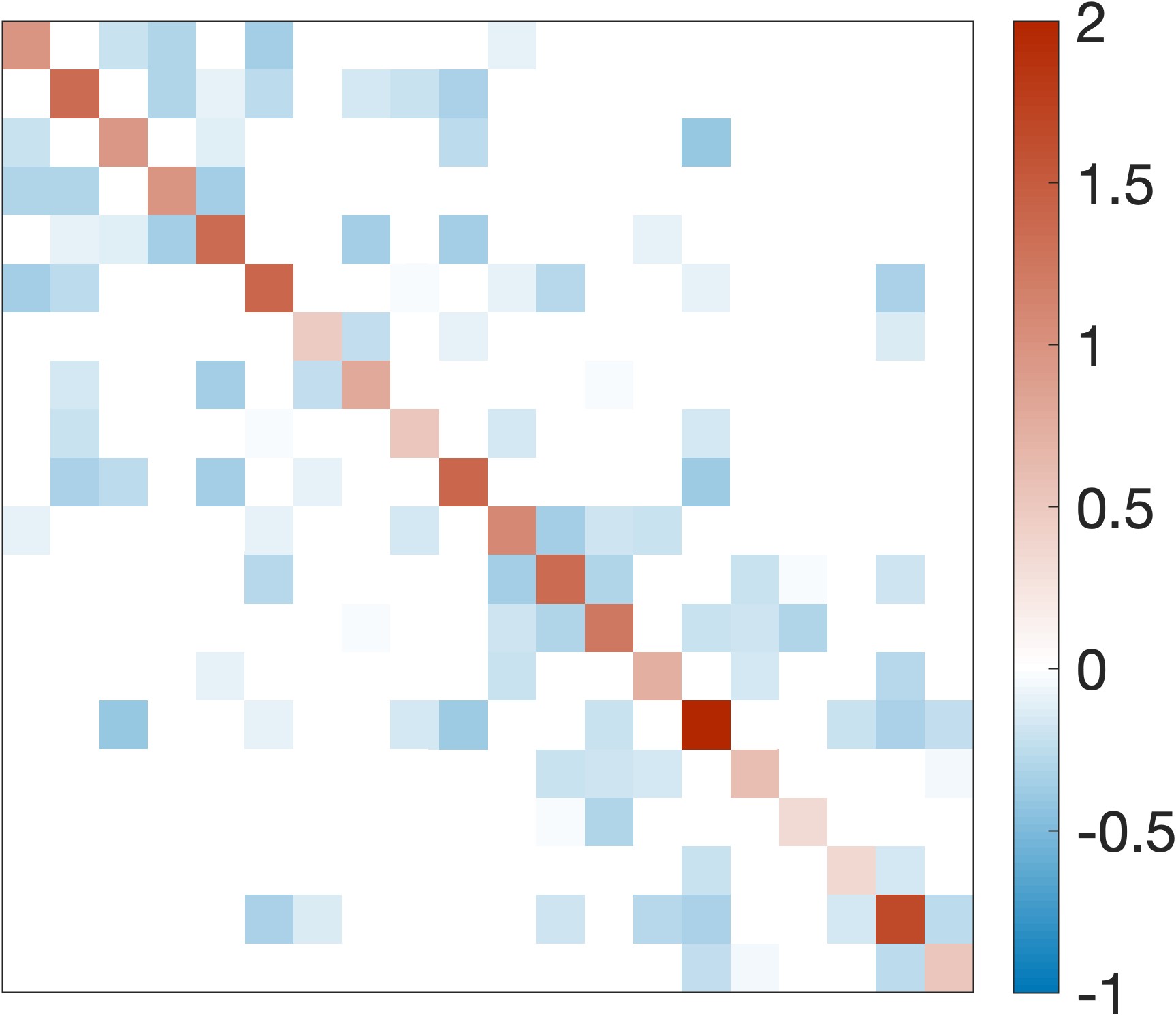}\\

    \includegraphics[width=2cm]{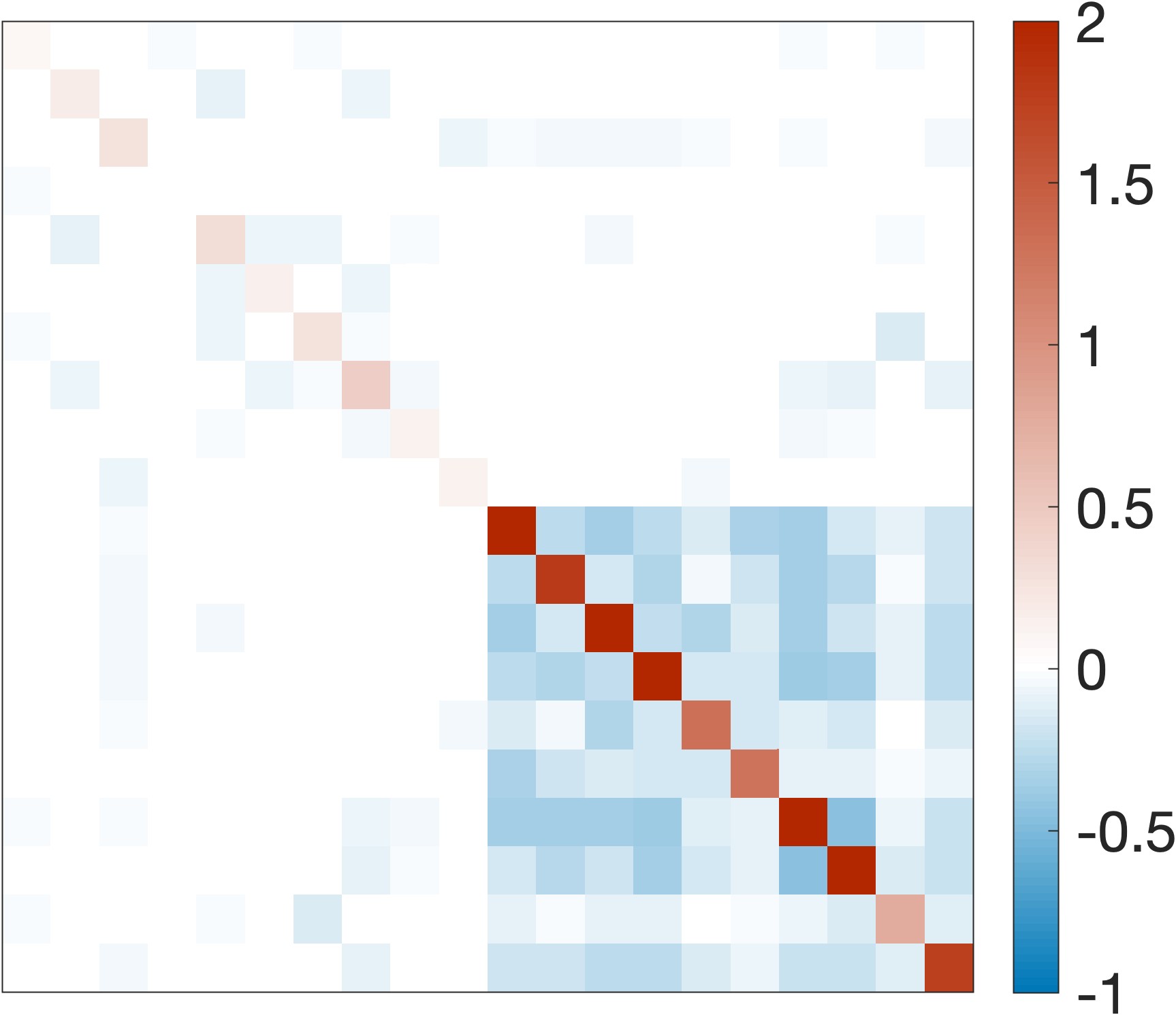}     &  \includegraphics[width=2cm]{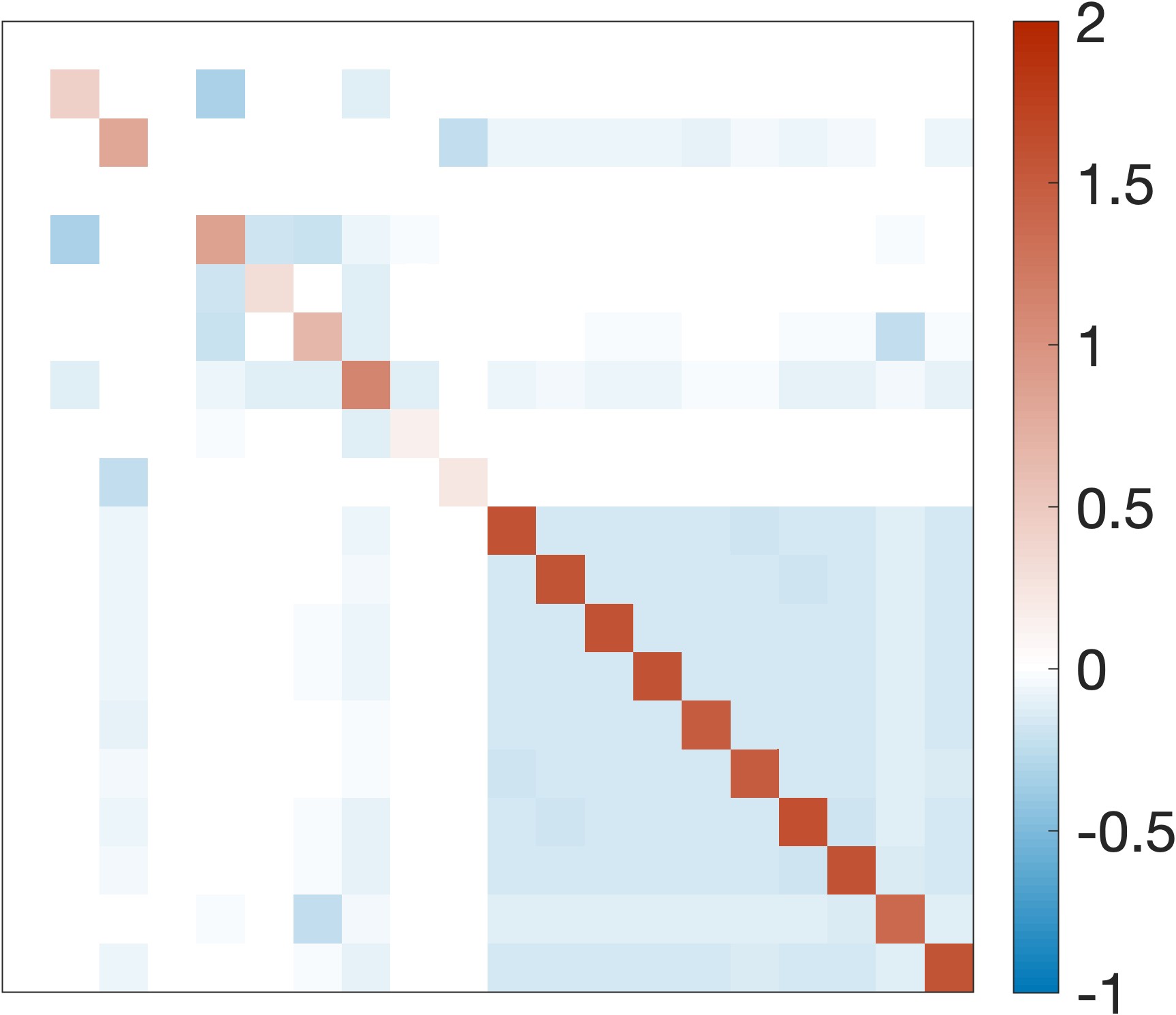}  &
    \includegraphics[width=2cm]{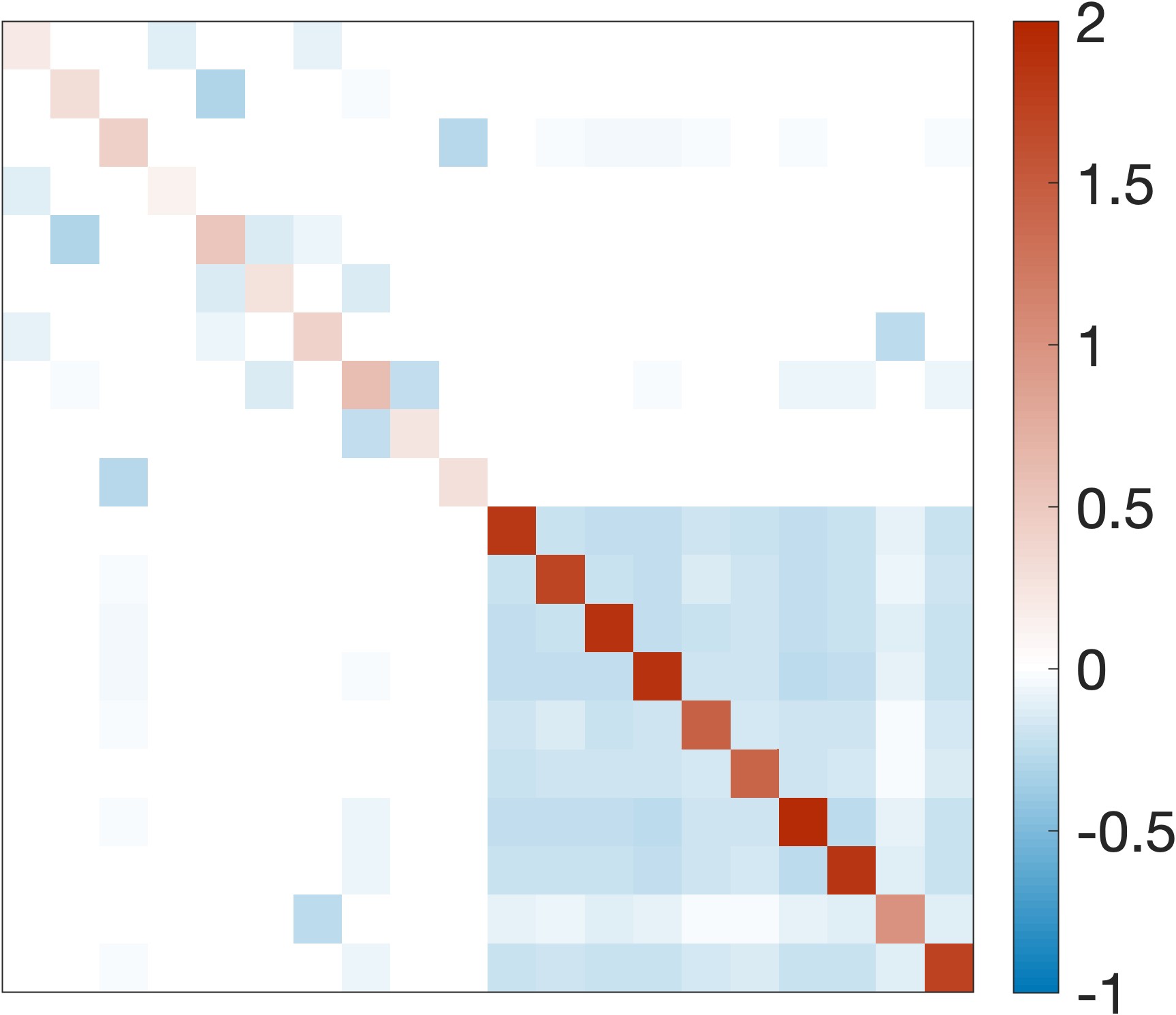} &
    \includegraphics[width=2cm]{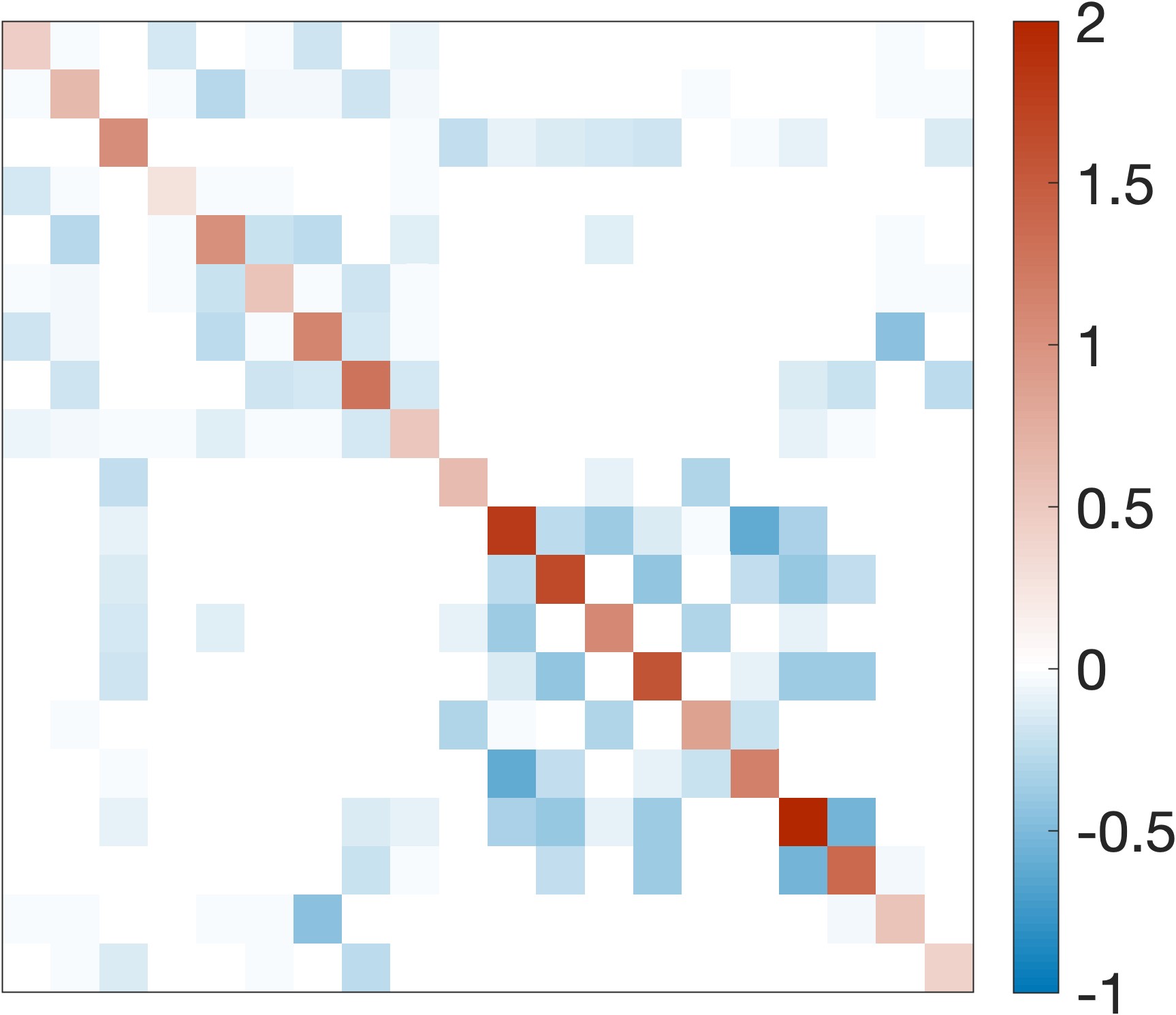}    &
    \includegraphics[width=2cm]{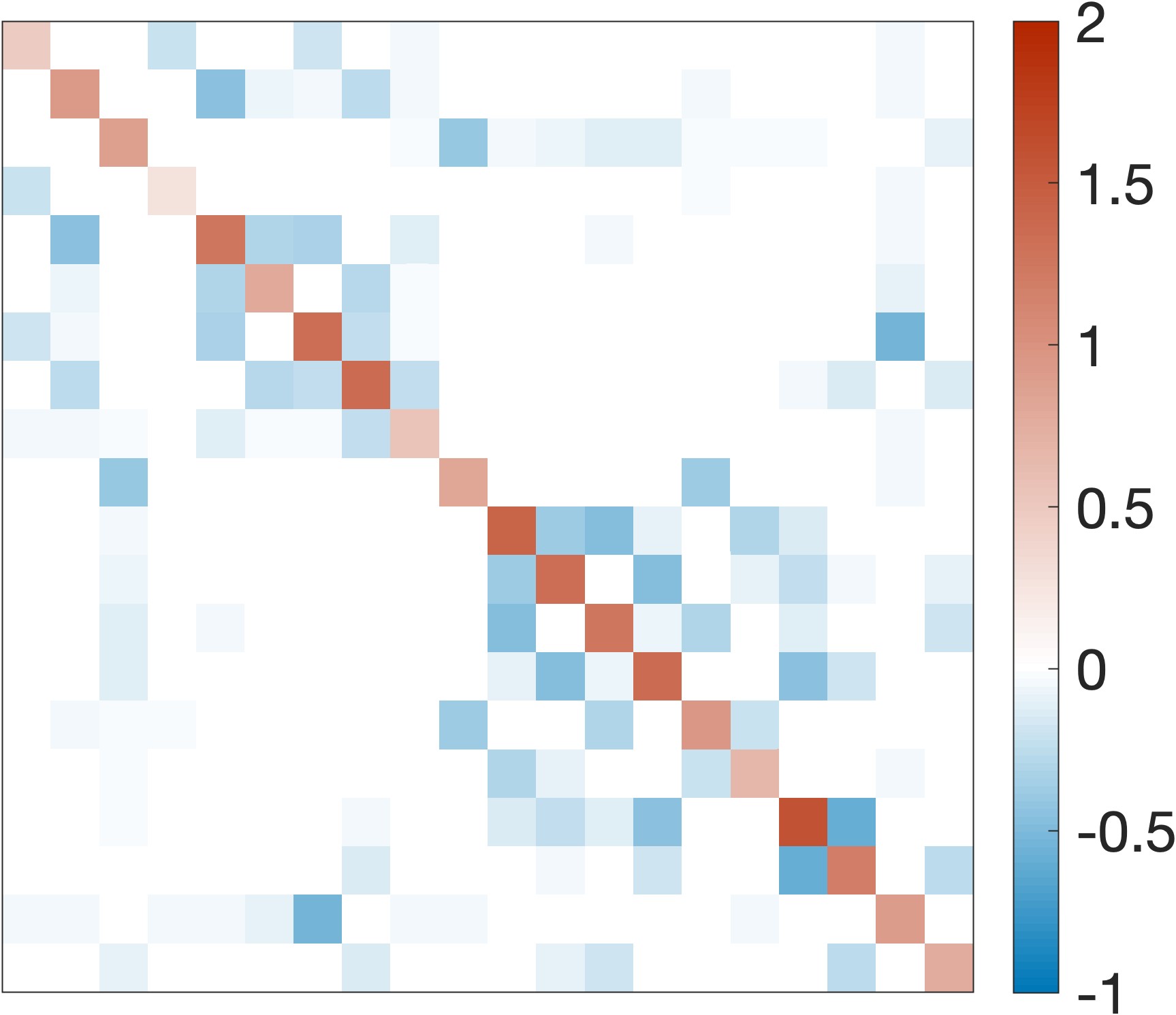}   &
    \includegraphics[width=2cm]{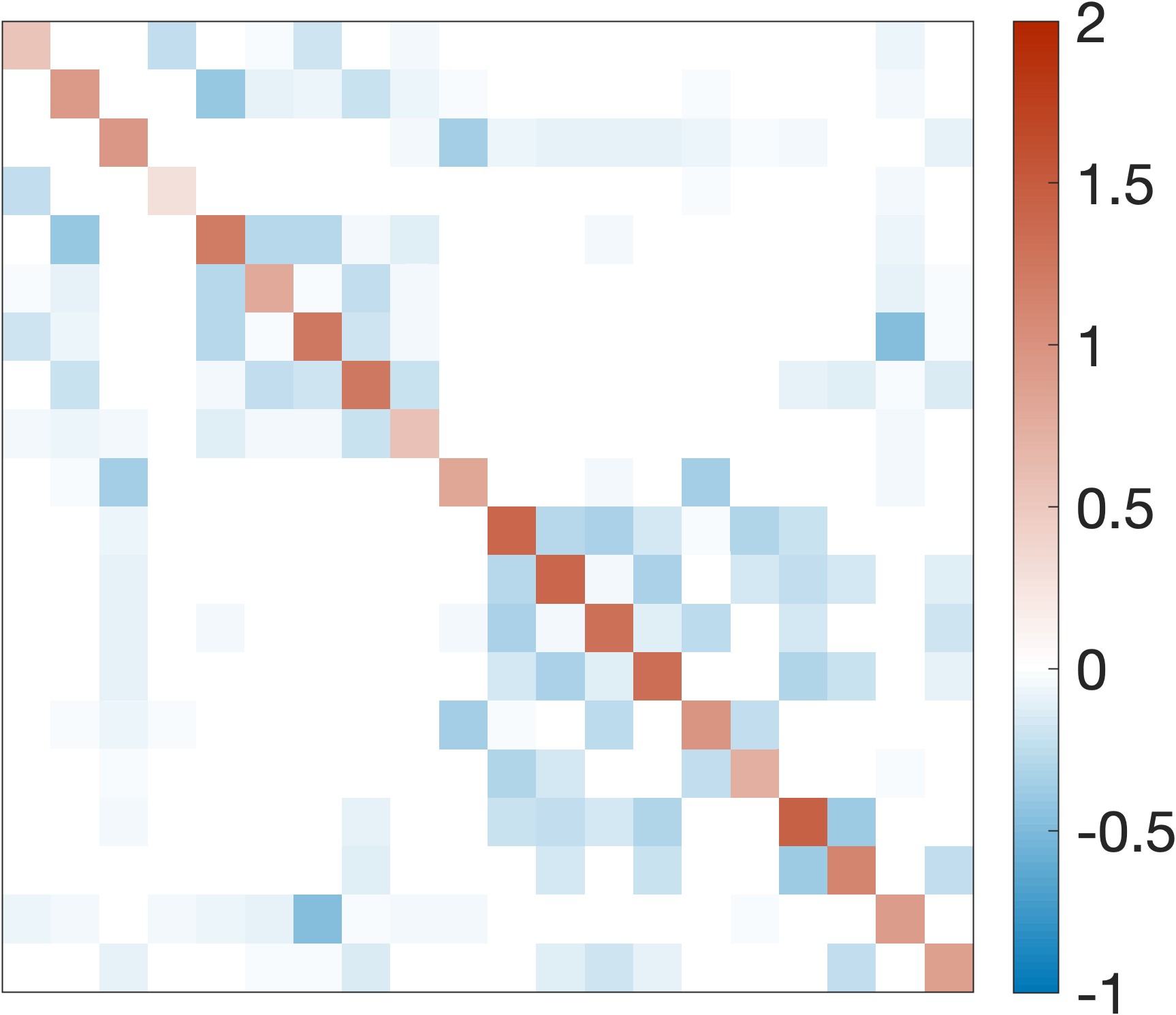} &
    \includegraphics[width=2cm]{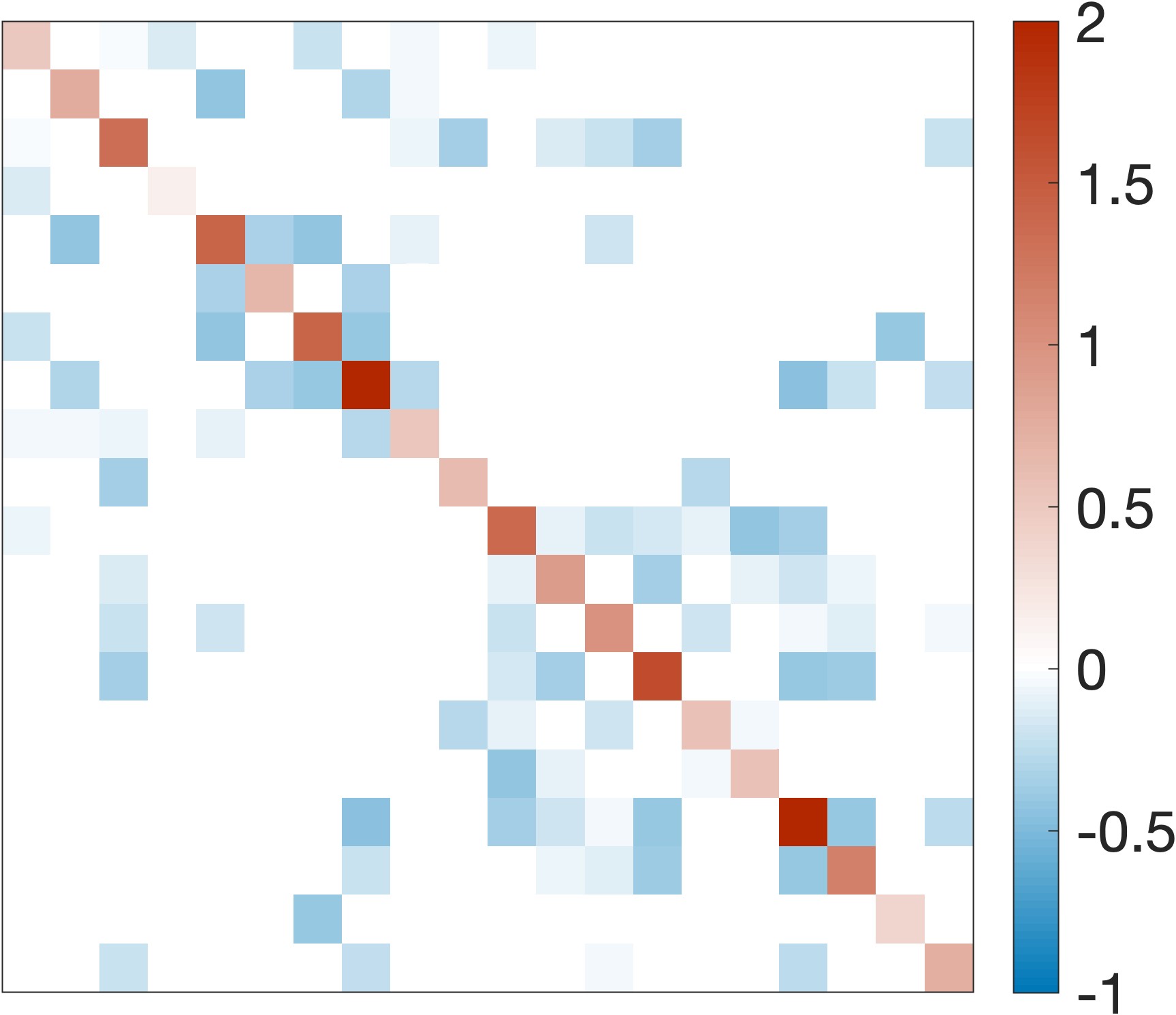}\\

    \end{tabular}
    \caption{Graph Laplacians estimated by different methods and the ground truth stochastic block graph Laplacian (p=0.4,q=0.1).}
    \label{fig:more_sbm_laplacians}

    \centering
    \begin{tabular}{cccccccc}
    SCGL    & GLS-1 & GLS-2 & CGL   & GLEN  &GLEN-VI    & Ground Truth \\

    \includegraphics[width=2cm]{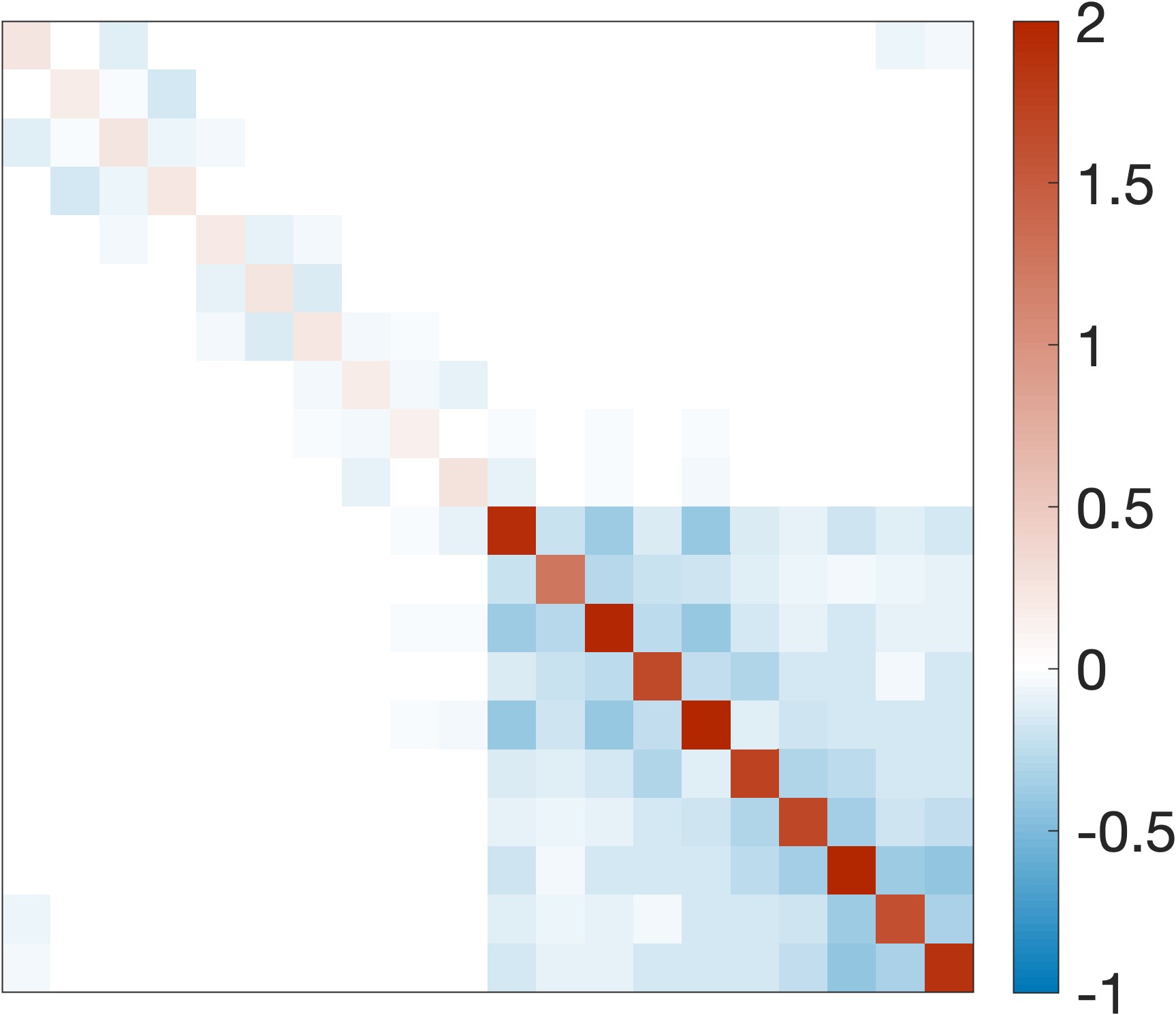}     &  \includegraphics[width=2cm]{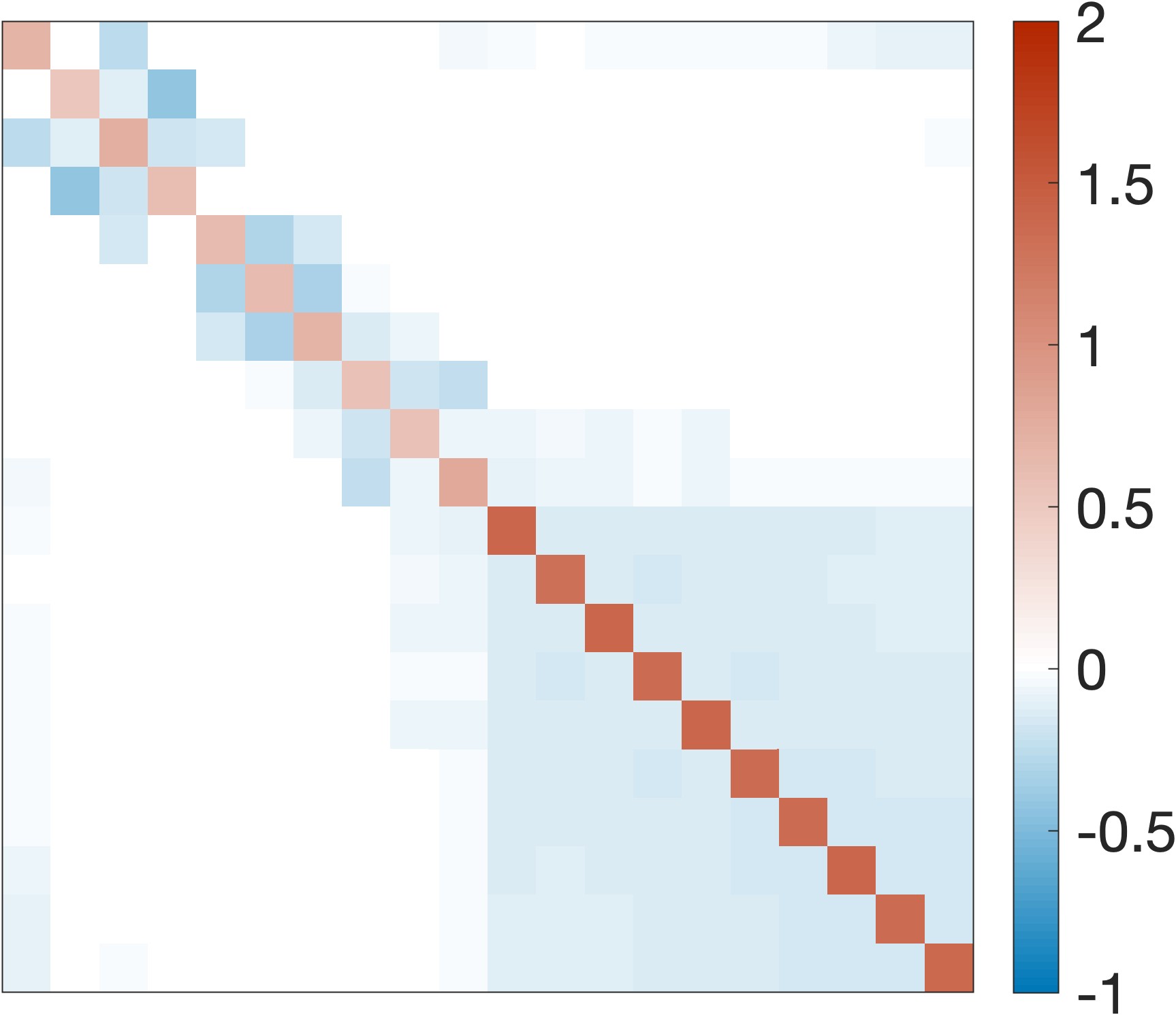}  &
    \includegraphics[width=2cm]{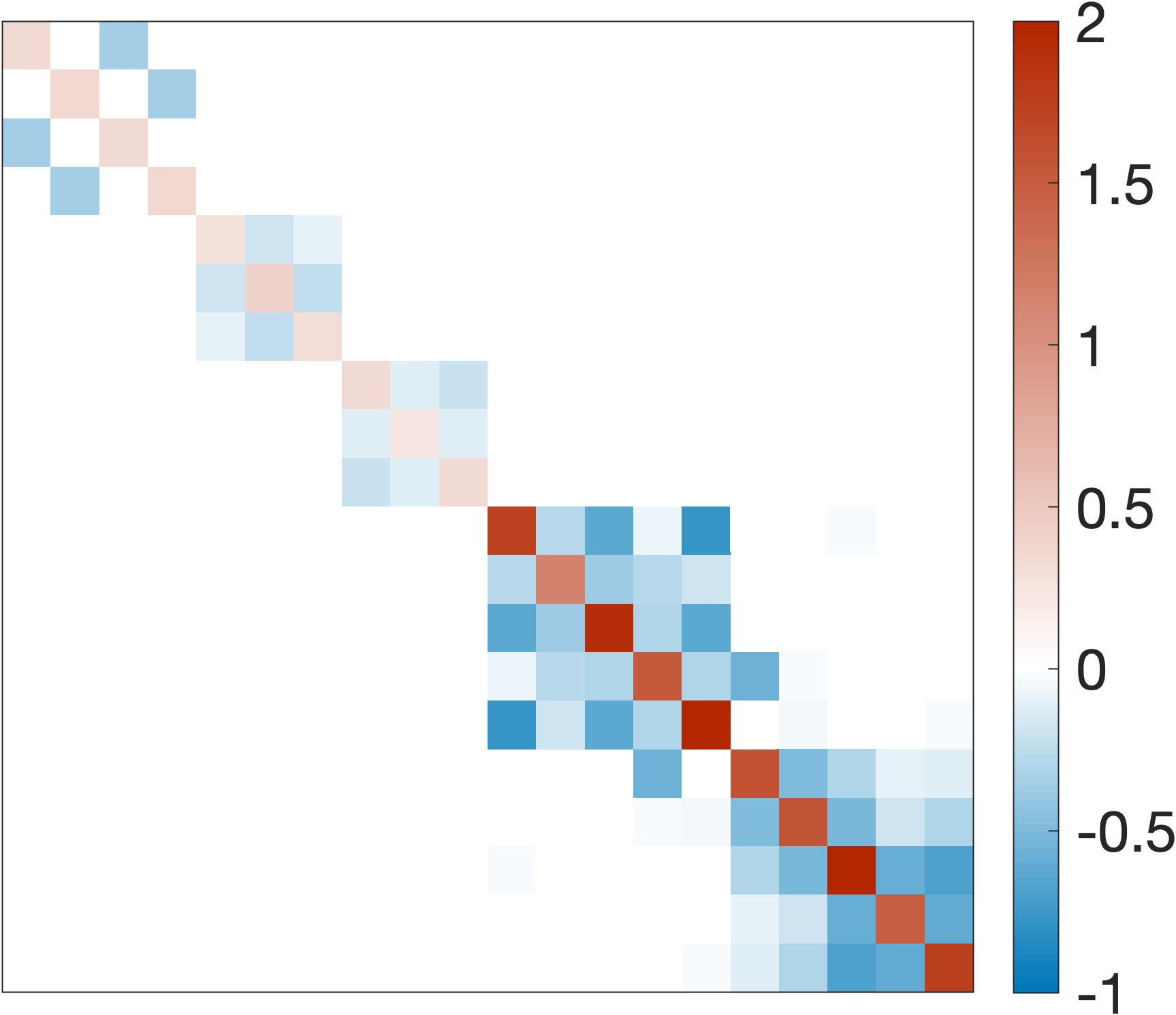} &
    \includegraphics[width=2cm]{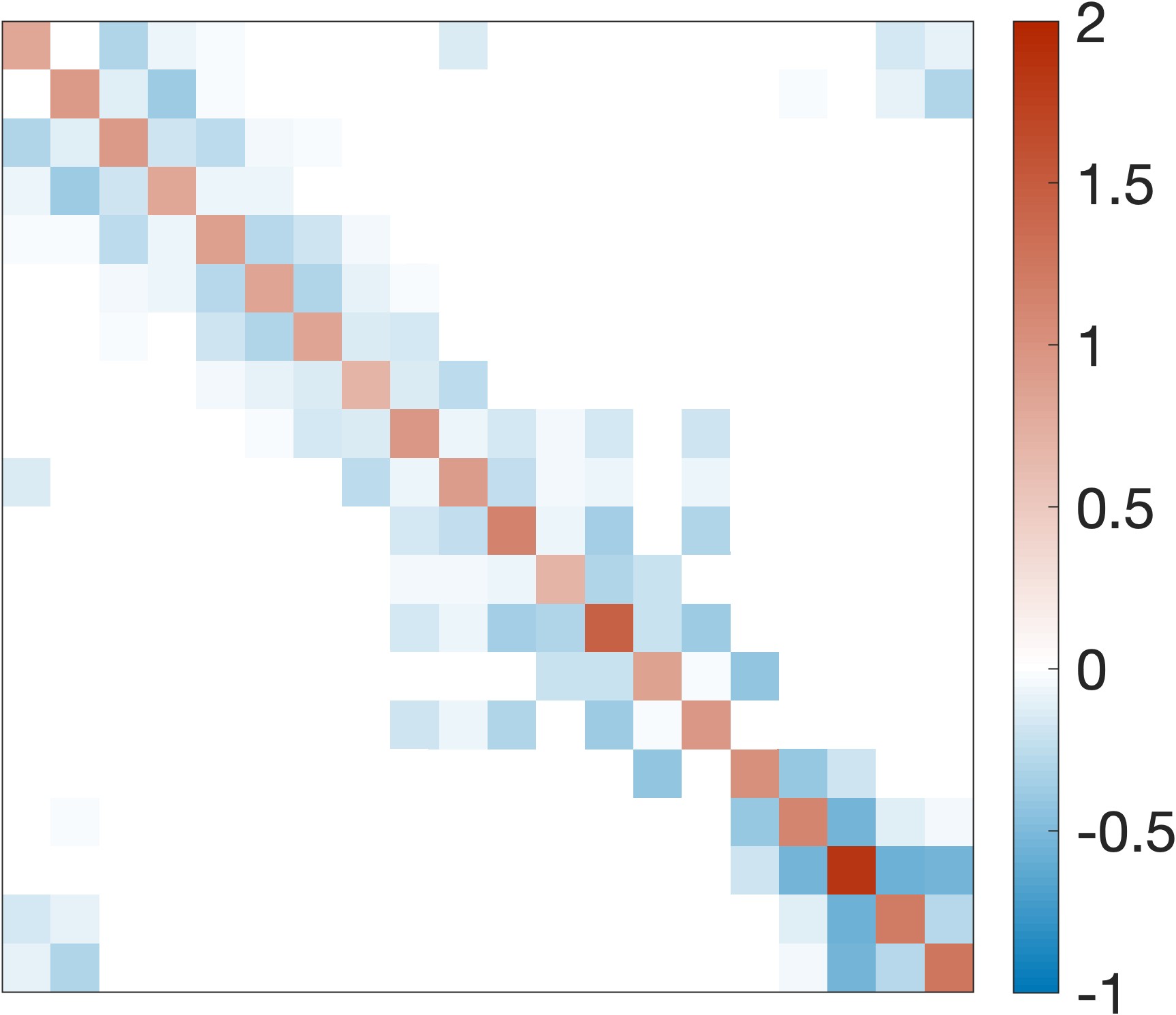}    &
    \includegraphics[width=2cm]{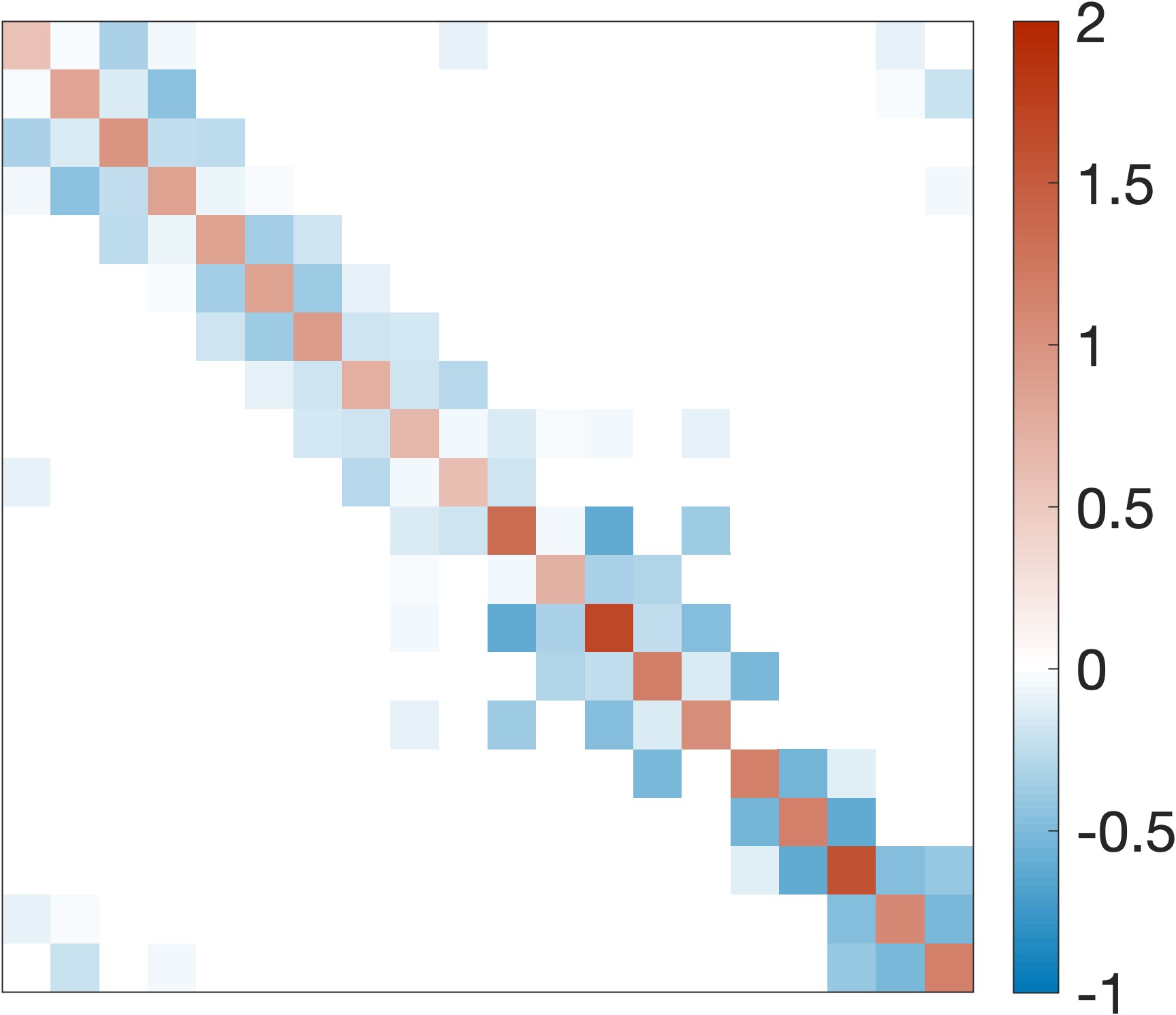}   &
    \includegraphics[width=2cm]{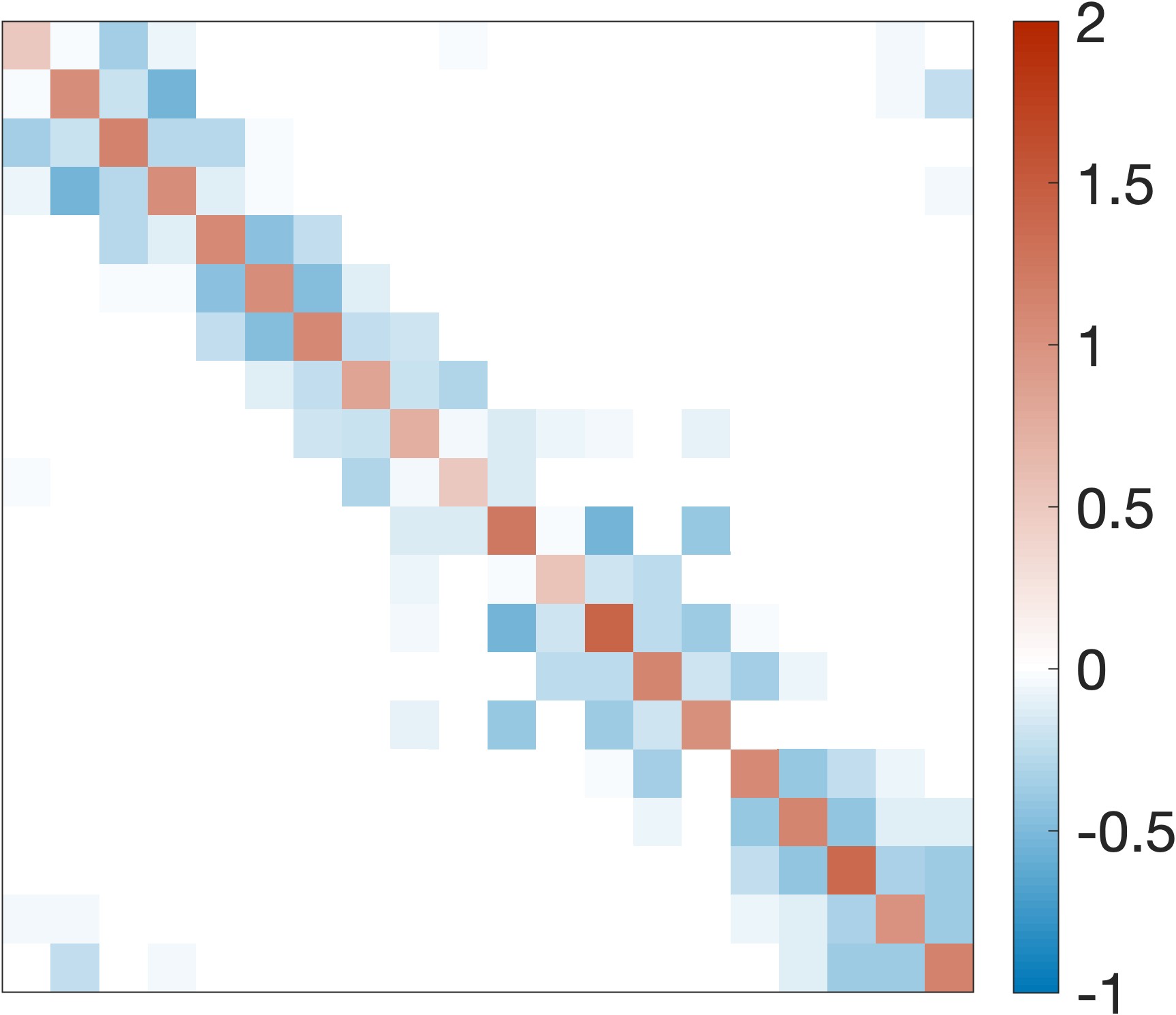} &
    \includegraphics[width=2cm]{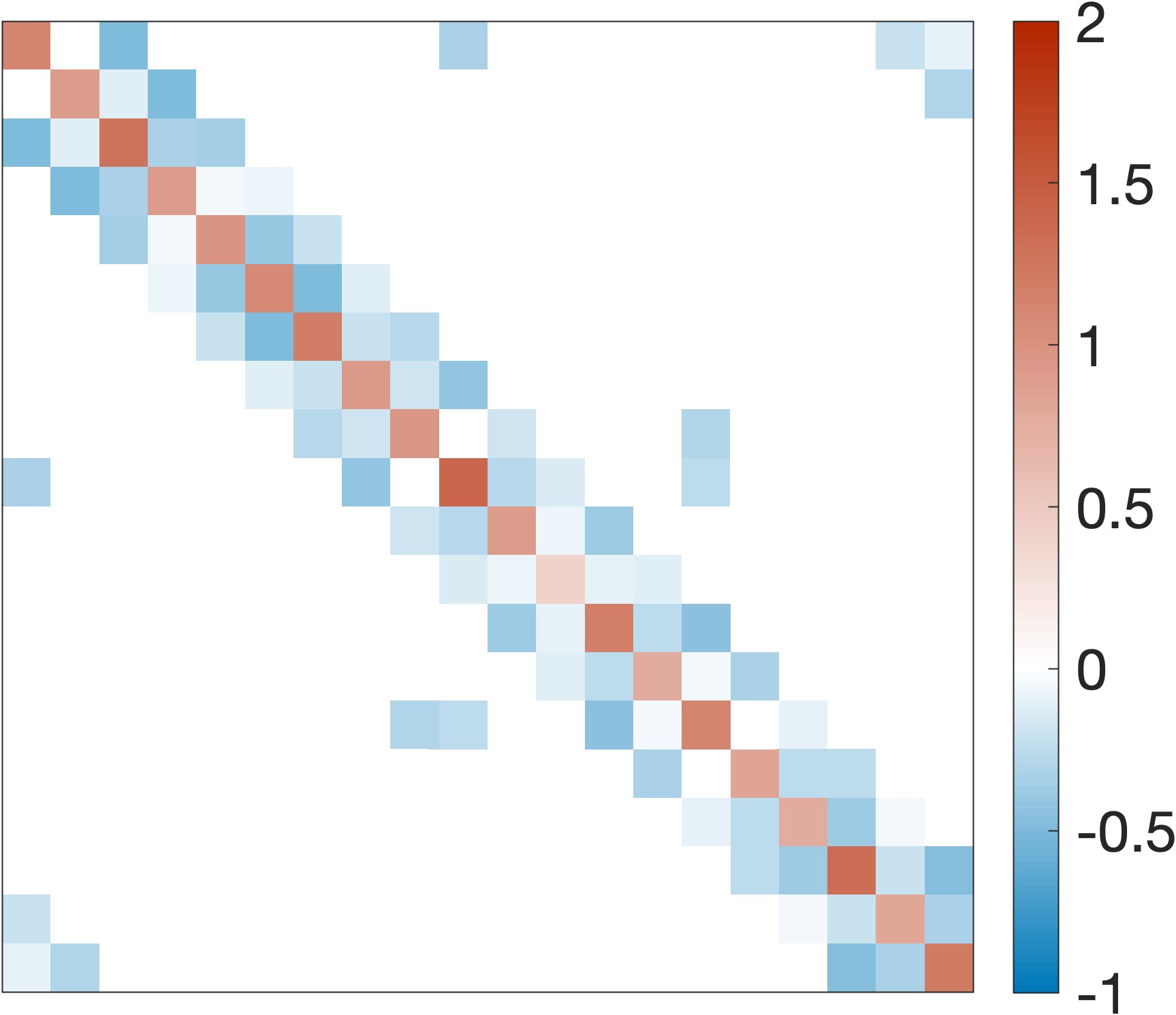}\\

    \includegraphics[width=2cm]{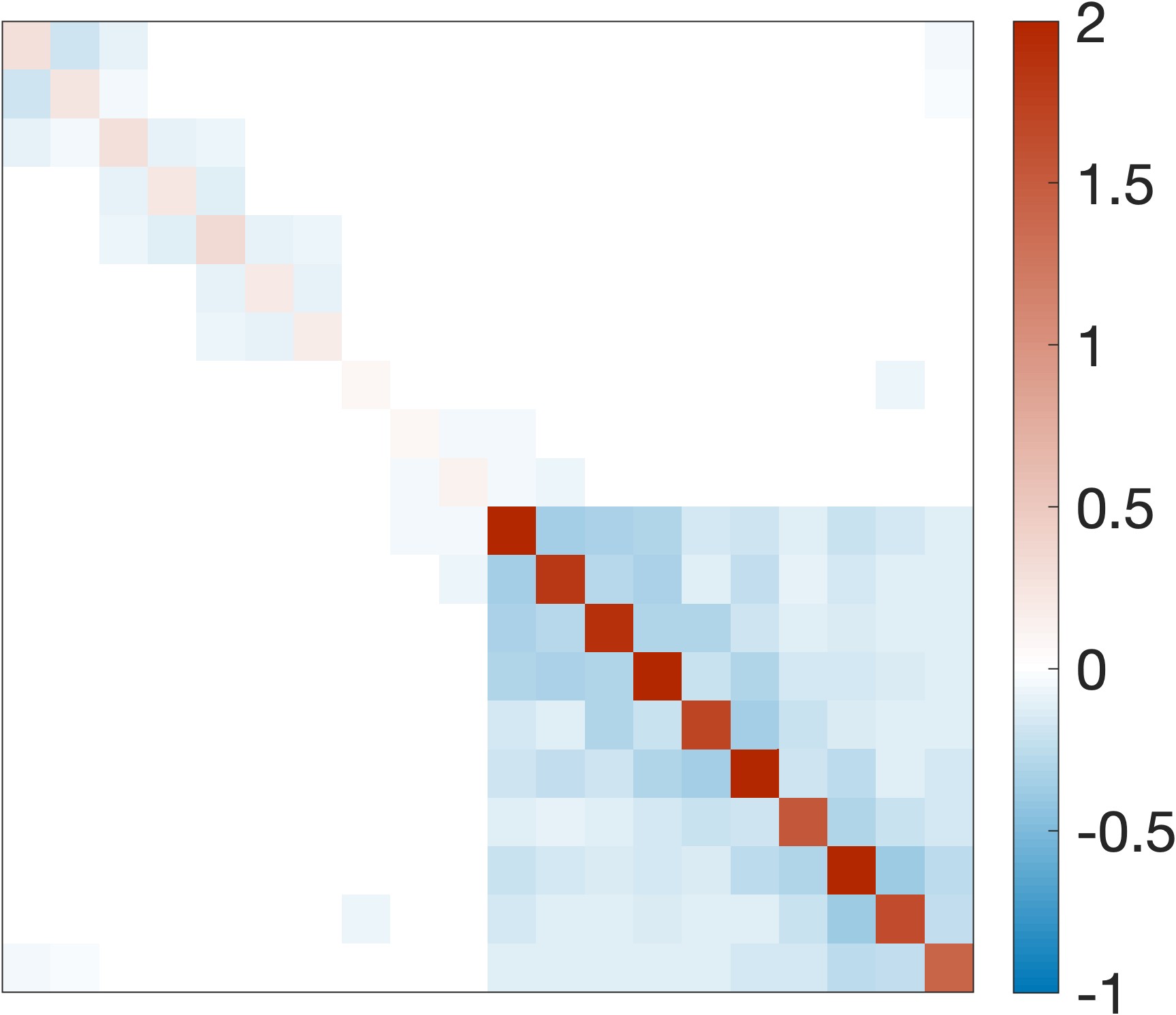}     &  \includegraphics[width=2cm]{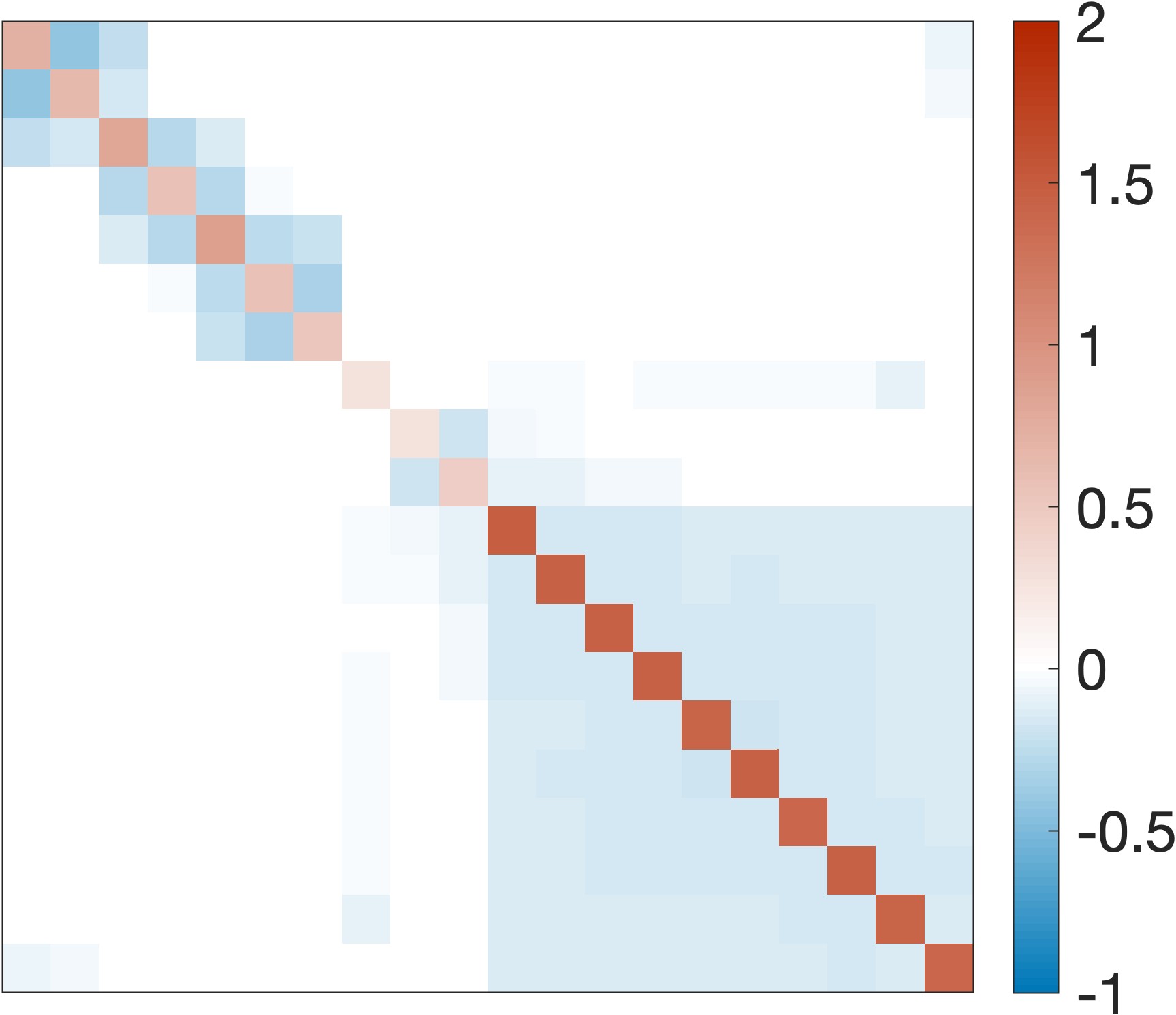}  &
    \includegraphics[width=2cm]{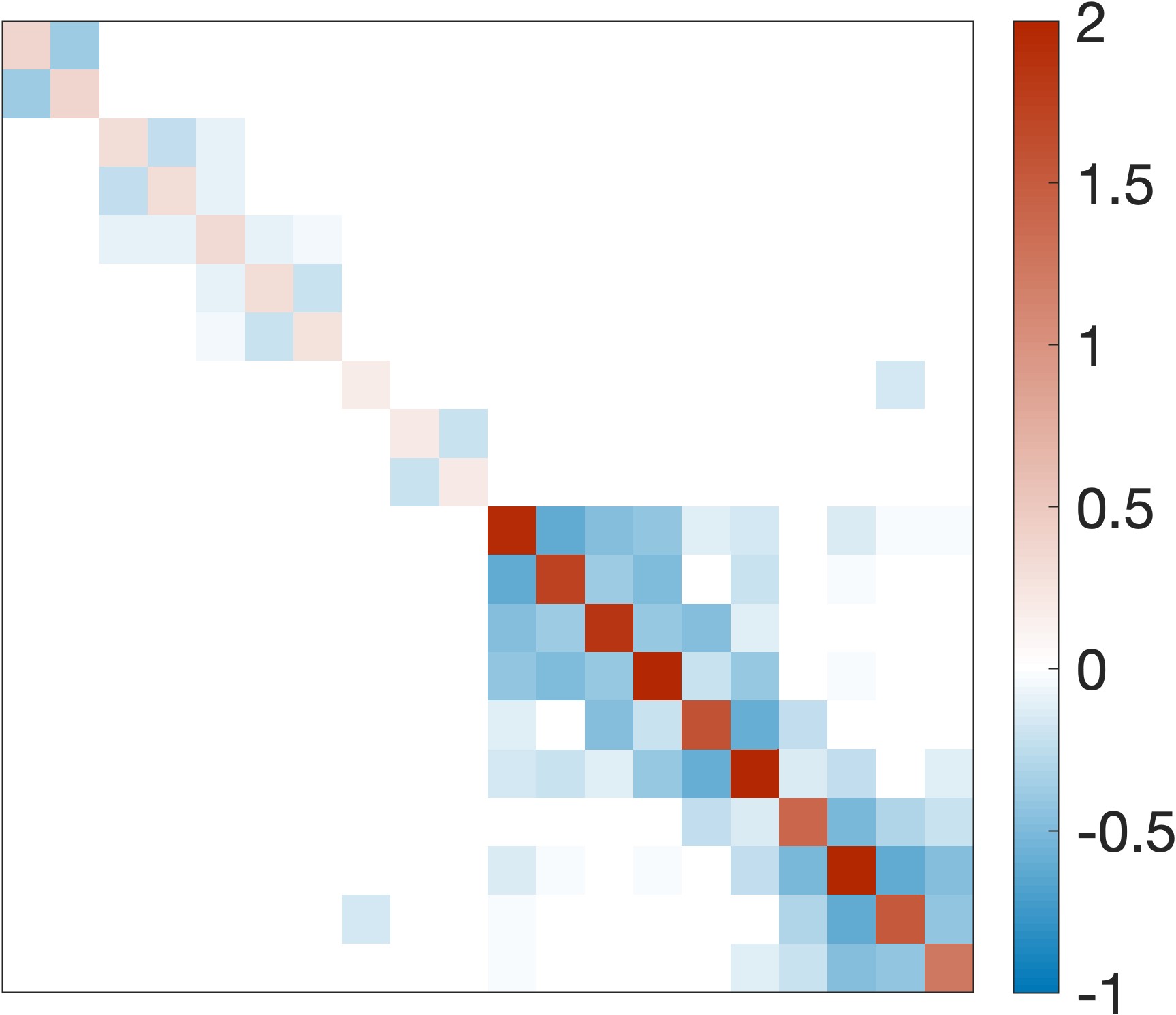} &
    \includegraphics[width=2cm]{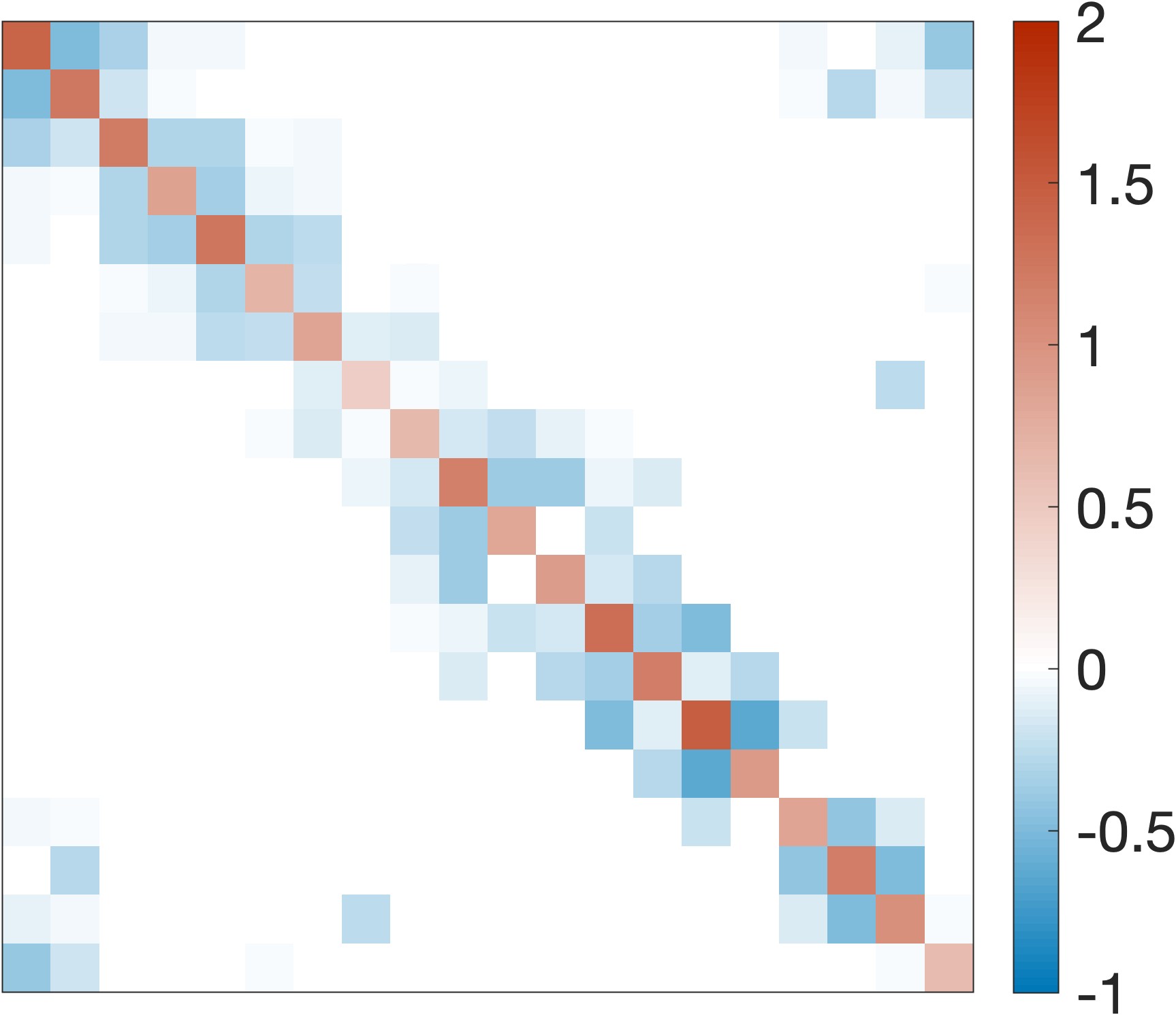}    &
    \includegraphics[width=2cm]{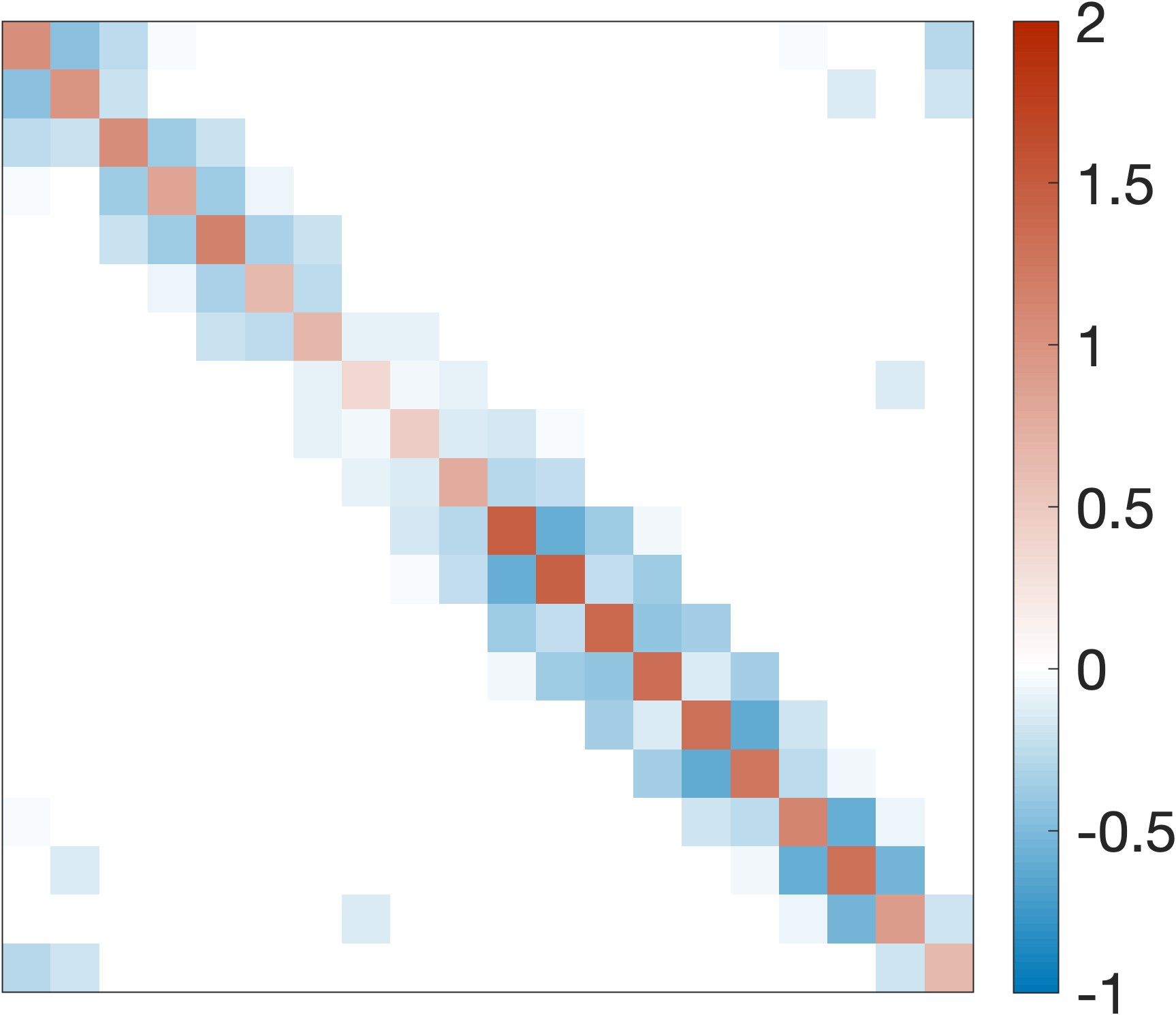}   &
    \includegraphics[width=2cm]{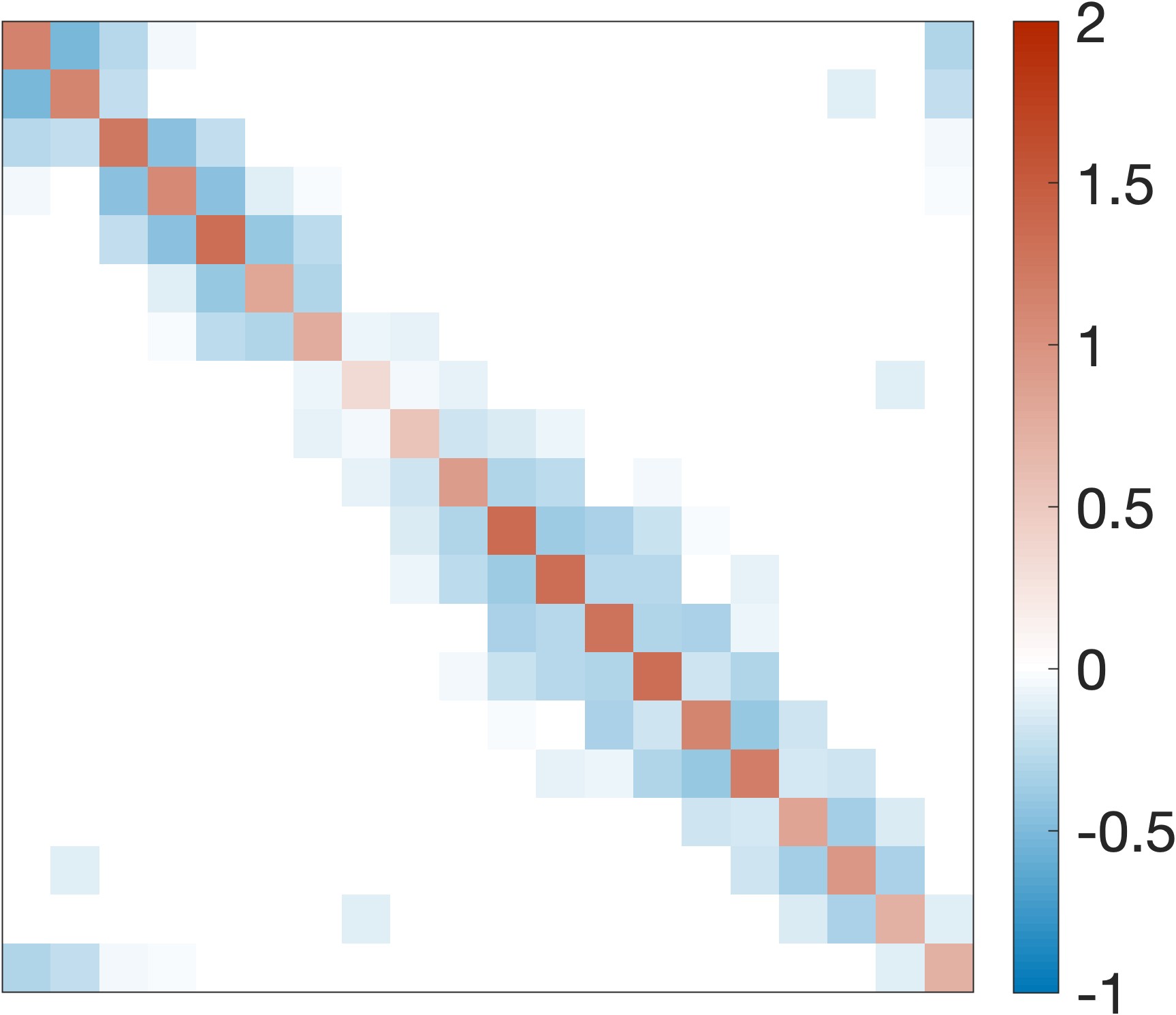} &
    \includegraphics[width=2cm]{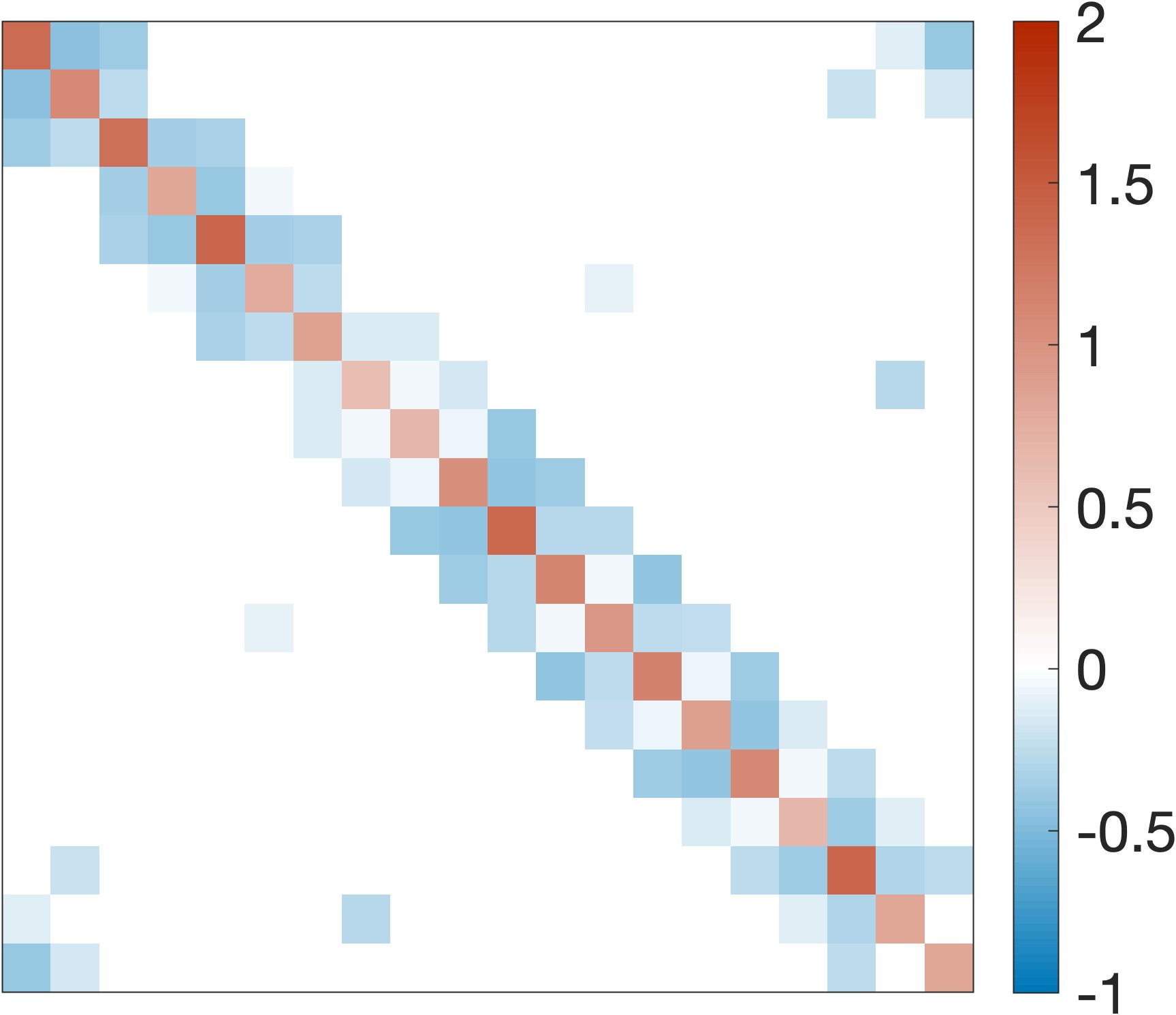}\\

    \includegraphics[width=2cm]{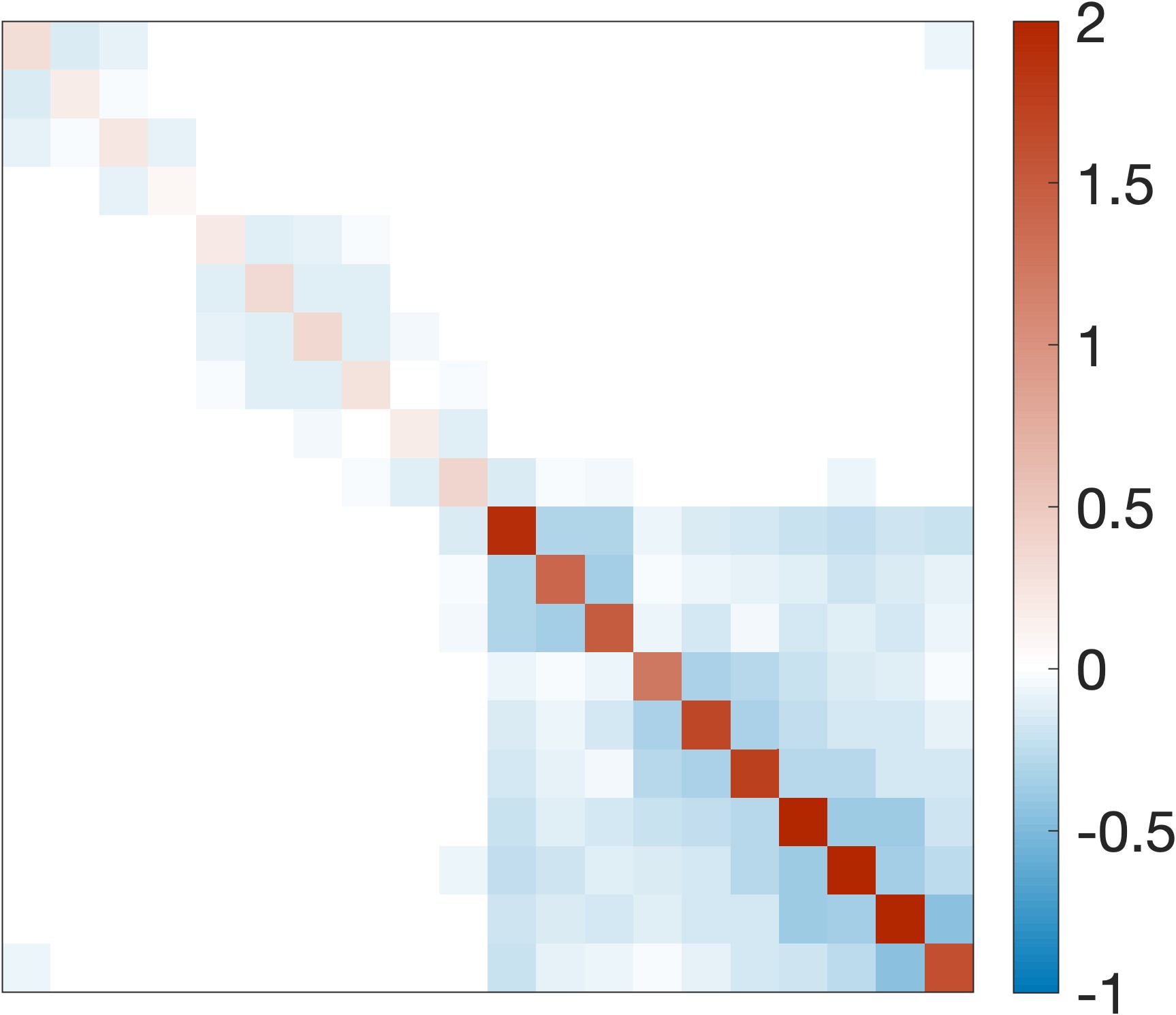}     &  \includegraphics[width=2cm]{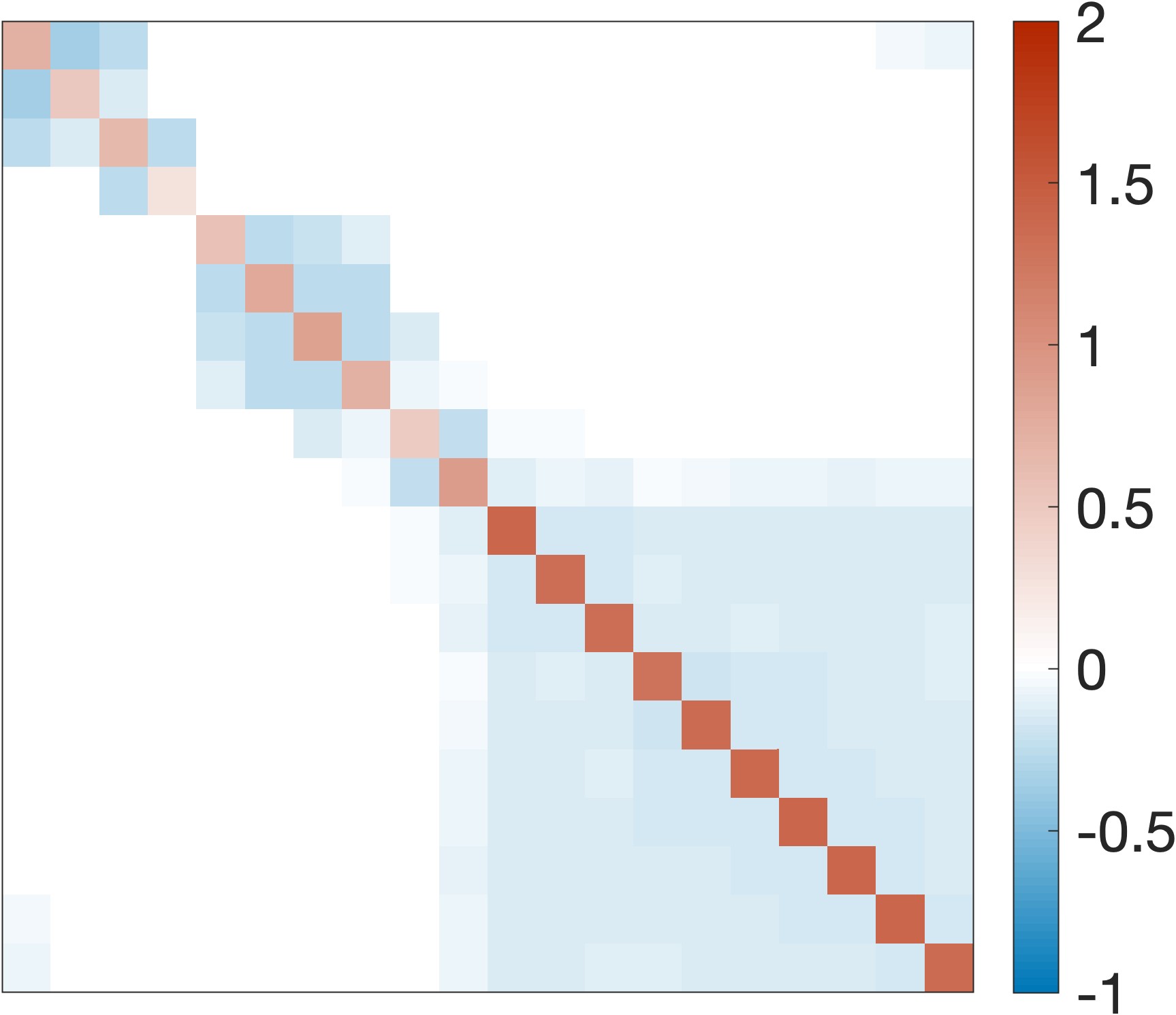}  &
    \includegraphics[width=2cm]{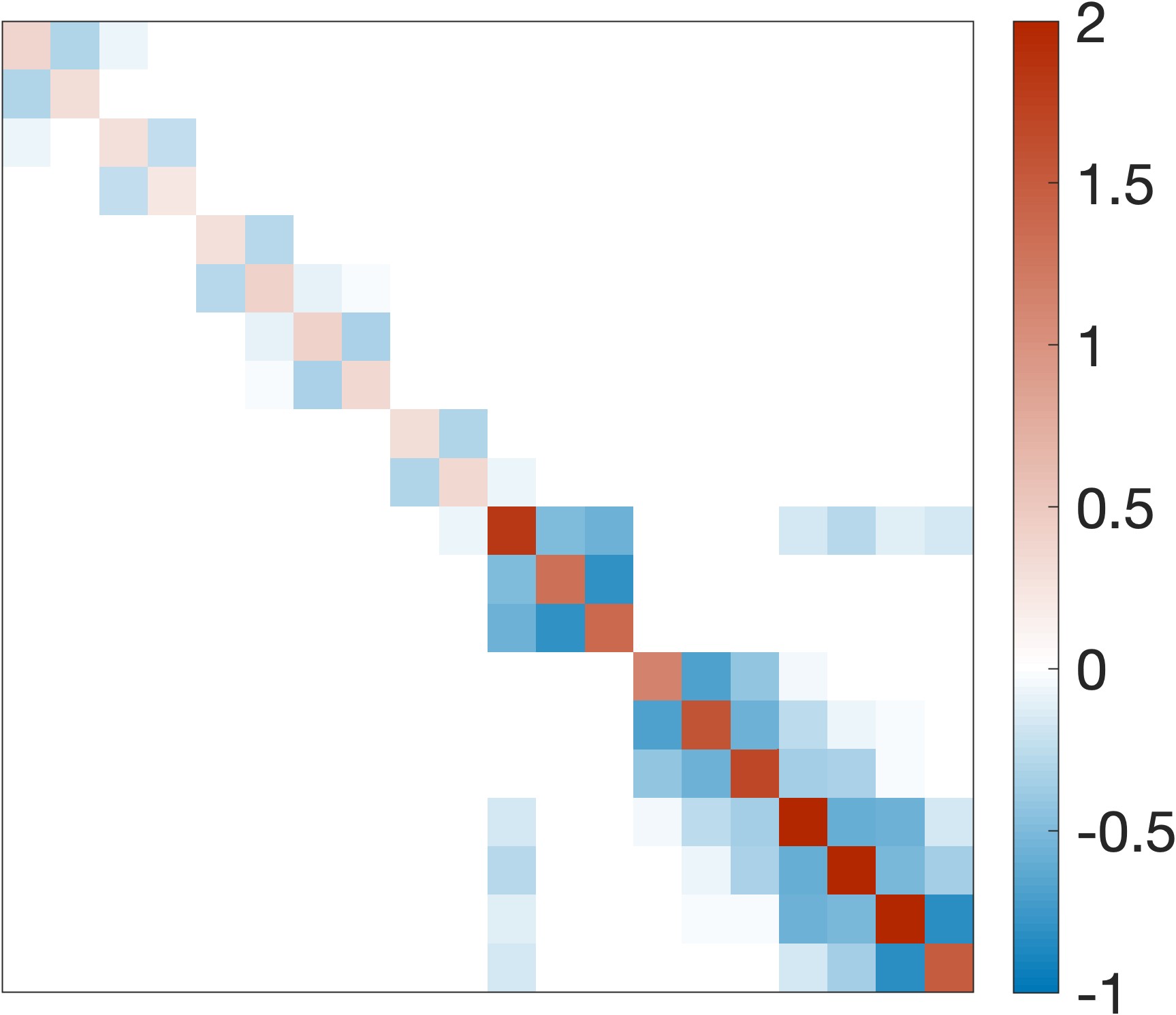} &
    \includegraphics[width=2cm]{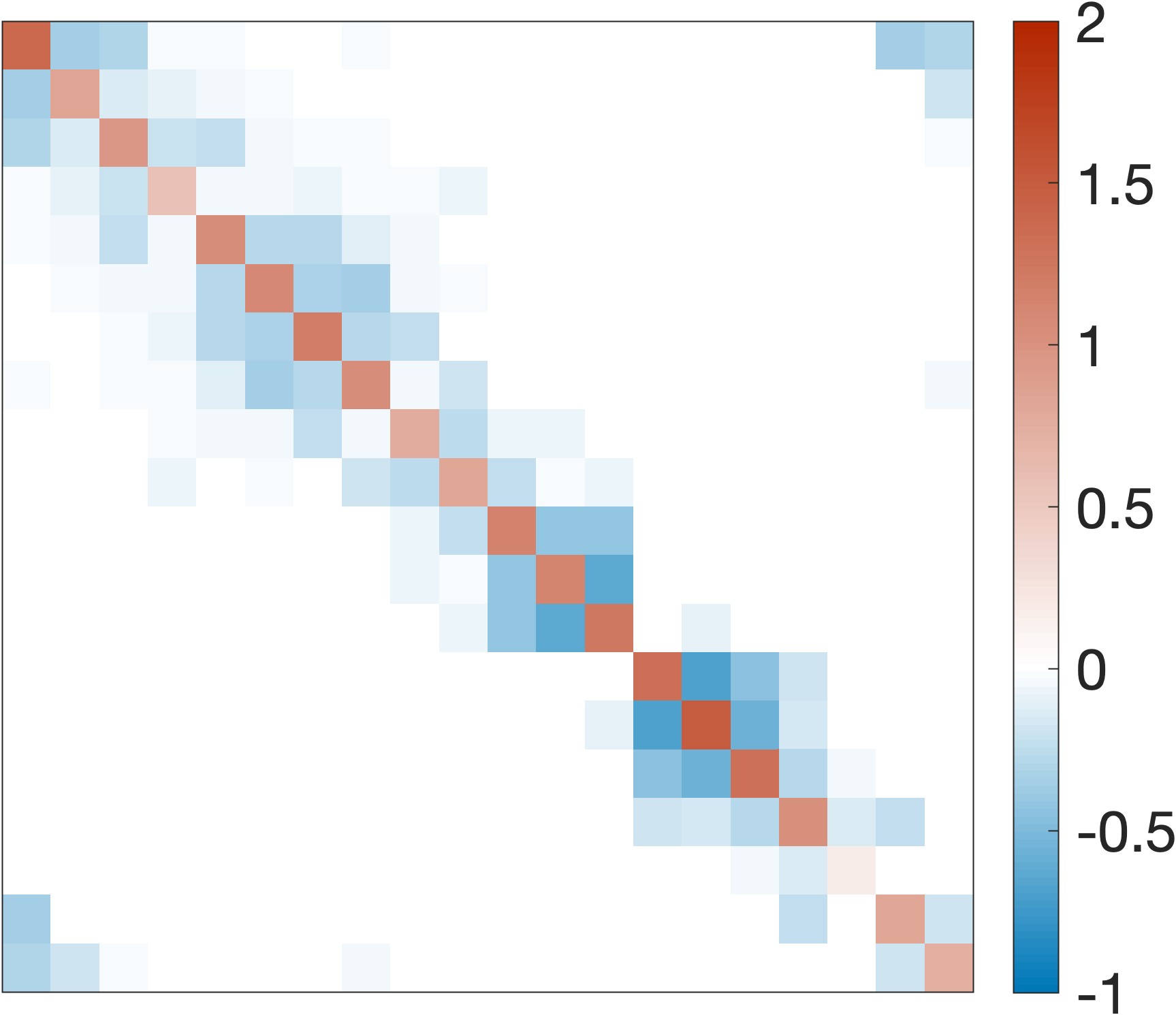}    &
    \includegraphics[width=2cm]{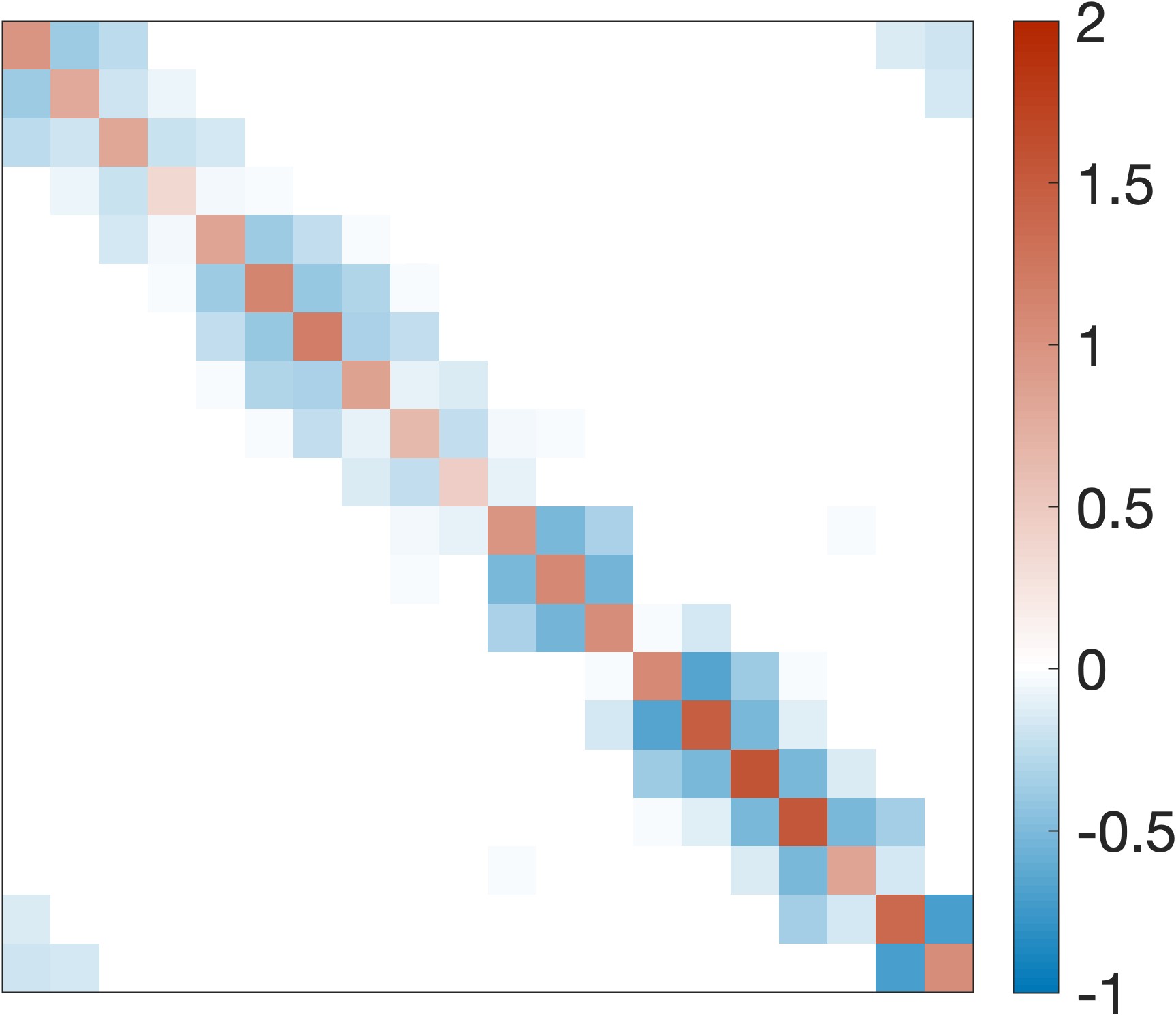}   &
    \includegraphics[width=2cm]{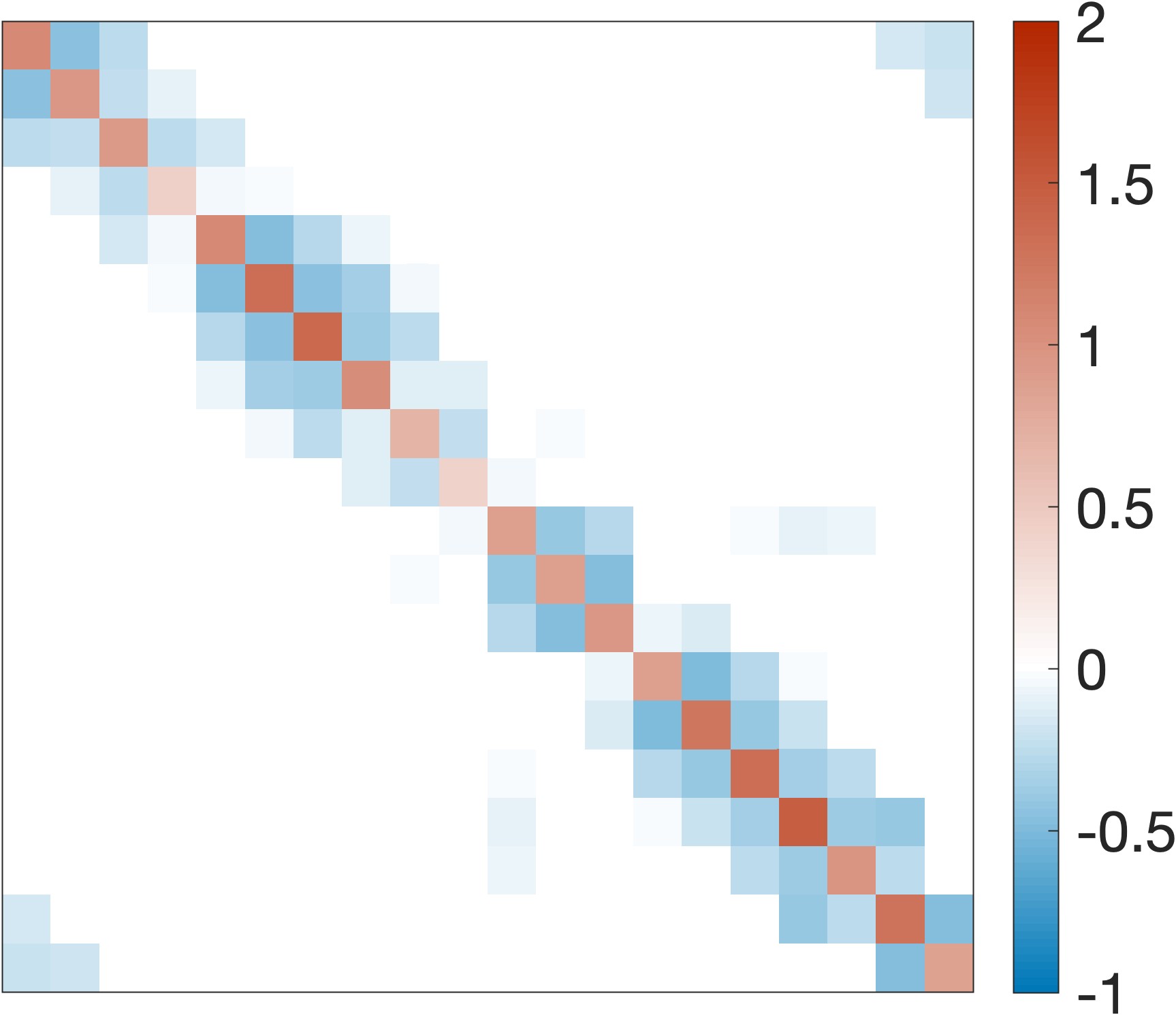} &
    \includegraphics[width=2cm]{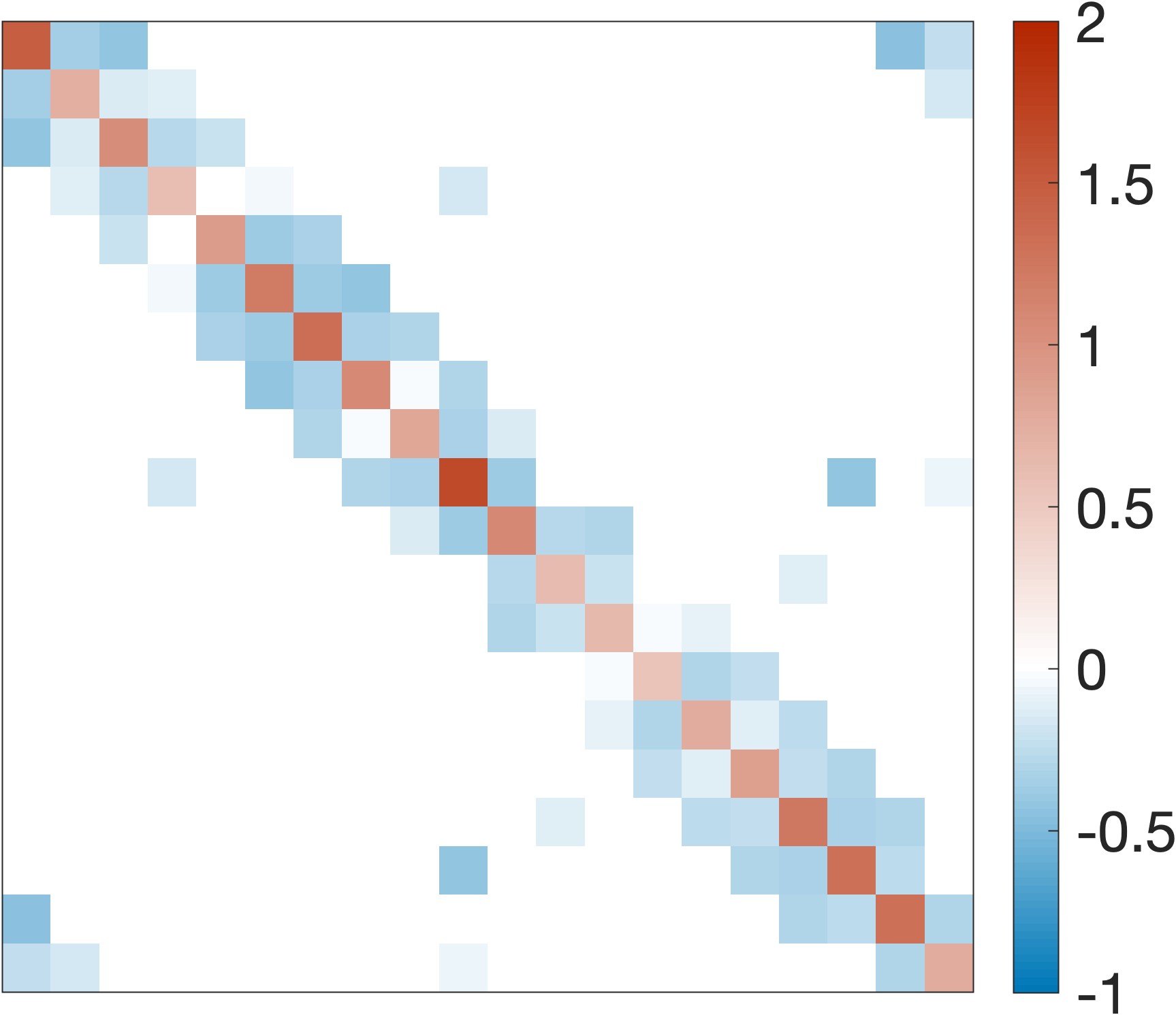}\\

    \end{tabular}
    \caption{Graph Laplacians estimated by different methods and the ground truth Watts-Strogatz small-world graph Laplacian (K=2, p=0.1).}
    \label{fig:more_ws_laplacians}
\end{figure*}

\looseness=-1
We visualize the ground truth Laplacian of several realizations as well as the inferred Laplacians of different methods in Fig.~\ref{fig:more_er_laplacians}-\ref{fig:more_ws_laplacians}.
These Laplacians are used for weight prediction.
All Laplacians are normalized by their trace.
As we can see, SCGL and both GLS methods fail to recover the structures among the second half of the nodes with negative offsets.
Surprisingly, CGL recovers those structures quite successfully even without the knowledge of Poisson distribution, but still underperforms GLEN.
Only our proposed methods obtain both accurate structure and weight estimation on all graph models.

\subsection{Chicago Crime Dataset}
\looseness=-1
Now we evaluate GLEN on a real dataset, the Chicago Crime Dataset \cite{sensorscity}.
Our goal is to learn a graph between different types of crime to reveal their patterns of concurrence.
The dataset contains 32 types of crimes that occurred in 77 Chicago communities during every hour from 2001 to 2017.
We bin the data by year over the last 10 years and aggregate the count numbers within each bin, resulting in 770 graph signals over 32 crime types.
We further remove ``Ritualism" from the crimes since there is no occurrence in the 10 years period, resulting in 31 nodes.
We apply CGL and GLEN, still with Poisson noise, to this $31\times770$ count matrix.
Evaluating on the Chicago crime dataset demonstrates the importance of considering non-zero offsets, since the average frequency of different types of crime can be very different, and we want to be invariant to that frequency.

\begin{figure}[htb]
\begin{minipage}[b]{0.48\linewidth}
  \centering
  \centerline{\includegraphics[width=4cm]{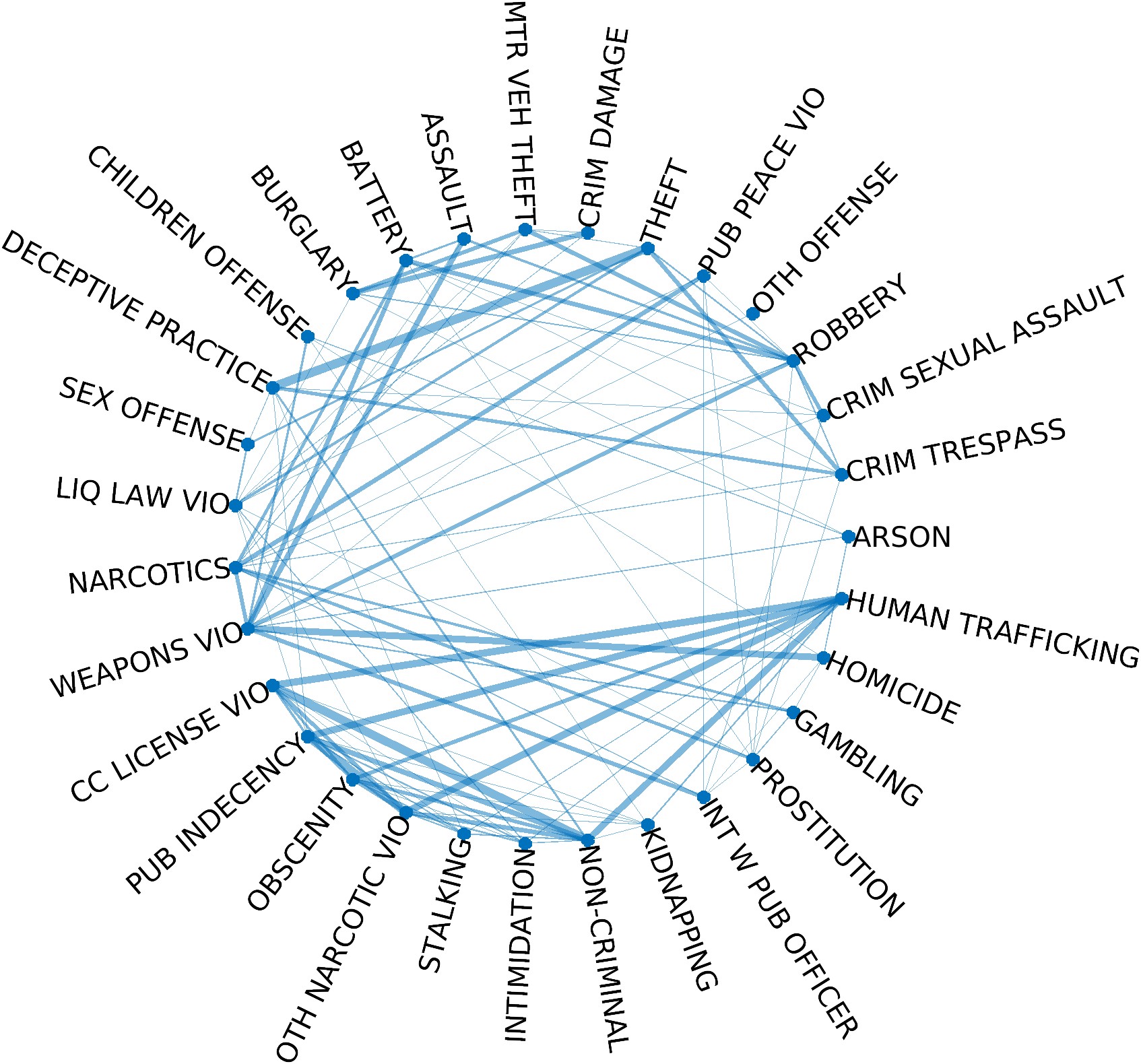}}
  \centerline{(a) CGL}\medskip
\end{minipage}
\hfill
\begin{minipage}[b]{.48\linewidth}
  \centering
  \centerline{\includegraphics[width=4cm]{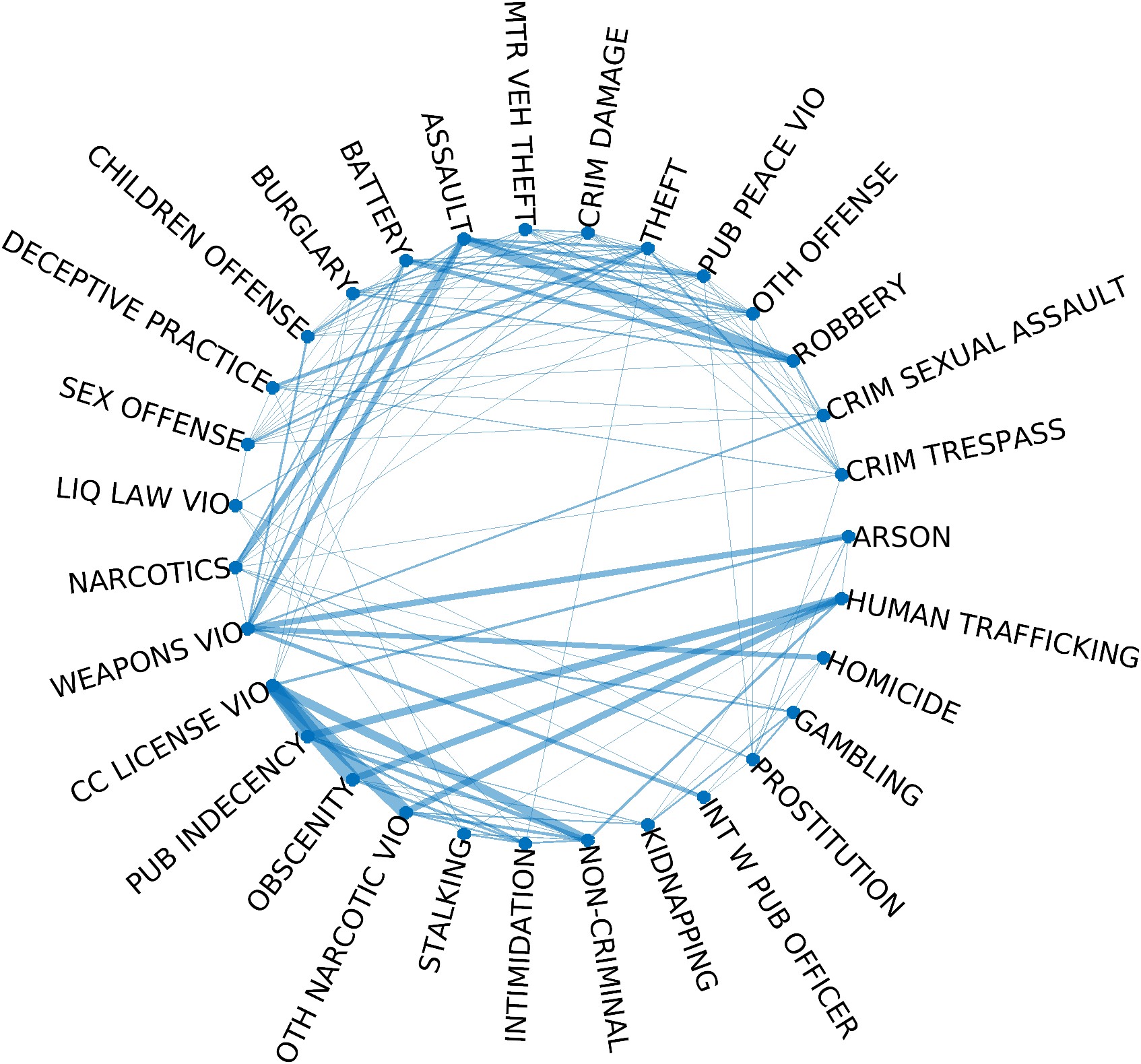}}
  \centerline{(b) GLEN}\medskip
\end{minipage}
\caption{Graphs (normalized) inferred from the Chicago crime dataset using original CGL and GLEN. Nodes correspond to crime types. Width of edges correspond to the edge weights.}
\label{fig:learned_crime_graphs}
\end{figure}

We tune the hyper-parameter to encourage sparsity and initialize GLEN with the graph learned by CGL.
The crime graphs learned by both methods are shown in  Fig.~\ref{fig:learned_crime_graphs}.
As we can see, the CGL method learns dense weak connections with very few dominant edges.
Furthermore, it does not converge when we increase the regularization term to estimate a sparser graph.
On the other hand, GLEN improves the CGL estimations and learns a more interpretable sparse graph, better linking similar crimes together.
For example, ``Assultant'' connects strongly to ``Weapons'', ``Nacortics'', and ``Robbury''.

\subsection{Animals Dataset} 
\begin{figure}[htb]
\begin{minipage}[b]{0.48\linewidth}
  \centering
  \centerline{\includegraphics[width=4cm]{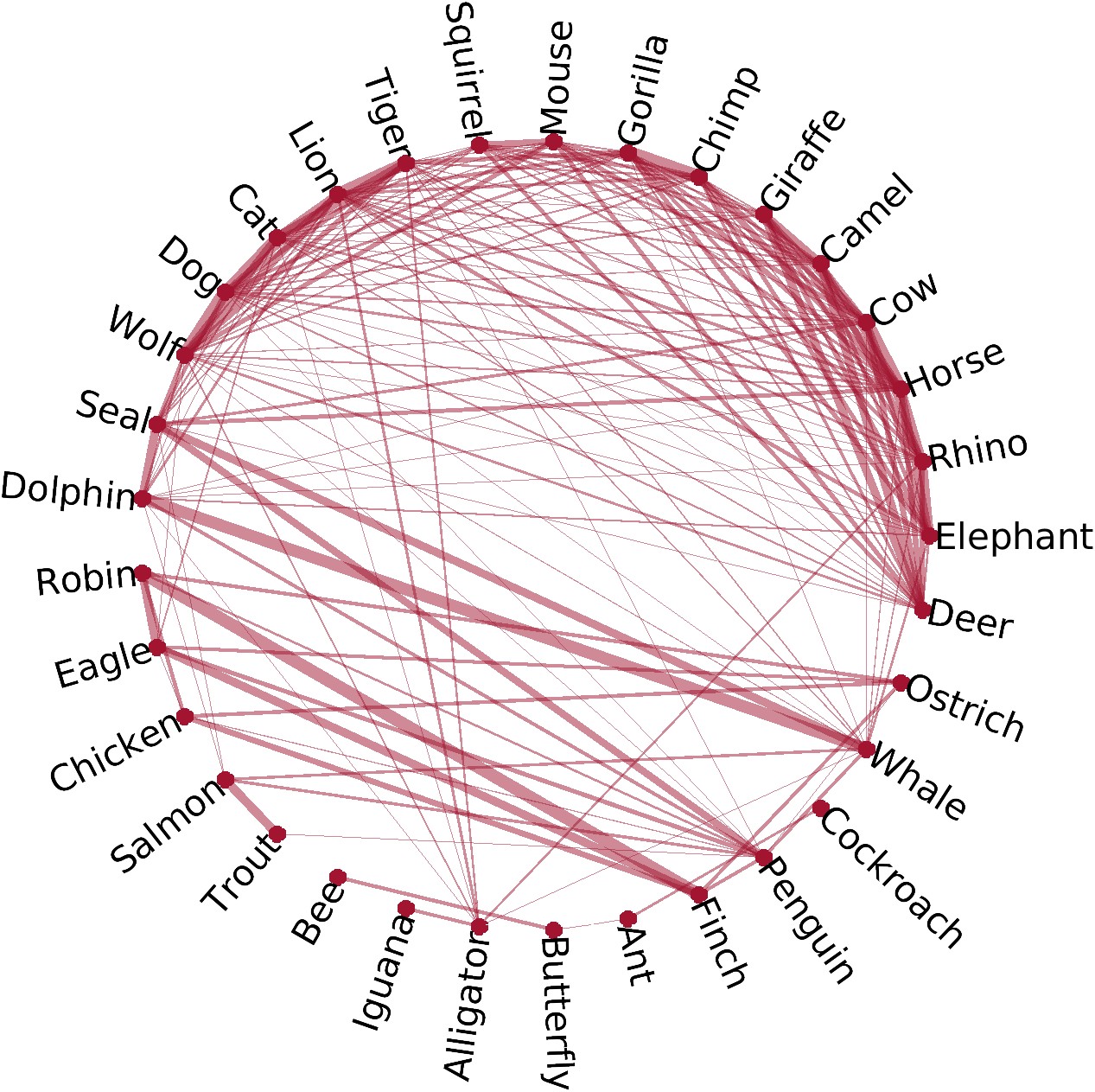}}
  \centerline{(a) CGL}\medskip
\end{minipage}
\hfill
\begin{minipage}[b]{.48\linewidth}
  \centering
  \centerline{\includegraphics[width=4cm]{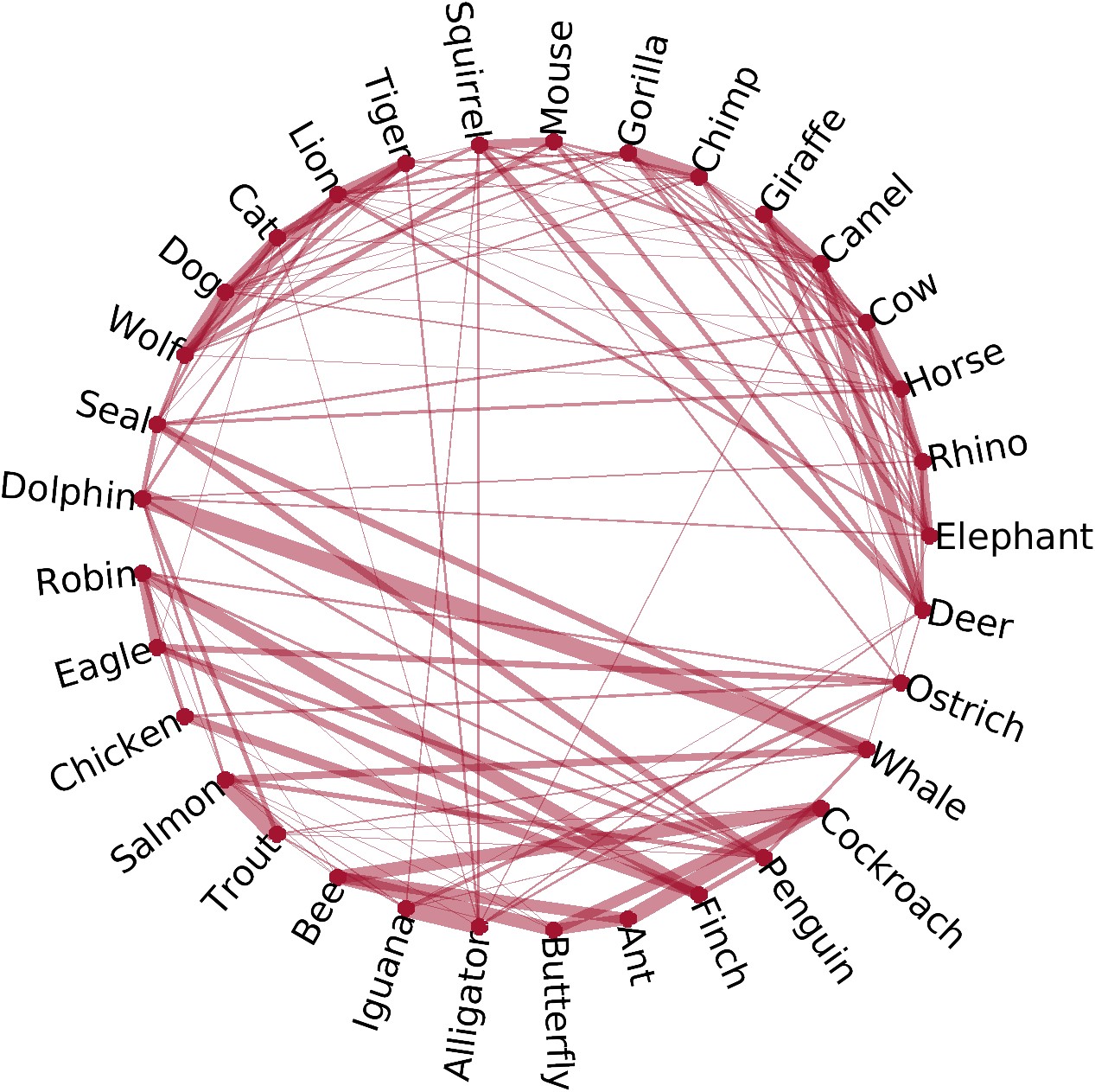}}
  \centerline{(b) GLEN}\medskip
\end{minipage}
\caption{Graphs (normalized) inferred from animal dataset using original CGL and GLEN. The nodes correspond to the species. The widths of the edges correspond to the weights of the edges.}
\label{fig:learned_animal_graphs}
\end{figure}

We also evaluate our method on the Animals dataset \cite{kemp2008discovery}.
The Animal dataset is a binary matrix of size $33 \times 102$,
where each row corresponds to an animal species and each column corresponds to a boolean feature such as ``has wings?", ``has lungs?", ``is dangerous?".
Our goal is to learn a graph where each node represents a specie and each edge represents the similarity between them. 
To accommodate smooth models to the binary signals, previous work \cite{egilmez2017graph,kumar2020unified} used a heuristic statistic $\mathbf{S}=\frac{1}{d}\mathbf{XX}^T+\frac{1}{3}\mathbf{I}$ suggested by \cite{banerjee2008model}.
We instead explicitly model the binary signals using the Bernoulli distribution, resulting in an improper Bernoulli-Logit-Normal model \cite{atchison1980logistic}, and use GLEN to learn a graph Laplacian.
We compare our results with CGL and plot the learned graphs in Fig.~\ref{fig:learned_animal_graphs}.
For CGL, we steadily increase the regularization to encourage sparsity so long as the algorithm converges and no isolated node exists.
Note that GLEN learns sparser graphs and more clear community structures.
Our method also learns meaningful structures that are ignored by CGL, such as the insect sub-network ``Bee-Butterfly-Ant-Cockroach".

\subsection{Neural Dataset}
We now turn to a dataset that has graph-temporal structure.
We evaluate our methods on a dataset of neural recordings, $Area2\_Bump$, from the neural latents benchmark (NLB) \cite{pei2021neural}.
This dataset consists of multiple trials of neural spiking activity and simultaneous behavior data of a macaque performing a bump task.
During the bump task, the macaque controls a cursor to perform center-out reaches towards one of eight target directions. In a subset of random trials, the macaque is interrupted by a bump shortly before the reach.
Neural activity is recorded from  Brodmann’s area 2 of the somatosensory cortex, which has been shown to contain information about whole-arm kinematics.
The neural recording are contained in a non-negative integer matrix $\mathbf{X} \in \mathbb{Z}_{+}^{N \times T \times K}$, where each entry $\mathbf{X}_{itk}$ counts the firing of neuron $i$ in time bin $t$ during trial $k$.
Following standard procedure in \cite{pei2021neural}, we resample the 1-ms resolution signals into 5-ms bins.

Learning the interactions between the neurons from these spiking activity matrices enables relating functional connectivity patterns to behavior.
A graph signal $\mathbf{X}_{:tk}$ is the simultaneous spiking of all neurons at time $t$ in trial $k$.
Since graph signals are not independent but temporally correlated, we use GLEN-TV to infer a graph of neurons for each trial and analyze the inferred graphs across different conditions (target directions).
We model the observation with Poisson distribution which is used in the standardized co-smoothing evaluation \cite{macke2011empirical}.
We plug-in Poisson distribution to Eq.~\eqref{eq:time_vertex_smoothness} to obtain the objective function
\begin{equation}
\label{eq:poisson_tvgraph_learning}
    \begin{gathered}
    \min_{\mathbf{Y},\boldsymbol{\mu},\mathbf{L} \in \mathcal{L}} \Bigl\{ - \Tr{((\mathbf{Y}^T+\mathbf{1}{\boldsymbol{\mu}}^T) \mathbf{X})} + \mathbf{1}^T \exp{(\mathbf{Y}+\boldsymbol{\mu}{\mathbf{1}}^T)} \mathbf{1}\\ + \gamma \Tr{(\mathbf{YTY}^T)}
    + \beta (\Tr{(\mathbf{Y}^T\mathbf{LY})} + \alpha h(\mathbf{L})) \Bigr\},\\
    s.t.\ \mathbf{Y}^T \mathbf{1} = \mathbf{0},
    \end{gathered}
\end{equation}
and the update rules
\begin{align}
    & \begin{gathered}
    \nabla_j = \left. - \mathbf{x}_j + \exp{(\boldsymbol{\mu}+\mathbf{y}_j)} + \beta \mathbf{L y}_j \right. \\
    \left. \ + 2 \gamma (2 \mathbf{y}_j - \mathbf{y}_{j-1} - \mathbf{y}_{j+1}) \right., 
    \end{gathered} \\
    & \nabla_j^2 = \textrm{diag}(\exp{(\boldsymbol{\mu}+\mathbf{y}_j)}) + \beta \mathbf{L} + 4 \gamma \mathbf{I}_N.
\end{align}

\begin{figure*}[htb]
\begin{minipage}[b]{0.48\linewidth}
\small
  \centering
    \begin{tabular}{rcc}
    & Average Spiking & Average Denoised Signals\\
    \rotatebox{90}{$0^{\circ}$}  &
    \includegraphics[align=c,width=3.6cm]{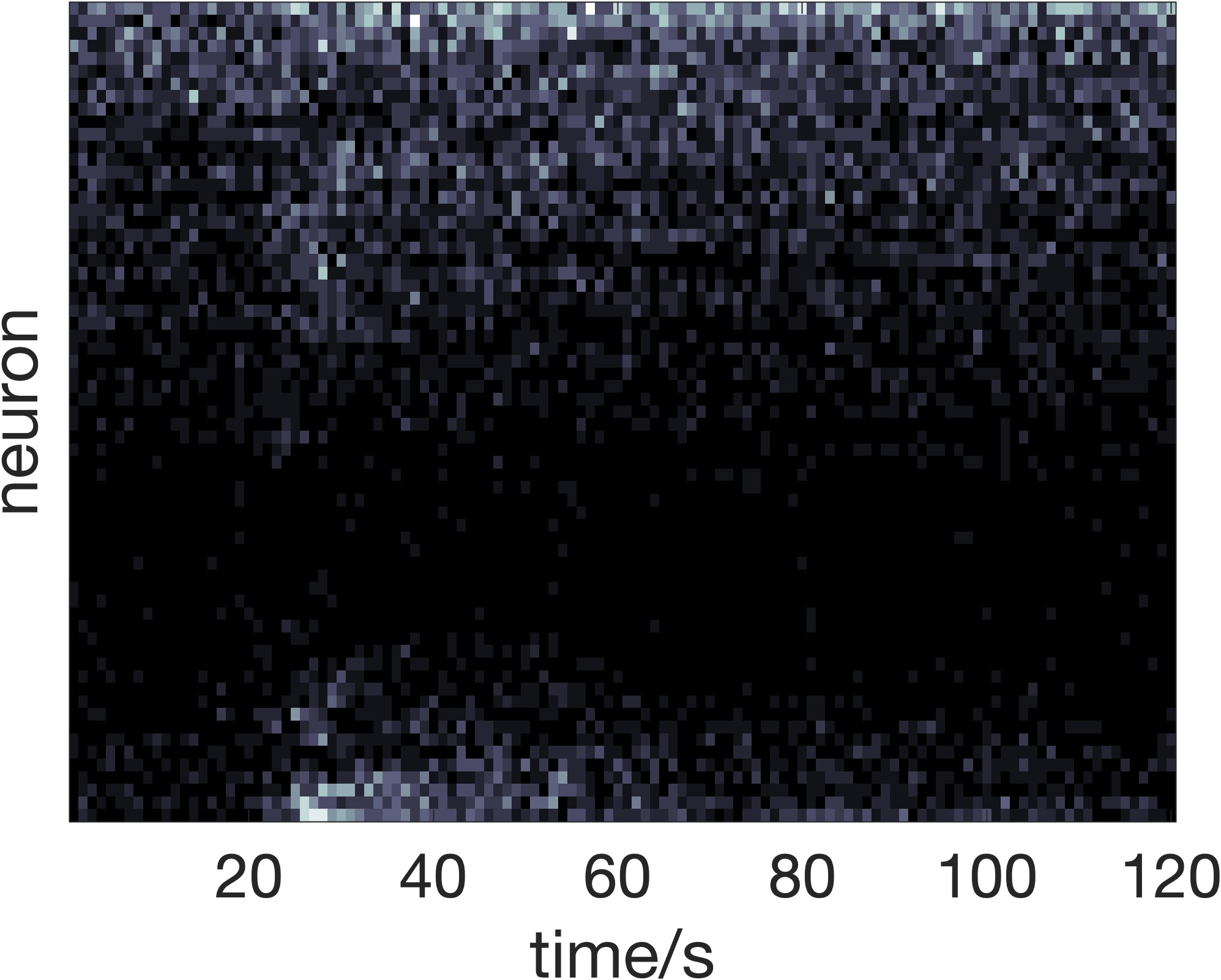}     &  \includegraphics[align=c,width=3.6cm]{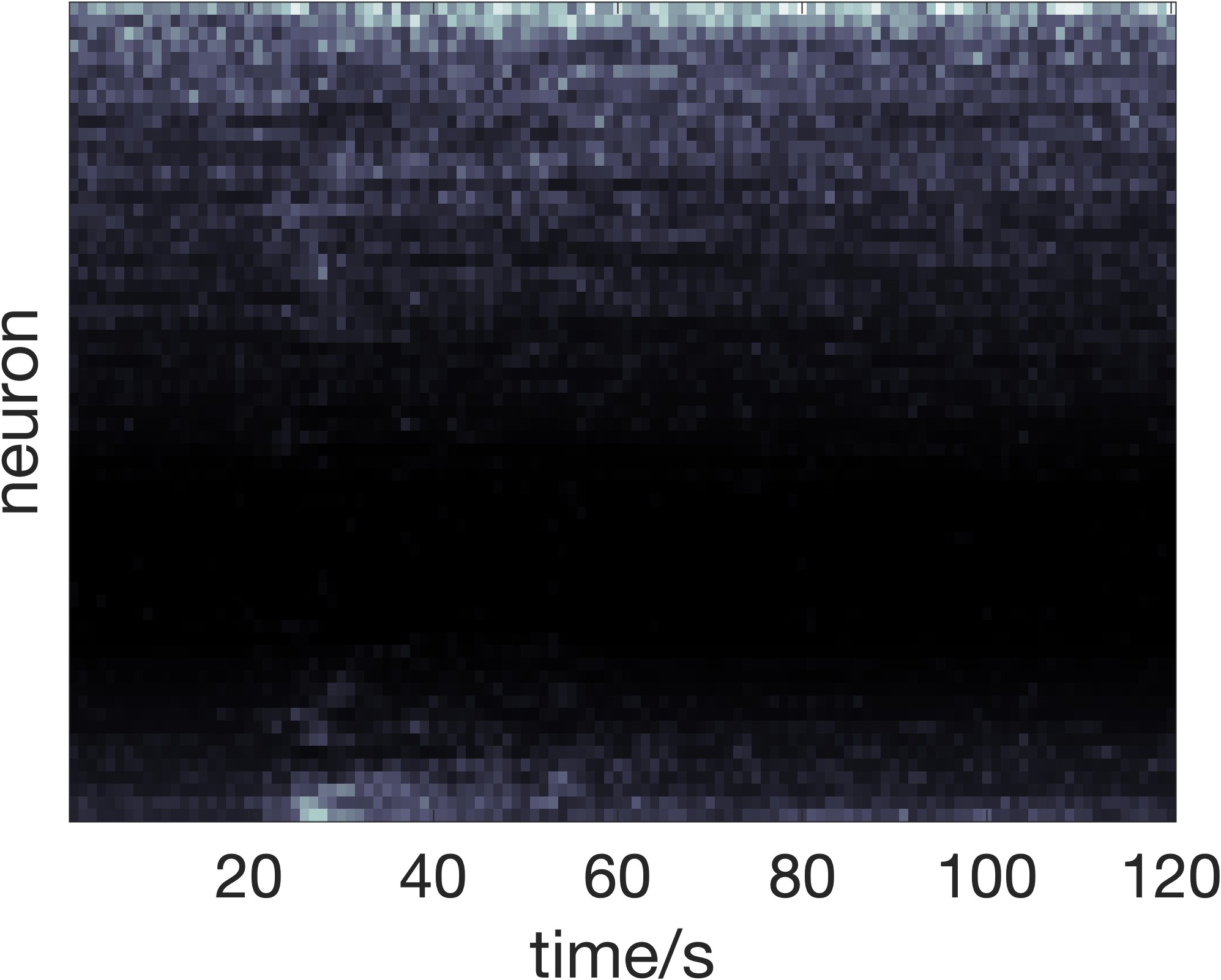}\\
    \rotatebox{90}{$45^{\circ}$}  &
    \includegraphics[align=c,width=3.6cm]{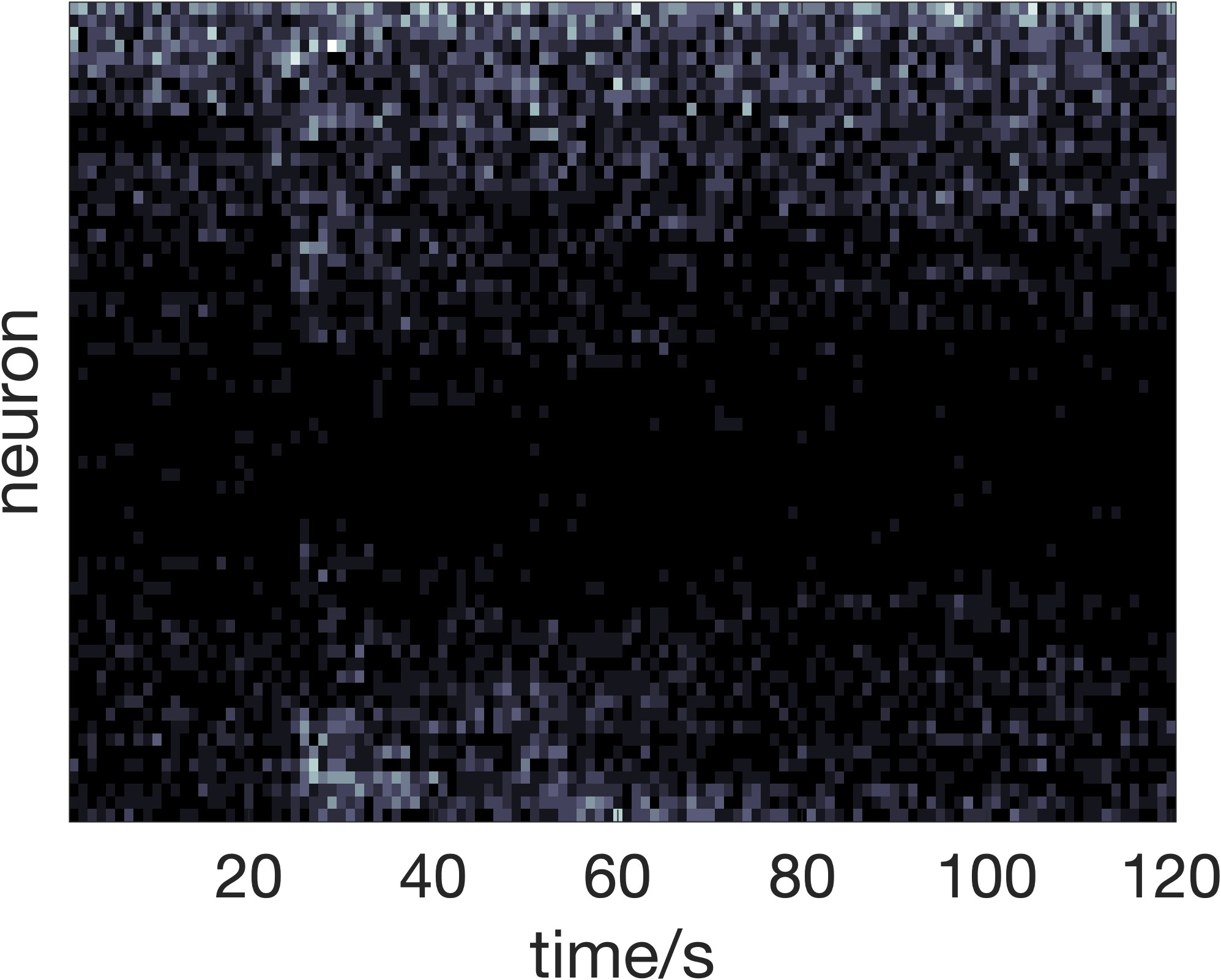}  &\includegraphics[align=c,width=3.6cm]{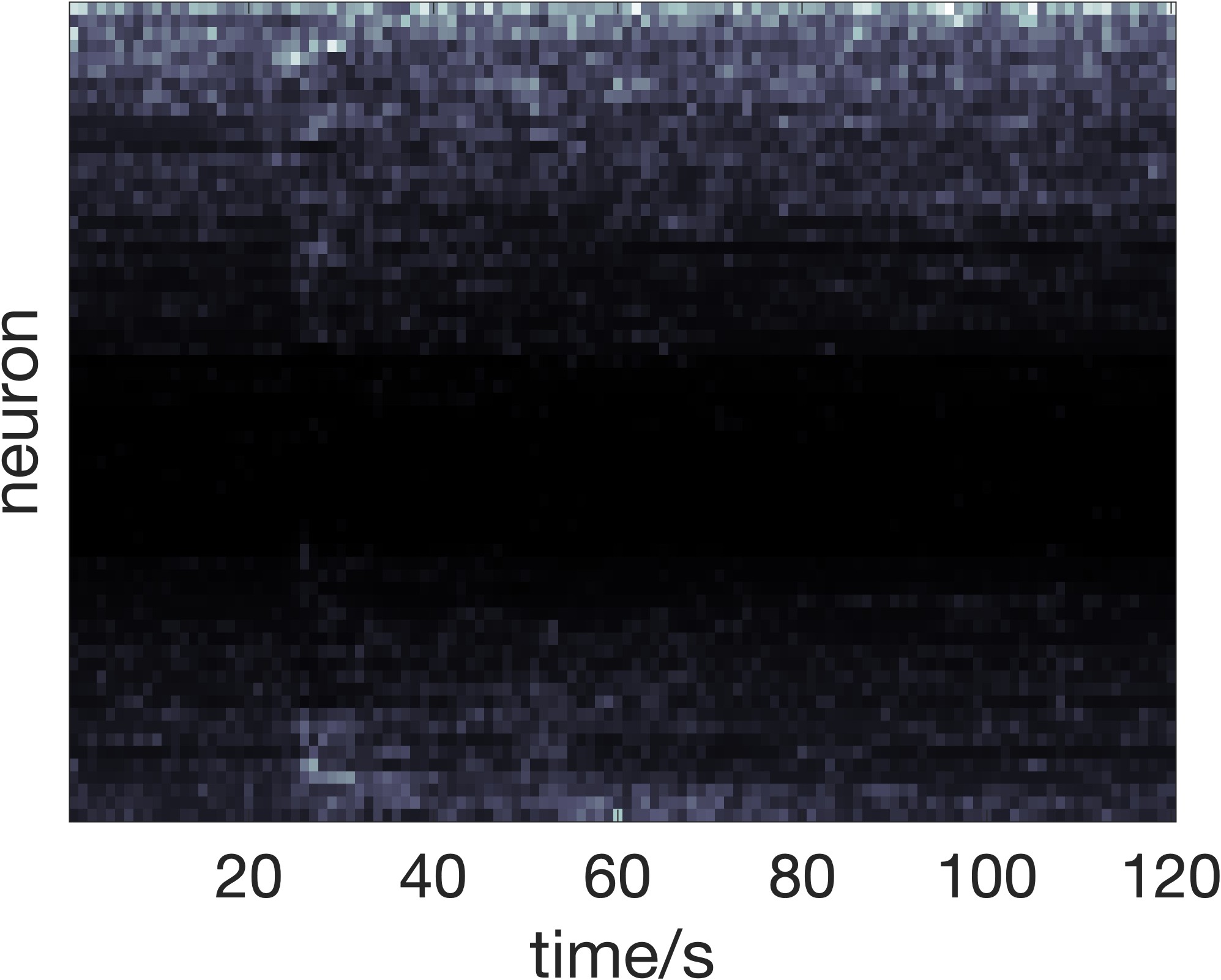}\\
    \rotatebox{90}{$90^{\circ}$}  &
    \includegraphics[align=c,width=3.6cm]{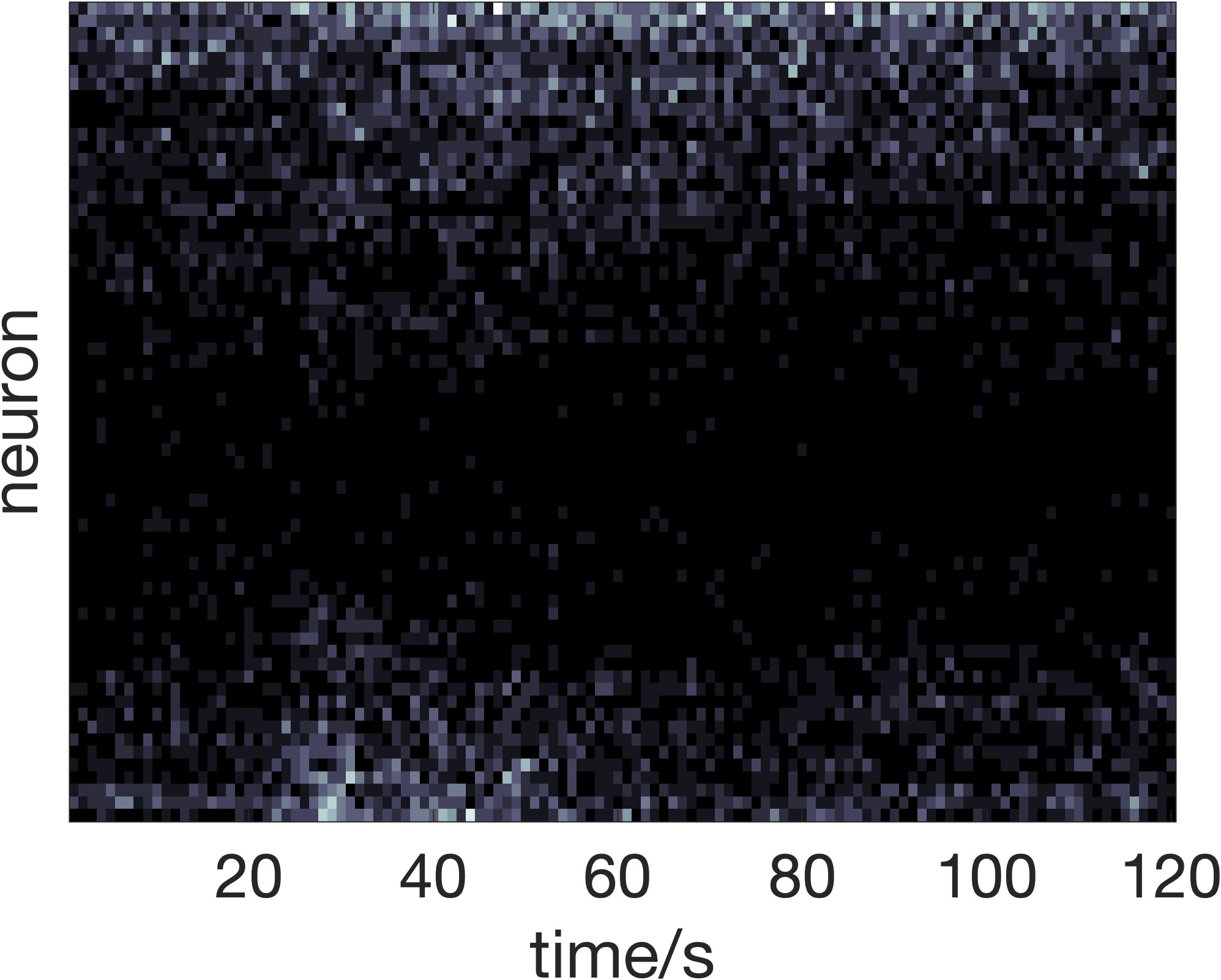} &   \includegraphics[align=c,width=3.6cm]{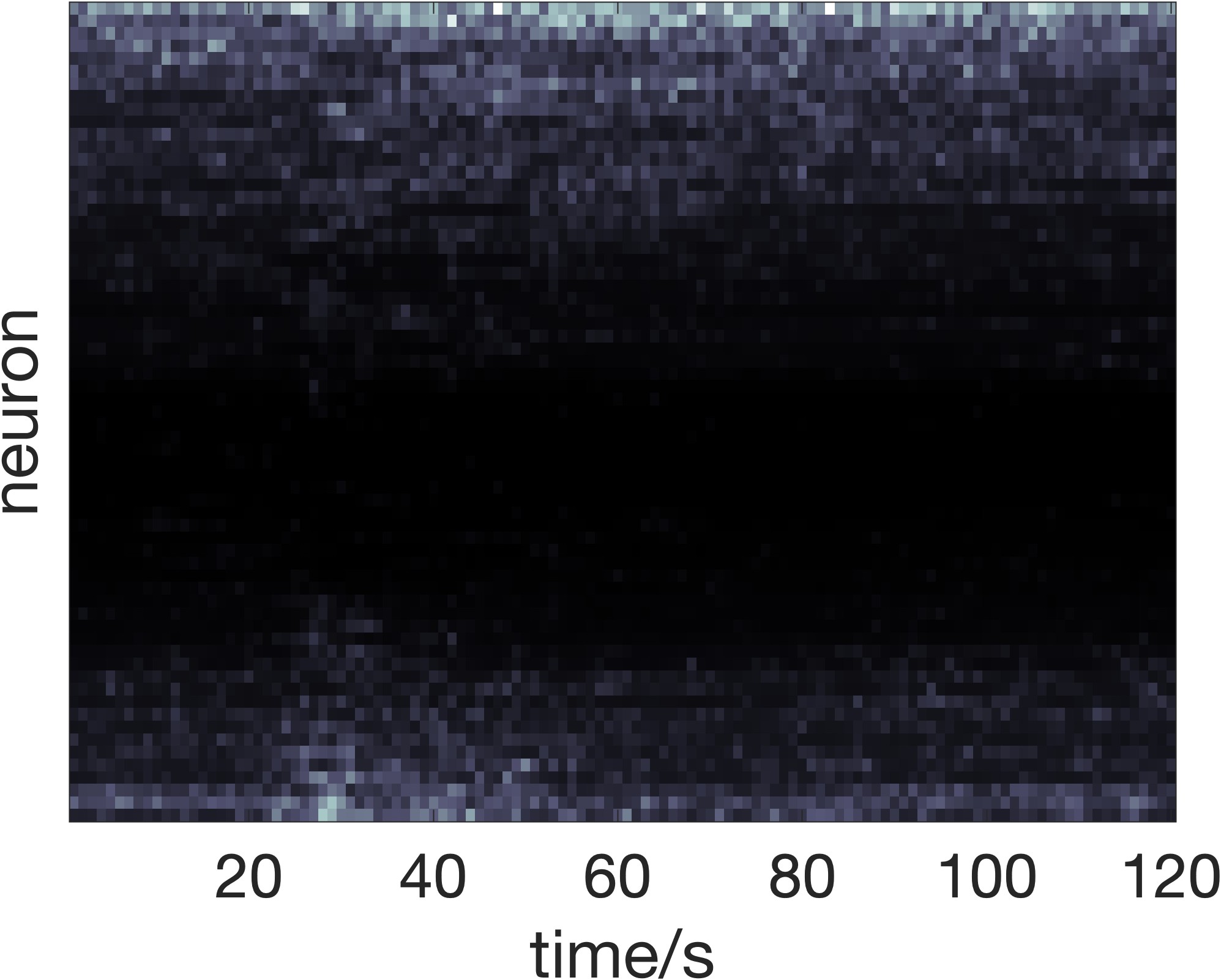}\\
    \rotatebox{90}{$135^{\circ}$}  &
    \includegraphics[align=c,width=3.6cm]{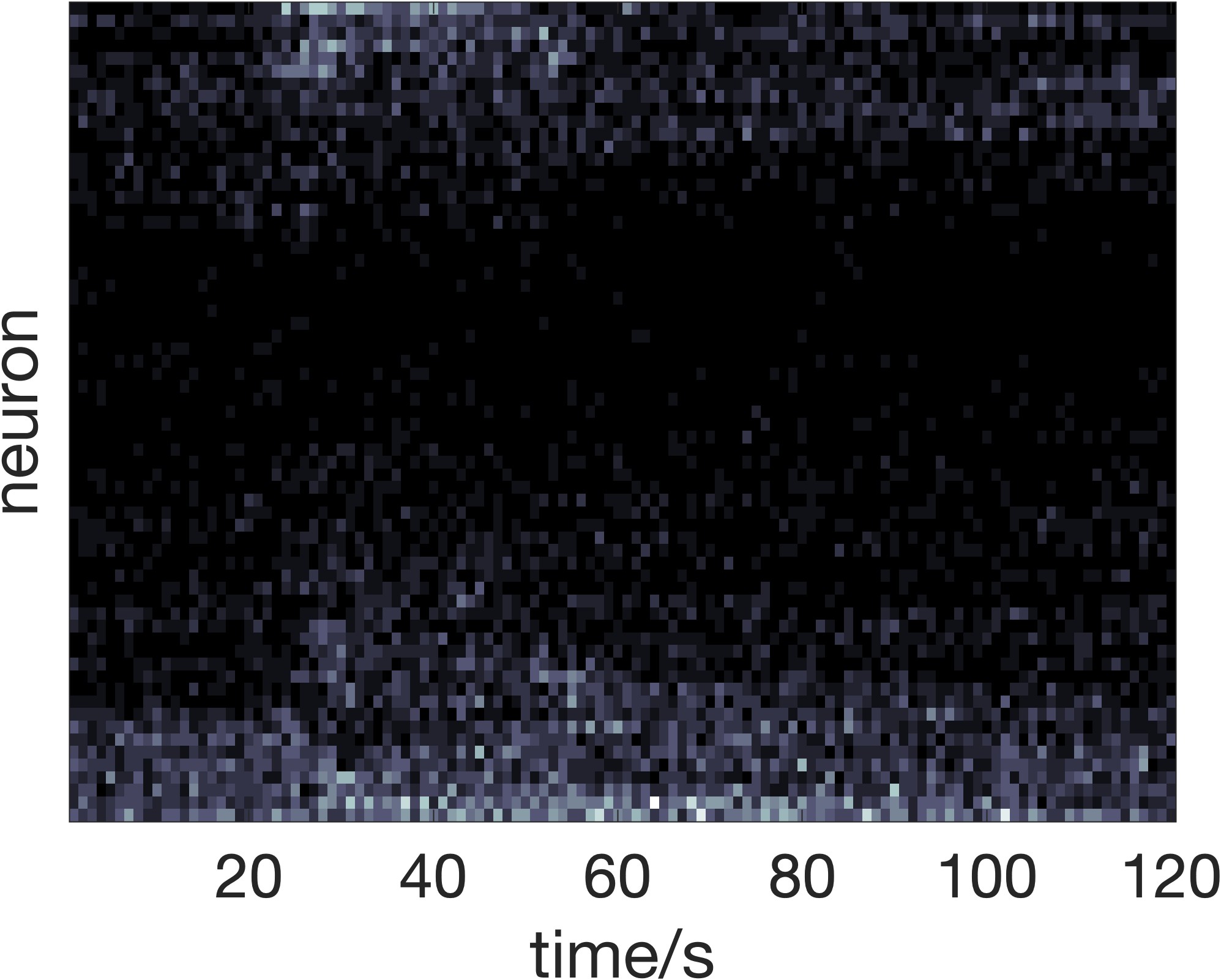}    &   \includegraphics[align=c,width=3.6cm]{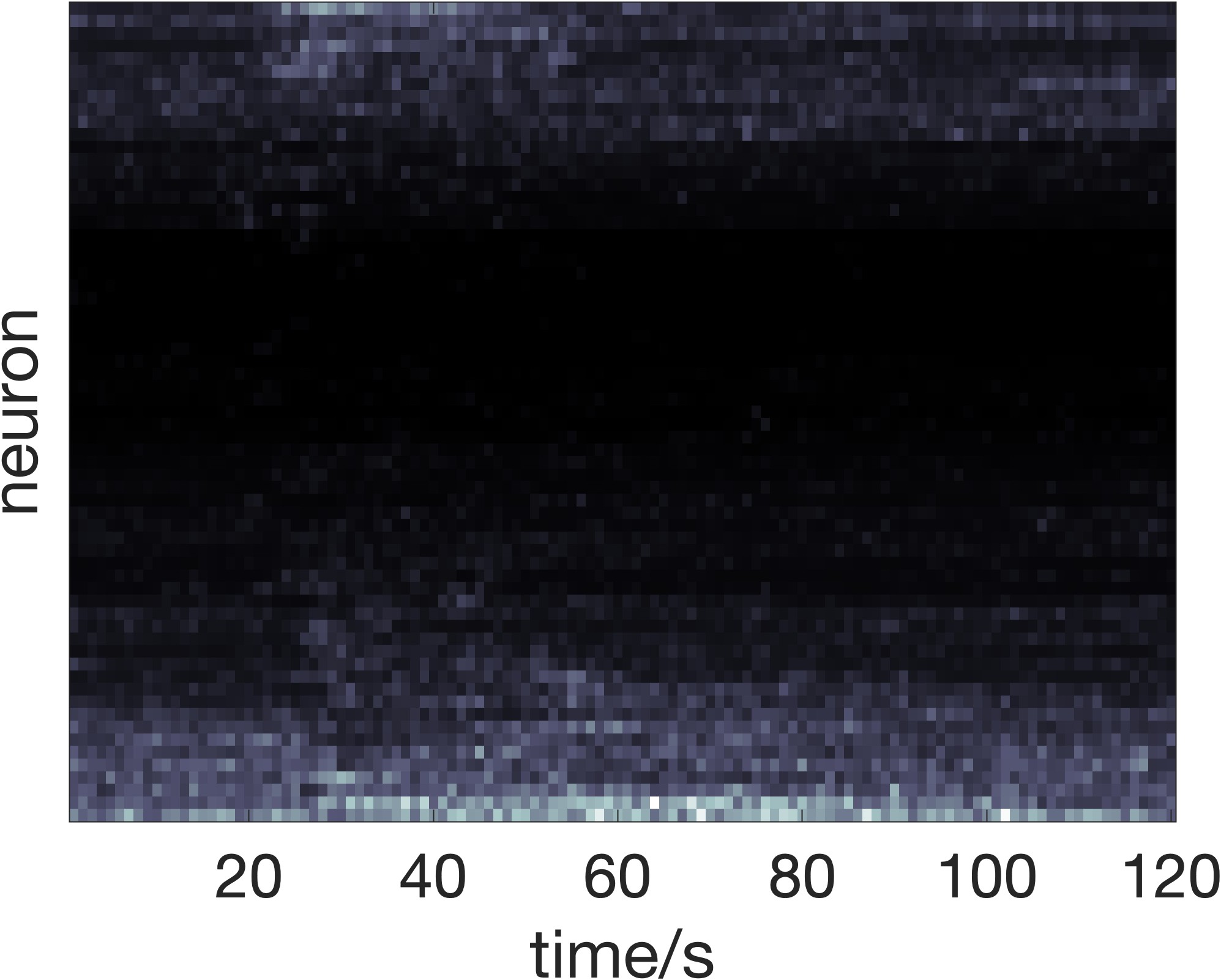}\\
    \end{tabular}\medskip
\end{minipage}
\hfill
\begin{minipage}[b]{.48\linewidth}
\small
  \centering
    \begin{tabular}{rcc}
    & Average Spiking & Average Denoised Signals\\
    \rotatebox{90}{$180^{\circ}$}  &
    \includegraphics[align=c,width=3.6cm]{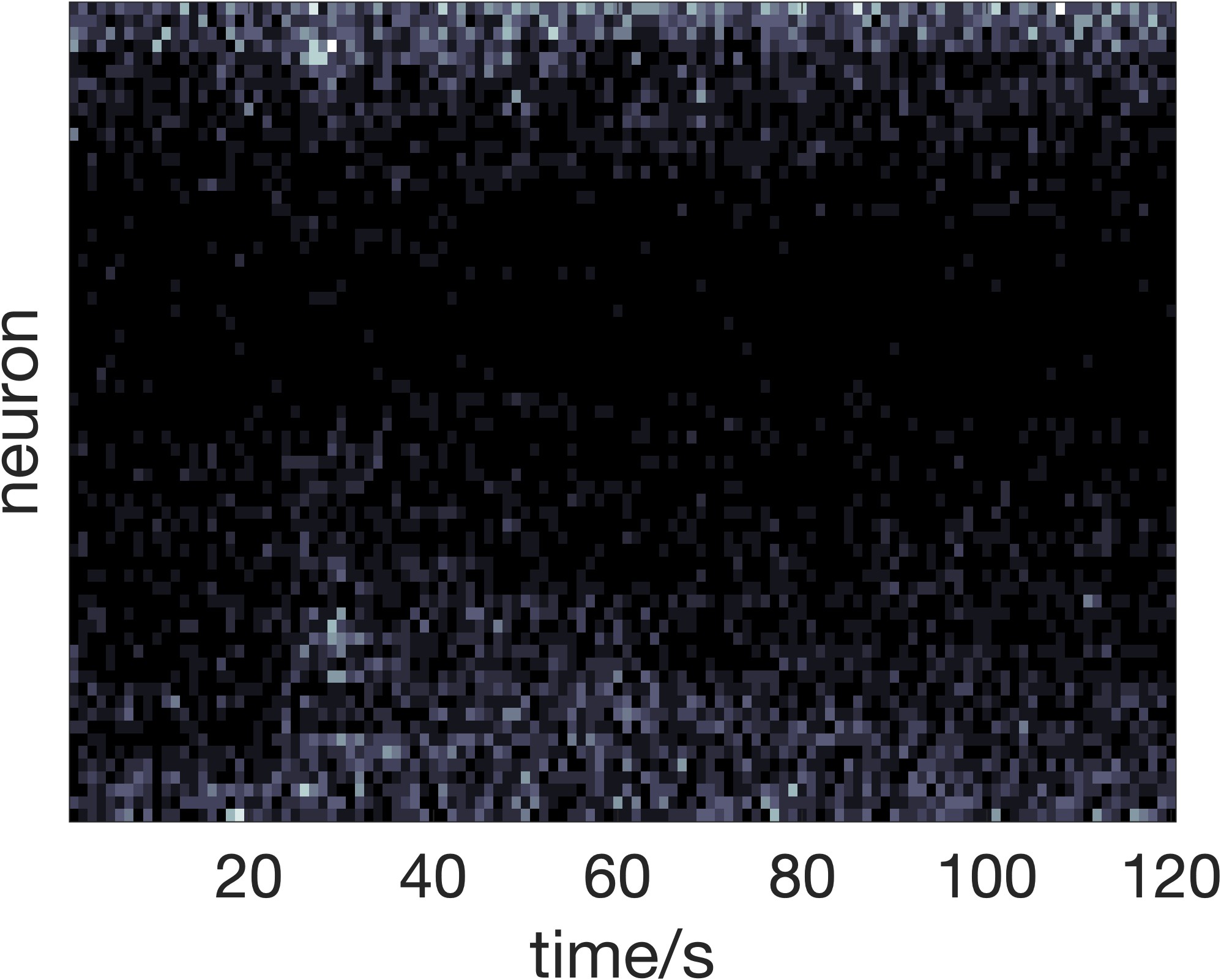}   &    \includegraphics[align=c,width=3.6cm]{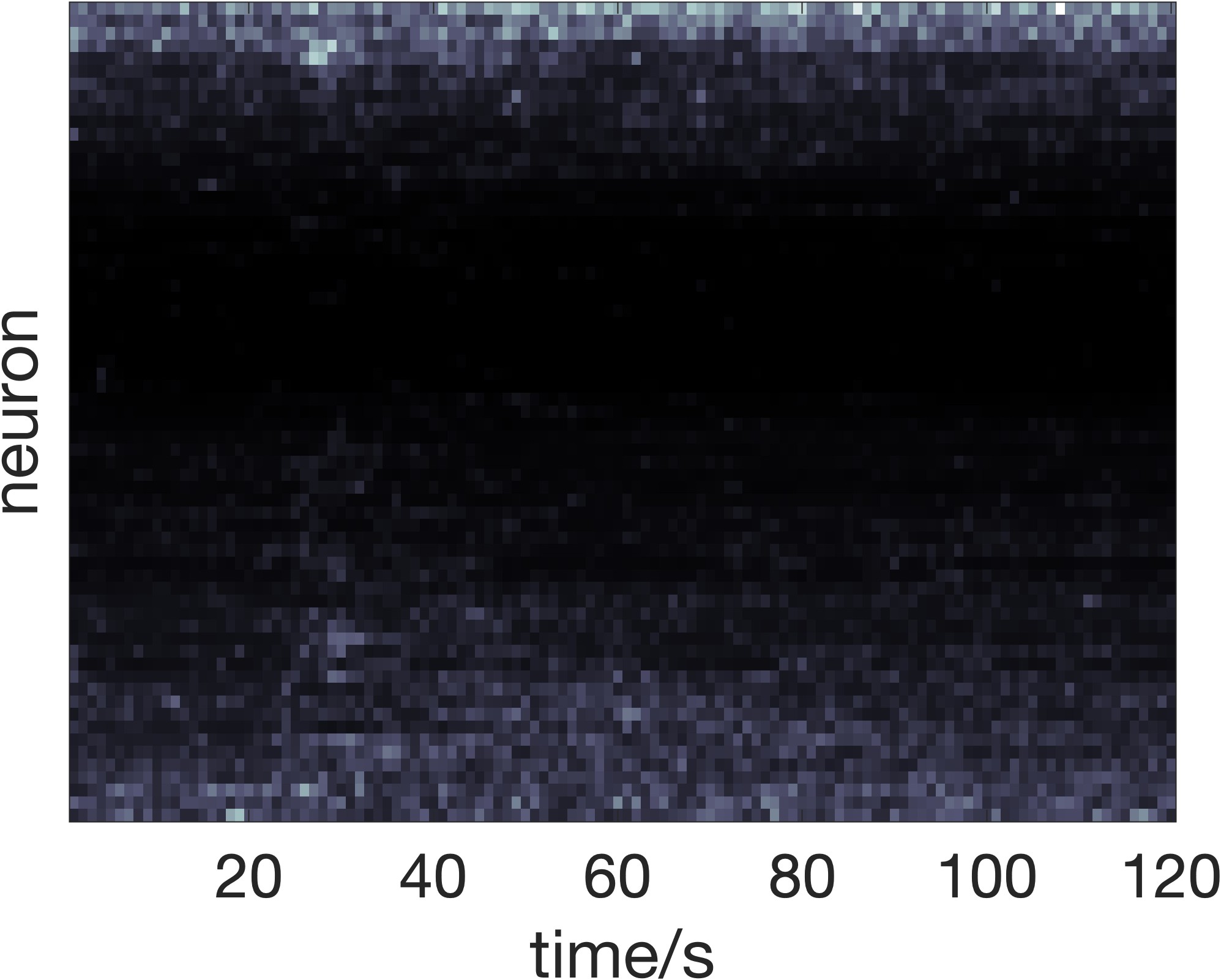}\\
    \rotatebox{90}{$225^{\circ}$}  &
    \includegraphics[align=c,width=3.6cm]{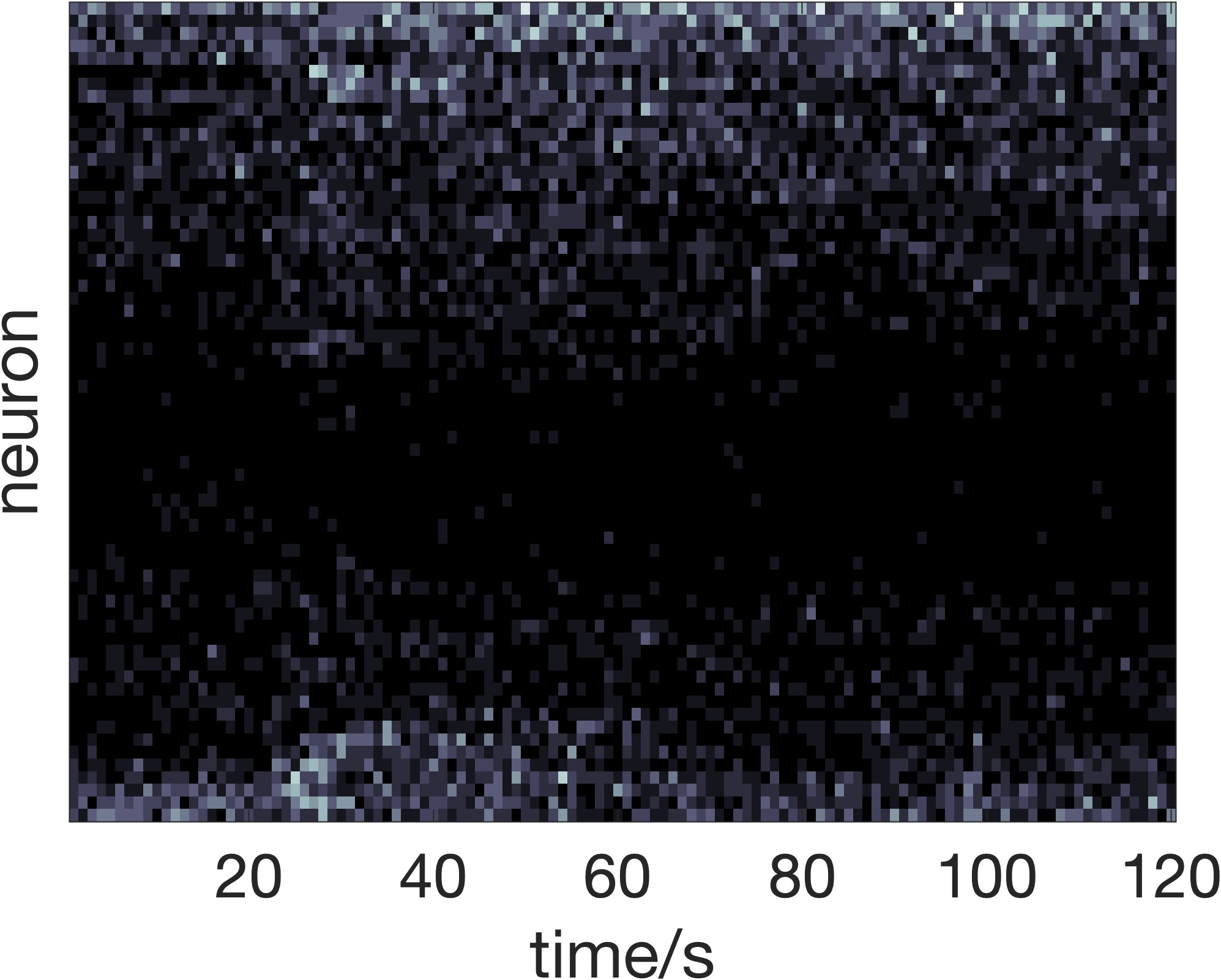} &  \includegraphics[align=c,width=3.6cm]{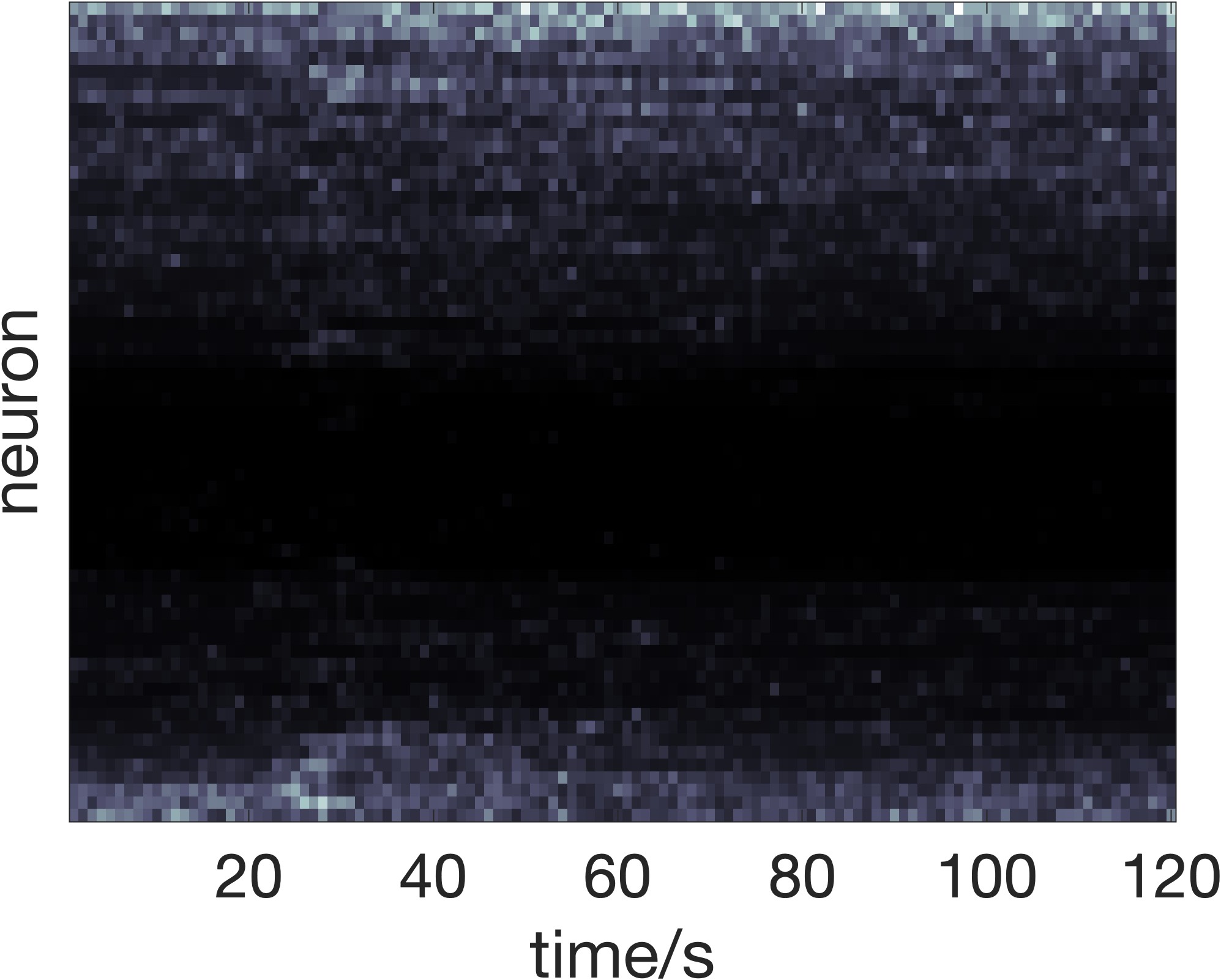}\\
    \rotatebox{90}{$270^{\circ}$}  &
    \includegraphics[align=c,width=3.6cm]{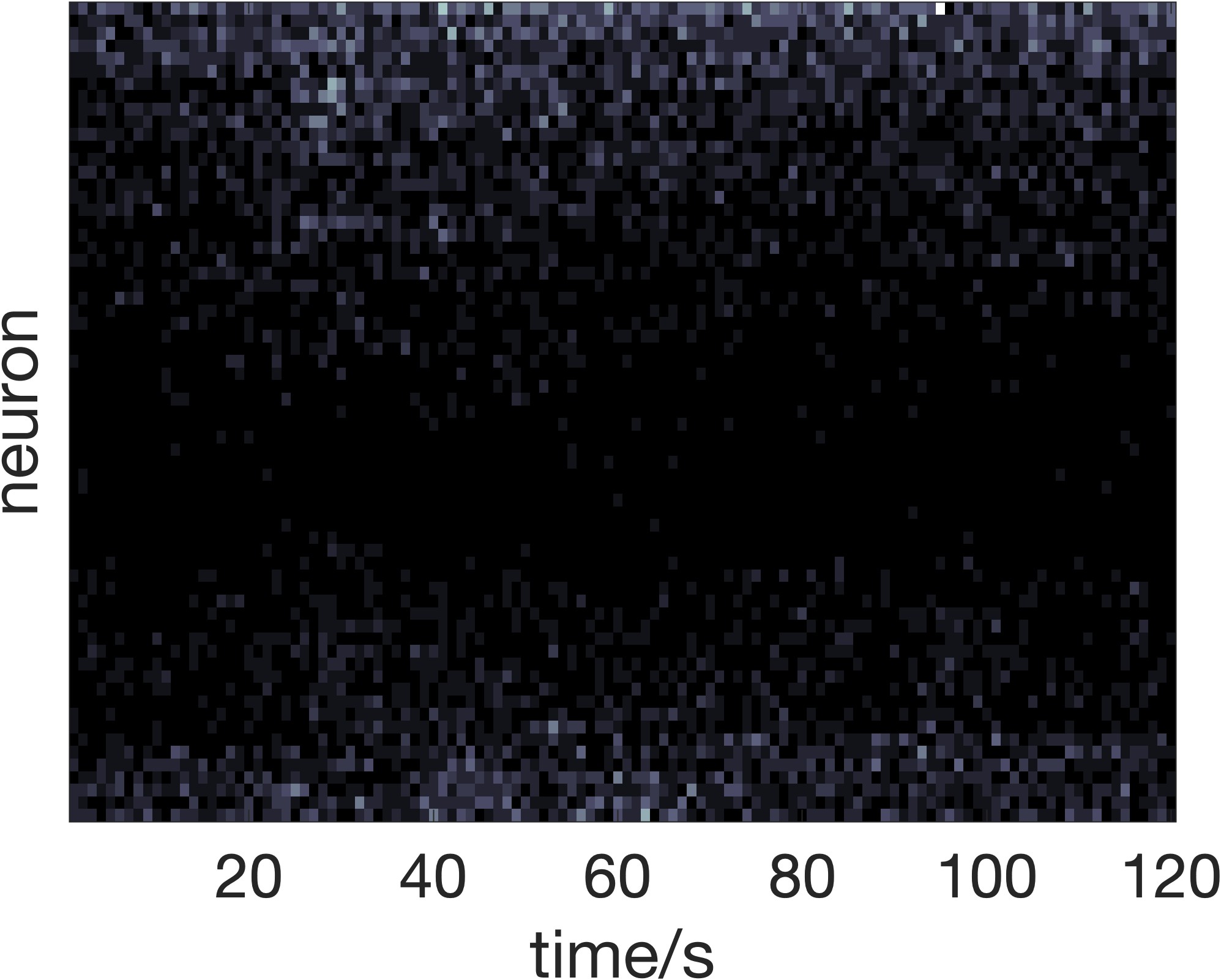}    &   \includegraphics[align=c,width=3.6cm]{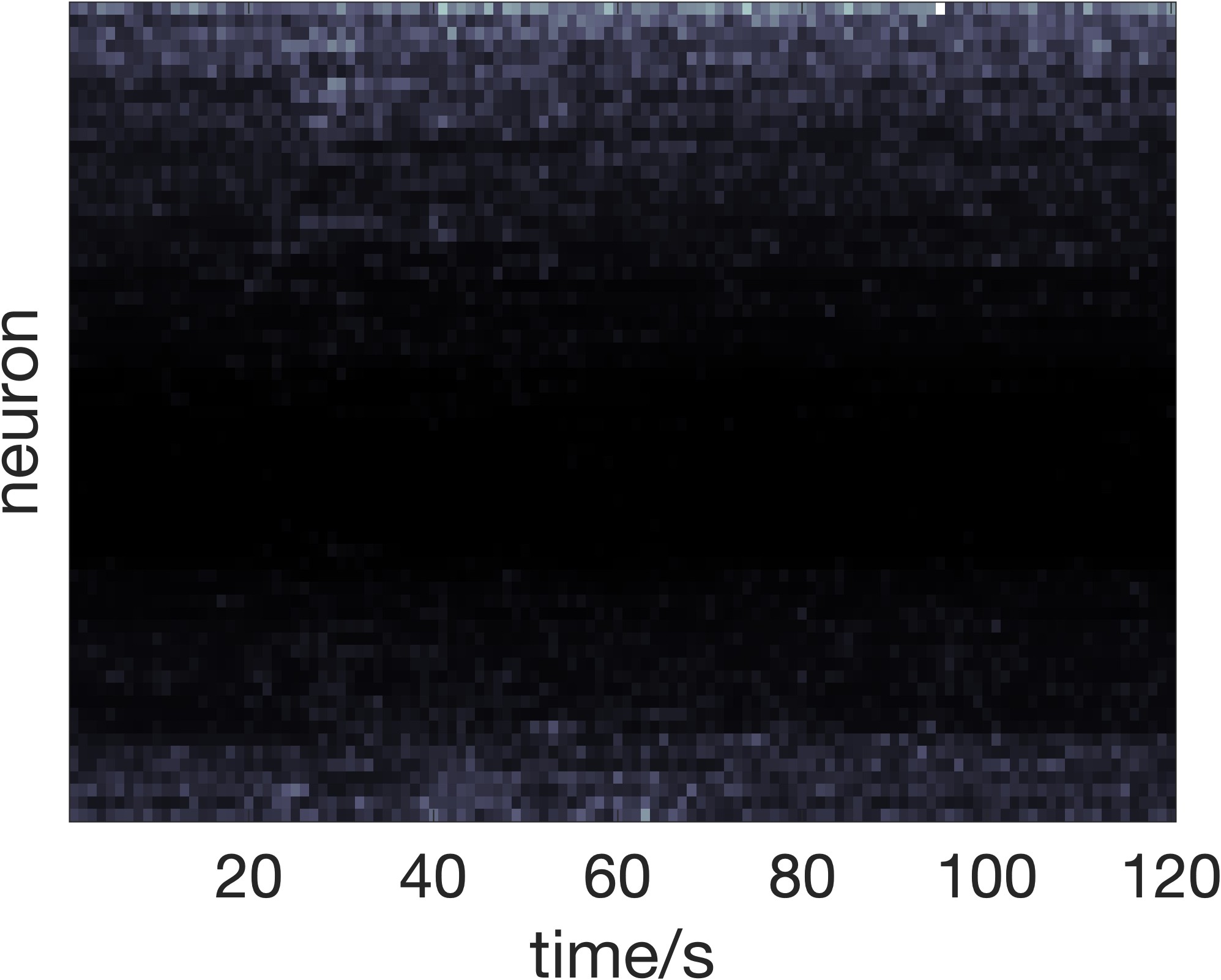}\\
    \rotatebox{90}{$315^{\circ}$}  &
    \includegraphics[align=c,width=3.6cm]{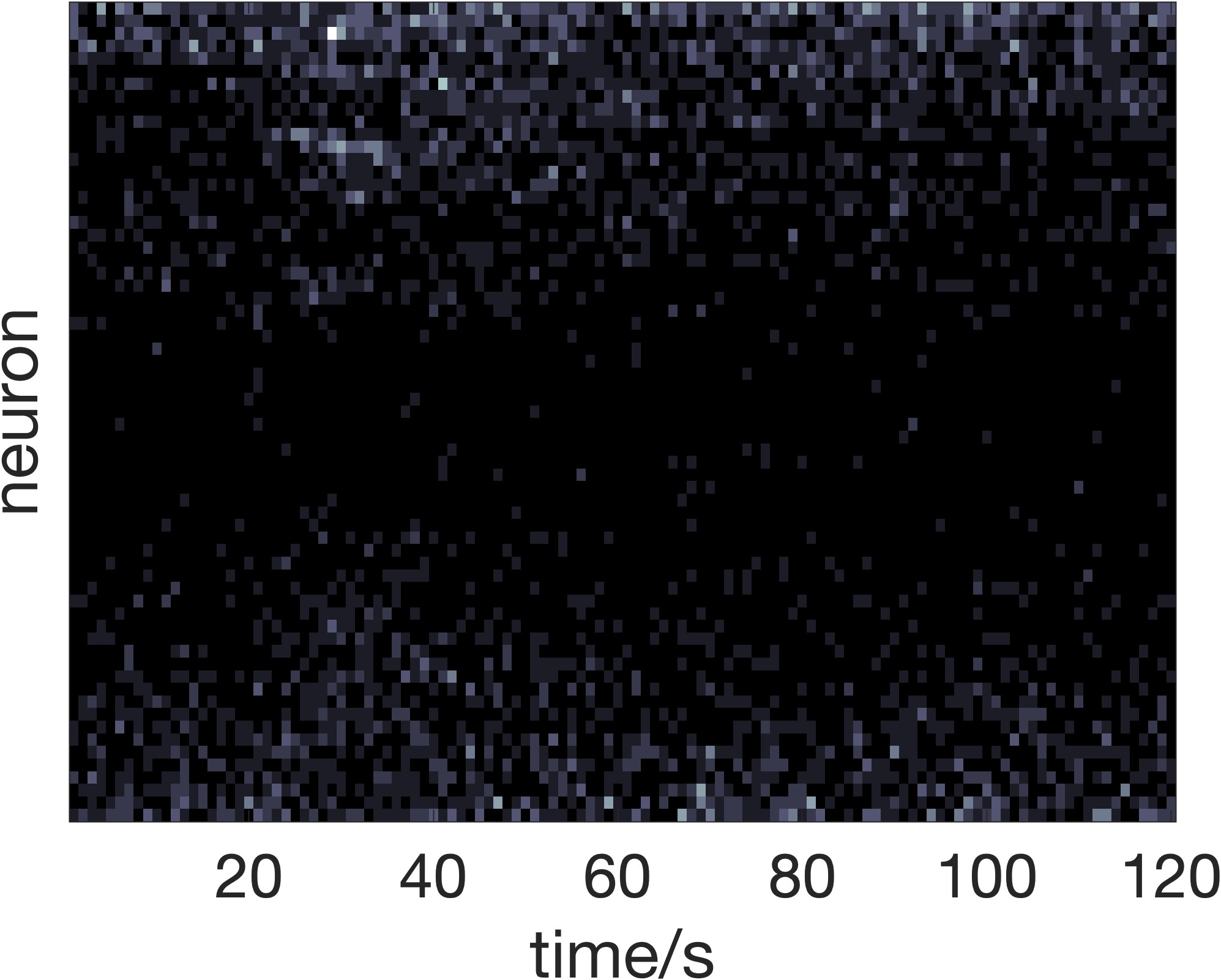}    &   \includegraphics[align=c,width=3.6cm]{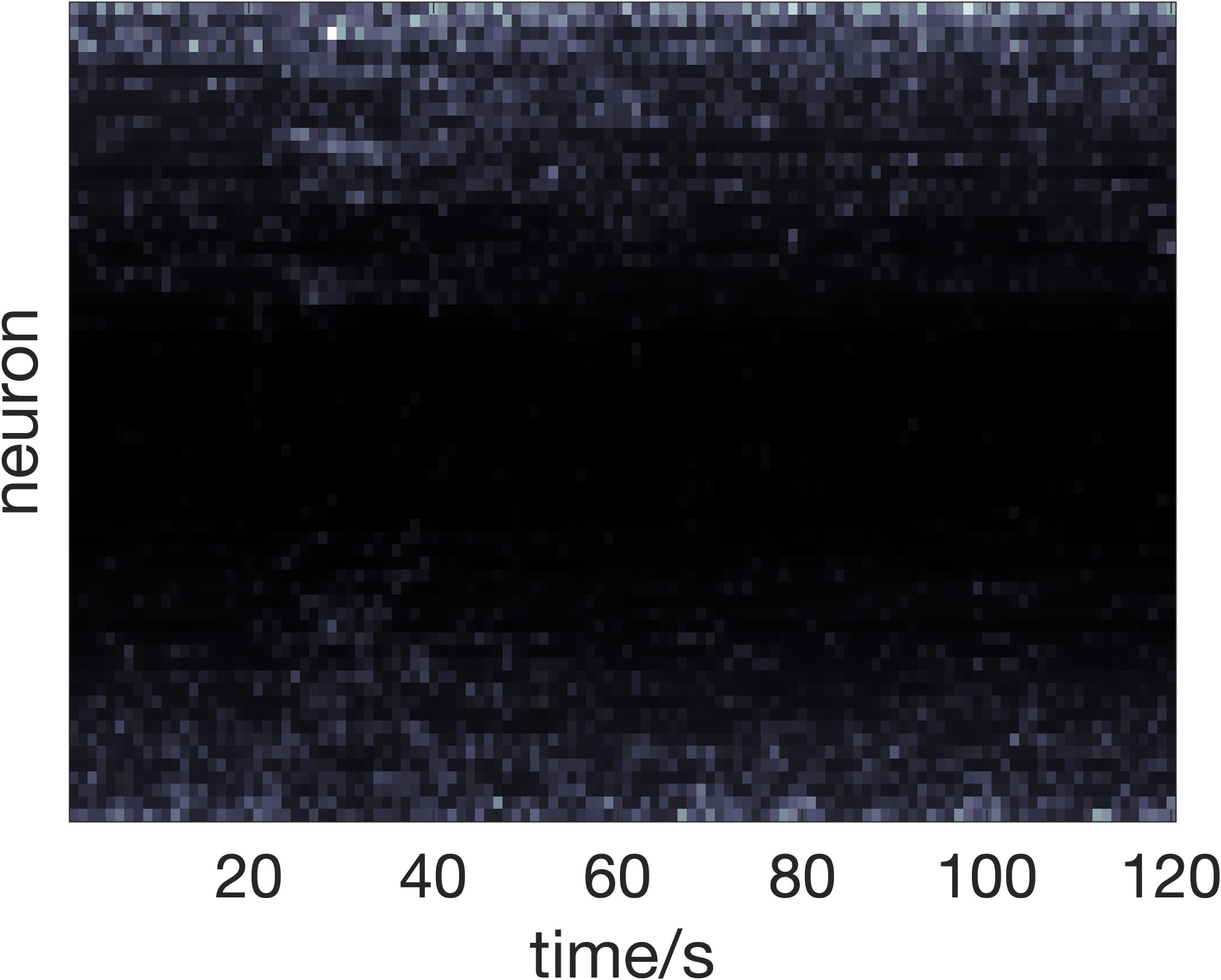}\\
    \end{tabular}\medskip
\end{minipage}
\centering
\caption{Average of raw neural spiking data compared against average denoised signals by GLEN-TV, for each target direction shown on the left.
Brighter indicates higher firing rate.
}
\label{fig:conditional_denoised_nlb_signals}
\end{figure*}

We perform linear discriminant analysis (LDA) on the degree vector of learned Laplacians, using direction conditions as 8 class labels.
LDA achieves $67.86\%$ accuracy which indicates that the structural information in our learned graphs encode the class conditions.
When applying GLEN without the temporal modeling, LDA only achieves $61.26\%$ accuracy, which indicates the importance of modeling temporal correlations.
We also visualize the average spiking activity and the averaged denoised signals for all 8 conditions in Fig.~\ref{fig:conditional_denoised_nlb_signals}.
The denoised signals are the exponential of learn smooth representation $\mathbf{Y}$, which can be considered as the firing rate of neurons.
Note that GLEN-TV smooths the signals both spatially and temporally.

\section{Conclusions}
\label{sec:conclude}

In this paper, we generalize combinatorial graph Laplacian learning from continuous and unbounded noisy signals to various data types.
Inspired by generalized linear models, we established a GSP-based generative model that produces signals of different responses of exponential family distributions whose expected values are underlying smooth signal representations.
To effectively learn the graph Laplacian, we proposed GLEN which alternates between a graph learning step that refines the estimation of the Laplacian, and a denoising step that refines the underlying smooth signal representations, until convergence.
We further extend GLEN using variational methods to compensate for the neglected stochasticity of latent smooth signal representation.
Since real-world graph signals are often not independent but temporal correlated, we also extended GLEN to the time-vertex framework for learning from time-varying graph signals.
We demonstrated that our methods achieve more accurate Laplacian estimation on different synthetic graph models compared to competing approaches, and can learn meaningful graphs on various real datasets. 
While our experiments highlighted Poisson and Bernoulli distributions, our framework is general for exponential family distributions. 

This paper serves as a first step toward a GSP model of multivariate non-Gaussian distributions and opens multiple directions for future works.
One direction is to study the MLE of the graph Laplacian.
Our variational approximation alleviates the computational difficulty of the likelihood, but other approaches, such as Monte Carlo simulation, might yield better approximations.
Another interesting direction is to obtain theoretical results for GLEN.
Since GLEN heavily relies on regularization, quantifying how regularization affects estimation errors can help us better understand its behavior. 
Studying the consistency and variance of the estimator is also desired.
Finally, the exponential family model can be further generalized.
From the perspective of noise distributions, although we dramatically enlarged the scope of data types of the GSP-based graph inference methods, our model can be more versatile by encompassing dispersion or Bayesian inference.
From the graph learning perspective, generalizing our model to multi-dimensional tensors is of great interest \cite{stanley2020multiway}.

\section*{Appendix}
\subsection*{Derivation of Variational Loss}
\label{sec:exp_noise_vi}
We now consider the underlying smooth signal representation as latent variables, and derive the evidence lower bound (ELBO) for the variational approach.
Following standard procedure, we have the following
\begin{align}
\label{eq:standard_elbo}
     & \log \int p(\mathbf{x}|\mathbf{y})p(\mathbf{y}) d\mathbf{y} \\ = & \log \mathbb{E}_{q(\mathbf{y})} [\frac{p(\mathbf{x}|\mathbf{y})p(\mathbf{y})}{q(\mathbf{y})}] \\ 
     \geq & \mathbb{E}_{q(\mathbf{y})} [\log \frac{p(\mathbf{x}|\mathbf{y})p(\mathbf{y})}{q(\mathbf{y})}].
\end{align}
We plug-in $p(\mathbf{x}|\mathbf{y})$ given in Eq.~\eqref{eq:noisy_exp_signal_generation} and use a Gaussian distribution with mean $\bar{\mathbf{y}}$ and covariance $\mathbf{\Lambda}$ for $q(\mathbf{y})$
\begin{align}
    & \mathbb{E}_{q(\mathbf{y})} [\log \mathbf{1}^T k(\mathbf{x}) + {(\mathbf{y}+\boldsymbol{\mu})}^T T(\mathbf{x}) - \mathbf{1}^T A(\mathbf{y}+\boldsymbol{\mu}) \nonumber\\ & - \frac{N}{2} \log 2\pi - \frac{1}{2} \log |\mathbf{\Sigma}| - \frac{1}{2} \mathbf{y}^T \mathbf{\Sigma}^{-1}\mathbf{y} - \log q(\mathbf{y})] \\
    = & \boldsymbol{\mu}^T T(\mathbf{x}) - \frac{1}{2} \log |\mathbf{\Sigma}| + \mathbb{E}_{q(\mathbf{y})} [\mathbf{y}^T T(\mathbf{x}) - \mathbf{1}^T A(\mathbf{y}+\boldsymbol{\mu}) \nonumber\\ & - \frac{1}{2} \mathbf{y}^T \mathbf{\Sigma}^{-1}\mathbf{y} - \log q(\mathbf{y})] \\
    = & \boldsymbol{\mu}^T T(\mathbf{x}) + \frac{1}{2} \log |\mathbf{L}| + \mathbb{E}_{q(\mathbf{y})} [\mathbf{y}^T T(\mathbf{x}) - \mathbf{1}^T A(\mathbf{y}+\boldsymbol{\mu}) \nonumber\\ & - \frac{1}{2} \mathbf{y}^T \mathbf{Ly} - \log q(\mathbf{y})] \\
    = & \boldsymbol{\mu}^T T(\mathbf{x}) + \frac{1}{2} \log |\mathbf{L}| + \bar{\mathbf{y}}^T T(\mathbf{x}) - \mathbf{1}^T \mathbb{E}_{q(\mathbf{y})} A(\mathbf{y}+\boldsymbol{\mu}) \nonumber\\ & - \frac{1}{2} \bar{\mathbf{y}}^T \mathbf{L}\bar{\mathbf{y}} - \frac{1}{2}Tr(\mathbf{L\Lambda}) + \frac{1}{2} \log |\mathbf{\Lambda}| \\
    = & \boldsymbol{\mu}^T T(\mathbf{x}) + \frac{1}{2} \log |\mathbf{L}| - \frac{1}{2}Tr(\mathbf{L}(\mathbf{\Lambda}+\bar{\mathbf{y}}\bar{\mathbf{y}}^T)) + \bar{\mathbf{y}}^T T(\mathbf{x}) \nonumber\\ & + \frac{1}{2} \log |\mathbf{\Lambda}| - \mathbf{1}^T \mathbb{E}_{q(\mathbf{y})} A(\mathbf{y}+\boldsymbol{\mu}) \\ 
    = & {(\boldsymbol{\mu} + \bar{\mathbf{y}})}^T T(\mathbf{x}) - \mathbf{1}^T \mathbb{E}_{q(\mathbf{y})} A(\mathbf{y}+\boldsymbol{\mu}) - \frac{1}{2}Tr(\mathbf{L}(\mathbf{\Lambda}+\bar{\mathbf{y}}\bar{\mathbf{y}}^T)) \nonumber\\ & + \frac{1}{2} \log |\mathbf{L}| + \frac{1}{2} \log |\mathbf{\Lambda}| ,
\end{align}
where $\mathbf{\Sigma}=\mathbf{L}^{\dagger}$.
The above objective function is for a single graph signal, we now sum over all graph signals $\mathbf{y}_j$ and write the above objective function in matrix form with respect to $\mathbf{y}$
\begin{align}
    & \Tr{((\bar{\mathbf{Y}}^T+\mathbf{1}{\boldsymbol{\mu}}^T) T(\mathbf{X}))} - \mathbf{1}^T \mathbb{E}_{q(\mathbf{Y})} A(\mathbf{Y}+\boldsymbol{\mu}{\mathbf{1}}^T) \mathbf{1} \nonumber\\ & - \frac{1}{2} \Tr{(\bar{\mathbf{Y}}^T \mathbf{L} \bar{\mathbf{Y}})} - \frac{N}{2} \Tr(\mathbf{L\Lambda}) + \frac{N}{2} \log |\mathbf{L}| \nonumber\\ & + \frac{N}{2} \log |\mathbf{\Lambda}|.
    \label{eq:elbo_matrix}
\end{align}
Note that we constraint that all $q(\mathbf{y}_j)$ have the same covariance $\mathbf{\Lambda}$.
Once $\mathbf{\Lambda}$ is fixed, maximizing Eq.~\eqref{eq:elbo_matrix} is equivalent to Eq.~\eqref{eq:variational_generalized_graph_learning} up to re-weighting.
For Poisson distributions, the last expectation term has a closed-form solution
\begin{equation}
\label{eq:variational_generalized_graph_learning}
    \mathbb{E}_{q(\mathbf{y})} A(\mathbf{y}+\boldsymbol{\mu}) = \exp{(\mathbf{y}+\boldsymbol{\mu}+\frac{1}{2}\mathbf{I}_N\mathbf{\Lambda})}.
\end{equation}
For other exponential family distribution such as Bernoulli, computing this term might need numerical approximation.

In Eq.~\eqref{eq:variational_generalized_graph_learning}, the mean parameter $\mathbf{y}_j$ plays a similar role as the smooth signal representation in Eq.~\eqref{eq:generalized_graph_learning}, and $\mathbf{\Lambda}_j$ accounts for the stochasticity of $\mathbf{y}$ which we ignored.
For simplicity, we fix $\mathbf{\Lambda}_j=r\mathbf{I}_N$.
We use $r=0.5$ for the synthetic experiments.

\bibliographystyle{IEEEtran}
\bibliography{references}

\end{document}